\documentclass[sn-apa]{sn-jnl}

\usepackage{graphicx}%
\usepackage{multirow}%
\usepackage{amsmath,amssymb,amsfonts}%
\usepackage{amsthm}%
\usepackage{mathrsfs}%
\usepackage[title]{appendix}%
\usepackage{xcolor}%
\usepackage{textcomp}%
\usepackage{manyfoot}%
\usepackage{booktabs}%
\usepackage{algorithm}%
\usepackage{algorithmicx}%
\usepackage{algpseudocode}%
\usepackage{wrapfig}
\usepackage{listings}%
\usepackage{soul}

\theoremstyle{thmstyleone}%
\theoremstyle{thmstyletwo}%

\theoremstyle{thmstylethree}%

\begin{document}

\title[Article Title]{Offline Reinforcement Learning for Learning to Dispatch for Job Shop Scheduling}


\author*[1]{\fnm{Jesse} \spfx{van} \sur{Remmerden}}\email{j.v.remmerden@tue.nl}

\author[1]{\fnm{Zaharah} \sur{Bukhsh}}\email{z.bukhsh@tue.nl}

\author[1]{\fnm{Yingqian} \sur{Zhang}}\email{yqzhang@tue.nl}

\affil*[1]{\orgdiv{Information Systems IE\&IS}, \orgname{Eindhoven University of Technology}, \orgaddress{\street{De Zaale}, \city{Eindhoven}, \postcode{5600 MB}, \country{Netherlands}}}


\abstract{The Job Shop Scheduling Problem (JSSP) is a complex combinatorial optimization problem. While online Reinforcement Learning (RL) has shown promise by quickly finding acceptable solutions for JSSP, it faces key limitations: it requires extensive training interactions from scratch leading to sample inefficiency, cannot leverage existing high-quality solutions \textcolor{black}{from} traditional methods like Constraint Programming (CP), \textcolor{black}{and require simulated environments to train in, which are impracticable to build for complex scheduling environments. We introduce Offline Learned Dispatching (Offline-LD), an offline reinforcement learning approach for JSSP, which addresses these limitations by learning from historical scheduling data. Our approach is motivated by scenarios where historical scheduling data and expert solutions are available or scenarios where online training of RL approaches with simulated environments is impracticable}. Offline-LD \textcolor{black}{introduces maskable variants of two Q-learning methods, namely, Maskable Quantile Regression DQN (mQRDQN) and discrete maskable Soft Actor-Critic (d-mSAC), that are able to learn from historical data, through Conservative Q-Learning (CQL). Moreover, we present a novel entropy bonus modification for d-mSAC, for maskable action spaces. Moreover, we introduce a novel reward normalization method for JSSP in an offline RL setting}. Our experiments demonstrate that Offline-LD outperforms online RL on both generated and benchmark instances \textcolor{black}{when trained on only 100 solutions generated by CP. Notably, introducing noise to the expert dataset yields comparable or superior results to using the expert dataset, with the same amount of instances, a promising finding for real-world applications, where data is inherently noisy and imperfect.}}




\maketitle

\section{Introduction}
The job shop scheduling problem (JSSP) is a widely known combinatorial optimization problem with critical applications in manufacturing, maintenance, and industrial operations~\citep{jssp_survey}. The objective of JSSP is to optimally schedule a set of \textit{jobs} on available \textit{machines}, typically aiming to minimize the total completion time (makespan). However, finding an optimal schedule is computationally intensive, as JSSP is an NP-hard problem.

The traditional \textcolor{black}{methods} for JSSP fall into two categories: exact and heuristic methods. Exact methods such as Constraint Programming (CP) \citep{cp_jssp} and Mathematical Programming~\citep{FAN2022105998} guarantee optimality but face scalability issues for large-sized instances. Therefore, in most cases, improve heuristics, such as Genetic Algorithms \citep{ga_jssp_survey} and construction heuristics, such as Priority Dispatching Rules (PDR)~\citep{dispatching_rules}, are preferred, as they might find acceptable solutions in a reasonable time. 
Recently, reinforcement learning has emerged as a promising approach to learn PDRs, with methods like Learning to Dispatch (L2D)~\citep{l2d} demonstrating that graph neural networks can learn policies that generalize well to larger instance sizes while trained on smaller ones. Learning-based approaches like L2D offer two key advantages over traditional methods. First, once trained, the learned policy can solve new instances orders of magnitude faster than exact solvers or evolutionary algorithms \citep{reijnen2023job}.  Second, these policies demonstrate robust performance in uncertain and dynamic environments, where traditional methods often struggle to adapt~\citep{smit2024graphneuralnetworksjob}. 

All existing \textcolor{black}{Deep Reinforcement Learning (DRL)} approaches to learn to dispatch are online RL methods\textcolor{black}{, whereby an RL agent interacts with a simulator. In this simulator, also known as environment, the RL agent learns by trial and error what the best scheduling policy is~\citep{l2d, smit2024graphneuralnetworksjob}}. Despite their compelling advantages, online RL approaches face fundamental limitations: they require extensive training in simulation environments. These simulations often fail to capture the full complexity of real-world scheduling problems, as creating accurate digital twins of (manufacturing) systems is non-trivial~\citep{jssp_digital_twin}. The gap between simplified simulations and real-world complexity significantly limits the practical deployment of online RL methods in industrial settings. \textcolor{black}{Furthermore, even when suitable simulations exist, the requirement of learning from scratch through trial-and-error interaction makes the online RL sample inefficient, whereby it can require millions of interactions with a simulation model.}


In contrast, offline reinforcement learning does not rely on simulations and tends to be more sample-efficient, making it a promising alternative for solving JSSP, or more generally, combinatorial optimization problems. However, developing offline RL for JSSP is non-trivial. Traditional online RL methods cannot be used directly to train with these existing datasets due to the distributional shift that exists between the dataset and when a trained method is deployed in the real world~\citep{sergey_survey}. \textcolor{black}{This distributional shift happens when an agent encounters situations or action choices that were not present in its training data. Due to this, an agent can overestimate the value of those unseen actions (called out-of-distribution actions (OOD)). When deployed in real-world setting, the agent might select these poorly understood OOD actions, resulting in poor or unsafe decisions.} Offline reinforcement learning presents a promising solution to these challenges, \textcolor{black}{by limiting the overestimating of OOD actions}. Offline RL has been widely studied for robotics~\citep{kumar2023pretrainingrobotsofflinerl}, where they encounter similar challenges of being difficult to simulate and having expert datasets, such as human demonstration, to train on. However, to the best of our knowledge, offline RL has not yet been researched for JSSP or any other combinatorial optimization problem.

\textcolor{black}{In this paper, we present Offline Learned Dispatching (Offline-LD), the first offline RL approach for learning dispatching policies in JSSP. Our method bridges the gap between traditional optimization and learning-based approaches by leveraging existing high-quality solutions while maintaining the generalization capabilities that make learning-based approaches attractive. By learning from existing data without environment interaction, Offline-LD can utilize high-quality solutions from traditional optimization algorithms, such as constraint programming methods, while avoiding the limitations of simulation environments. A key advantage of Offline-LD is its ability to learn comprehensive dispatching policies for JSSP by analyzing both optimal and suboptimal scheduling decisions in the training data. Our work offers the following contributions:}

 
\begin{itemize}
    \item We introduce Offline-LD, the first fully end-to-end offline RL method for JSSP that learns dispatching policies directly from existing (sub)optimal solutions without environment interaction.
    \textcolor{black}{\item We propose two Q-learning methods capable of masking infeasible actions. Building upon Conservative Q-Learning (CQL)~\citep{cql} for effective offline training, we also introduce novel approaches for entropy regularization and reward normalization to enhance learning.}
    \item We demonstrate that our Offline-LD achieves comparable or superior performance compared to the online RL counterpart.  
    \item  We show that Offline-LD can learn effective policies from a small training dataset of just 100 instances, highlighting its data efficiency compared to online RL approaches that require millions of interactions.
    \item We show that our Offline-LD method generalizes well to varying benchmarks and instance sizes, showing it ability to learn on smaller instance sizes and generalizing to larger ones.
    \item We demonstrate that incorporating ``noisy`` datasets in training leads to enhanced performance.
\end{itemize}

\section{Related Work}

Existing RL methods for JSSP are end-to-end online RL approaches, where RL agents directly learn a policy for JSSP by interacting with the environment, which is either a simulation model or simply an evaluation function. The Learning to Dispatch (L2D) method \citep{l2d} learns priority dispatching rules (PDR) by using a graph isomorphism network (GIN) to represent a disjunctive graph, trained by proximal policy optimization (PPO).  \cite{song2022flexible} develop an end-to-end online RL method to learn PDR to solve a flexible job shop scheduling (FJSP) problem, and \cite{wang2023flexible} propose DAN based Reinforcement Learning (DANIEL), which adopts self-attention models for solving FJSP. Once trained, these DRL approaches are much faster in solving large instances, compared to exact methods such as CP solvers.   


However, these online methods suffer from sample inefficiency and \textcolor{black}{do not} leverage existing data, including near-optimal examples generated by exact solvers. An alternative strategy is to employ offline RL, which allows using existing datasets. However, offline RL methods face challenges related to distributional shift~\citep{sergey_survey}. The distributional shift in RL is harder to deal with since taking the wrong actions can have compounding negative effects in a sequential decision setting. A common method to prohibit this distributional shift is to use regularization during training~\citep{offline_rl_survey}. One such method is Conservative Q-learning (CQL)~\citep{cql}, which ensures that the learned Q-function is the lower bound of the real Q-function. \textcolor{black}{A RL approach that uses a Q-learning method can be adapted to an offline RL approach since CQL is applied as a regularization term to the loss of the Q values.}

Another state-of-the-art offline RL method is Implicit Q-Learning (IQL)~\citep{iql}. IQL is different from CQL in that it does not regularize the Q-function, but rather IQL ensures that no out-of-distribution state-action pairs are queried during training. However, IQL has only been used for continuous action spaces~\citep{iql}. Therefore, IQL likely will learn sub-optimal policies for JSSP, in comparison to CQL, since JSSP has a discrete action space. Thus, in this paper, we will focus on CQL. Another approach for offline RL is to use transformers and model the RL problem as a sequence problem, which is learned \textcolor{black}{through} supervised \textcolor{black}{learning}~\citep{dt,sequence_transformer}. These methods have been shown to outperform both CQL and IQL in certain \textcolor{black}{offline RL} benchmarks; however, these methods have major downsides \textcolor{black}{in terms of efficiency and are} not invariant to the state and action space size. This makes them unsuitable for JSSP, where both the state and action space size is dependent on the instance size. 

Offline RL has \textcolor{black}{not been studied yet for} CO problems. Fully offline joint learners using behavioral cloning (BC) have been proposed \textcolor{black}{for the traveling salesman problem}~\citep{DaCosta2021}, and hybrid approaches have been explored, where BC is used to improve an online RL policy~\citep{zhang2024towards}, to jointly work with \textcolor{black}{Mixed-Integer Linear Programming (MILP)}. Recently, an approach for JSSP has been proposed by \citet{jssp_cp_rl}, whereby CP is combined with online RL to improve training; however, this approach still requires online training by interacting with an environment. The closest comparable fully offline RL method to ours is applied to an order dispatching problem~\citep{order_dispatching} that utilizes large-scale datasets with more than 20 million examples. In contrast, our approach achieves effective results using only 100 solutions \textcolor{black}{as training instances}, highlighting its efficiency and potential impact of offline RL for CO problems.

\section{Preliminaries}

\subsection{Job Shop Scheduling Problem} In JSSP, each problem instance has a set of job $\mathcal{J}$ and machines $\mathcal{M}$. Each job $J_{i} \in \mathcal{J}$ consists of a specific order of operations $O_{i,j} \in J_{i}$ that must be processed by $m_{i}$ machine in $\mathcal{M}$, so that the operations are processed as $O_{i,1}\rightarrow \dots \rightarrow O_{i,m_{i}}$. Moreover, each operation $O_{i,j}$ can only be processed by a specific machine $m_{i}$ and has processing time $p_{i,j} \in \mathbb{N}$. Each machine can only process a single job in a given timestep. The goal is to find a schedule that minimizes the makespan $C_{\max}=\max_{i,j}(C_{i,j}=Z_{i,j} + p_{i,j})$, where $Z_{i,j}$ is the starting time of operation $O_{i,j}$.

Any JSSP instance can be defined as a disjunctive graph $G=\left (\mathcal{O},\mathcal{C},\mathcal{D} \right )$~\citep{disjunctive_graph_jssp}. In this representation, $\mathcal{O}= \left \{O_{i,j} \mid \forall i,j \right\} \cup \left \{\text{Start},\text{End} \right \}$ is the set of nodes, which are the operations, including Start and End, which are dummy nodes representing the start and termination respectively, and have a processing time of zero. $\mathcal{C}$ is a set of undirected edges, which connect operation nodes in $\mathcal{O}$ that require the same machines. $\mathcal{D}$ is a set of directed edges representing the precedence of operations in a given job. A valid schedule of a JSSP instance can be found by setting the direction of all the edges in $\mathcal{C}$, such that $G$ becomes a DAG~\citep{l2d}.

\subsection{Offline Reinforcement Learning}
A reinforcement learning problem can be formulated as a Markov Decision Process (MDP) $\mathcal{M}=\langle S, A, P, R, \gamma\rangle$, where $S$ are the states, $A$ is the set of possible actions, $P:S \times A \times S \rightarrow [0,1]$ is the transition function, $R: S \times A \times S \rightarrow \mathbb{R}$ is the reward function, $\gamma$ the discount factor that determines the importance of future rewards. $Q(s_t, a_t)$, represents the Q-value and is the expected return when action $a_t$ is taken at step $s_t$. Moreover, this paper considers maskable action spaces, where the action space depends on the current state $A(s_{t}) \subseteq A$. In offline RL, a policy is not learned through interaction but rather through a fixed dataset $D= \{(s,a,r(s,a),{s}', {a}')_{i} \}$, where ${s}'$ and ${a}'$ are the next state and action.

\textcolor{black}{\section{Offline Learned Dispatching}}

In this paper, we \textcolor{black}{introduce Offline Learned Dispatching (Offline-LD)}, an end-to-end offline RL method for JSSP. \textcolor{black}{In this section, we first state the formulation of the Markov Decision Process (MDP). Afterwards, we explain our method, Offline-LD, in which we detail our two proposed RL methods for maskable action spaces, Maskable Quantile Regression Deep Q-Learning (mQRDQN) and Discrete Maskable Soft Actor-Critic (d-mSAC), and how they can learn from historical data using Conservative Q-learning (CQL)~\citep{cql}. Lastly, we show how we generate our datasets and explain our proposed reward normalization method. Algorithm \ref{code:pseudo_cql} shows the pseudocode of Offline-LD. }

\textcolor{black}{The proposed Offline-LD method use a Markov Decision Process (MDP), based on the one first introduced by \cite{l2d} for JSSP. Whereby the MDP is formulated as follows:}
\begin{itemize}

\item \textbf{State Space:} The state $s_t \in S$ is a disjunctive graph $G(t)=(\mathcal{O}, \mathcal{C} \cup \mathcal{D}_{u}(t), \mathcal{D}_{t})$, whereby $\mathcal{D}_{u}(t)$ are the (directed) edges that have been assigned before time step $t$ and $\mathcal{D}(t)$ are the edges to be directed. Each node has two features, namely $I(O, s_{t})$ a binary indicator to signify if operation $O$ has been scheduled at timestep $t$, and $C_{\text{LB}}(O, s_t)$, which is the lower bound of the completion time of operation $O$.

\item \textcolor{black}{\textbf{Action Space:} At each timestep $t$, the available actions $a_{t} \in A(s_{t})$ are the current operations that can be scheduled. Whenever, all the operation in a job are scheduled, we mask those actions, resulting in the size of the action space being smaller.}


\item \textbf{State Transition:} When an action $a_{t}$ is selected, the operation is allocated to the required machine when available. The disjunctive graph is updated accordingly.

\item \textbf{Reward Function:} The reward function measures the quality difference between partial solutions, calculated by: 
\begin{equation}  
R(s_t, a_t, s_{t+1})=\max_{i,j}\left \{C_{\text{LB}}(O_{i,j}, s_t) \right\} - \max_{i,j}\left \{C_{\text{LB}}(O_{i,j}, s_{t+1}) \right\},
\end{equation}
\textcolor{black}{where, $C_{\text{LB}}(O_{i,j}, s_{t})$ is the estimated completion time of operation $O_{i,j}$ at state $s_t$ at timestep $t$. The reward is therefore the difference between the estimated completion time between $s_{t+1}$ and $s_t$.}

\end{itemize}

\textcolor{black}{
\subsection{Conservative Q-Learning}
}

\begin{algorithm}[h!]

\caption{\textcolor{black}{
Offline-LD}}\label{code:pseudo_cql}
\begin{algorithmic}[1]
\color{black}
\State \textbf{Input:} Dataset $D= \{(s_t,a_t,r_t,s_{t+1})_{i} \}$, discount factor $\gamma$, CQL regularization coefficient $\alpha$, Number of training steps $T$, Target update rate $\eta$

\State \textbf{Initialize:} Q-network $Q_\theta$, target network $Q_{\theta'}$
\For{$i \gets 1$ to $T$}
    \State Sample batch $B = \{(s,a,r,s')\}$ from $D$.
    \State Compute TD error ($\mathcal{B}Q(s_{t+1})$ depends on the Q-learning method):
    \[
    J_{Q}(\theta) \gets \mathbb{E}_{s_t,a_t,r_t,s_{t+1} \sim B} \left [ Q_\theta(s_t,a_t) - r_t + \gamma \mathcal{B}Q_{\theta'}(s_{t+1}) \right ]
    \]

    \State Conservative penalty for discrete actions:
    \[
    J_{\text{CQL}}(\theta) =\alpha_{\text{CQL}} \mathbb{E}_{s \sim B} \Biggl[ 
    \log \sum_{a \in A(s)} \exp(Q_{\theta}(s, a)) - \mathbb{E}_{a \sim B}[Q_{\theta}(s, a)] 
    \Biggr] 
    \]
    \State Total Q loss:
    \[
    J(\theta) \gets J_{\text{CQL}}(\theta) + \frac{1}{2} J_{Q}(\theta)
    \]
    \State Apply loss to Q-network:
    \[
        \theta \gets \theta + \nabla_{\theta}\mathcal{L}(\theta)
    \]
    \If {$i \mod \eta =0$}
    \State Set the weights of the target network to the Q network:
    \[
        {\theta}' \gets \theta
    \]
    \EndIf
\EndFor

\end{algorithmic}

\end{algorithm}

With offline RL, the goal is to learn optimal policies from historical data sources, whereas online RL learns through interacting in an environment. An issue that arises from learning through data is the distributional shift between the data and when the learned policy is used in a real-world setting~\citep{sergey_survey}. When training data do not contain adequate information about certain scenarios, the policy might take actions not present in the data, \textcolor{black}{leading to} out-of-distribution action (OOD). For online RL, learning to take an OOD action is not an issue, since it can explore these actions and discover \textcolor{black}{their impact, during training}. However, with offline RL, this is not possible, since we do not interact with an environment and thus cannot explore the outcome of OOD actions.

Conservative Q-learning (CQL)~\cite{cql} \textcolor{black}{alleviates the} issue of OOD actions by being pessimistic about \textcolor{black}{Q values of} OOD actions. CQL introduces a regularization term to the loss of Q-networks, that minimizes the Q-values for OOD actions. With CQL, the loss for the Q-networks is formulated as:
\begin{align}
    J(\theta) =&\alpha_{\text{CQL}} \mathbb{E}_{s \sim D} \Biggl[ 
    \log \sum_{a \in A(s)} \exp(Q_{\theta}(s, a)) - \mathbb{E}_{a \sim D}[Q_{\theta}(s, a)] 
    \Biggr] \label{eq:cql_term} \\
    &+ \frac{1}{2} \mathbb{E}_{s, a, s', a' \sim D} \left[ 
    \left( Q_{\theta}(s, a) - \mathcal{B}Q(s', a') \right)^2 
    \right], \label{eq:bellman_update}
\end{align}
where the first part (Eq. \ref{eq:cql_term}) is the CQL regularizer and the second part (Eq. \ref{eq:bellman_update}) is the normal Q update, where $\mathcal{B}Q(s',a')$ is the target used by \textcolor{black}{either our proposed Maskable Quantile Regression Deep Q Networks (mQRDQN) or Discrete Maskable Soft Actor-Critic(d-mSAC), introduced in detail in the following sections. Alg.~\ref{code:pseudo_cql} shows how Offline-LD utilizes Conservative Q-Learning (CQL). Offline-LD calculates the TD-loss of either mQRDQN or d-mSAC. Afterwards, the CQL regression term is applied to ensure that high Q values are penalized for out-of-distribution (OOD) actions}. $\alpha_{\text{CQL}}$ determines the strength of the CQL regularizer, whereby the general rule is that less optimal datasets require a higher value $\alpha_{\text{CQL}}$. We made an adjustment to the CQL regularization term (Eq. \ref{eq:cql_term}), namely that we only apply the log-sum exponent on the actions available $a \in A(s)$ at the given state $s$. The use of log-sum-exponent requires that for all masked actions, the state-action value is set to negative infinity $Q(s,a)=-\infty,\, \forall a \notin A(s)$. Setting it to negative infinity ensures that the given action does not affect CQL since $\exp(-\infty)=0$.
\subsection{Maskable Quantile Regression DQN}
Quantile Regression DQN (QRDQN)~\citep{qrdqn} expands on \textcolor{black}{Deep Q-Learning} by learning a value distribution of the expected return for state-action pair. This value distribution is learned as a discrete set of $N$ quantiles. Therefore, the state-action value is represented as $Q_{\theta}(s,a)=\frac{1}{N}\sum^{N}_{i=1}\delta_{\theta_{i}}(s,a)$, which is the mean of all the quantiles $\delta_{\theta_{i}}$, whereby the network architecture must be adjusted to output $N$ quantiles for each action.
\begin{align}
    \mathcal{B}Q({s}',{a}') =& r + \gamma \mathbb{E}\left [ \underset{a \in A({s}')}{\max}Q({s}',a') \right ] \label{eq:dqn_up}  \\
    \mathcal{B}\delta_{\theta_{i}}({s}',{a}') =& r + \gamma\delta_{\theta_{i}}({s}',a_{\max}), \forall j\textcolor{black}{,} \label{eq:app_qrdqn_up}
\end{align}
\textcolor{black}{where $r$ is the reward. }In QRDQN, the DQN Bellman error update, \textcolor{black}{is shown in} Eq. \ref{eq:dqn_up}), is \textcolor{black}{adapted} to Eq. \ref{eq:app_qrdqn_up}, whereby $a_{\max}=\text{arg}\,\underset{{a}'\in A({s}')}{\text{max}}Q({s}',{a}')$. For mQRDQN, we only consider valid actions in the next state ${s}'$. This Bellman error update (Eq. \ref{eq:app_qrdqn_up}) is used to calculate the loss as follows:
\begin{equation}
    J_{Q}(\theta) = \sum_{i=1}^{N}\mathbb{E}_{j} \left [\rho_{\hat{\tau}_i}^{\kappa} \left ( \mathcal{B}\delta_{\theta_{j}}({s}',{a}') - \delta_{\theta_{i}}({s}',{a}')\right) \right],
\end{equation}
where $\rho_{\hat{\tau}_i}^{\kappa}$ is the quantile Huber loss. This loss is formulated as:
\begin{equation}
    \rho_{\hat{\tau}_i}^{\kappa}(u)= \left |\tau - \delta_{\left \{ u < 0\right\}} \right | \mathcal{L}_{\kappa}(u),
\end{equation}
where \textcolor{black}{$\tau$ is the current quantile and} $\mathcal{L}_{\kappa}(u)$ is the standard Huber loss.

\subsection{Discrete Maskable Soft Actor-Critic}
Discrete Soft Actor-Critic (d-SAC)~\citep{discrete_sac} is an actor-critic \textcolor{black}{method} for discrete action spaces. In d-SAC, a policy $\pi_{\phi}$ is learned using two Q-networks $Q_{\theta_{1}}$, and $Q_{\theta_{2}}$. \textcolor{black}{These Q networks act as critics and are used to update the policy $\pi_{\phi}$. The two Q-networks are used to minimize the effect of Q-value overestimation. However,  d-SAC does not consider a maskable action space. Therefore, we introduce discrete maskable Soft-Actor Critic (d-mSAC), to enable d-mSAC to learn in maskable action spaces.}

The policy loss of d-mSAC with two Q-networks is computed as follows:
\begin{equation}\label{eq:app_sac_policy}
    J_{\pi}{(\phi)}= \mathbb{E}_{s \sim D}  [\pi_{\phi}(s)^{T}  (\alpha_{\text{temp}}\log(\pi_{\phi}(s)) - \min(Q_{\theta_{1}}(s),Q_{\theta_{2}}(s)) )].
\end{equation}

$\alpha_{\text{temp}}$ is the entropy regularization term and encourages exploration in online RL and prevents the policy $\pi$ collapsing to a single action in offline RL. No changes had to be made for \textcolor{black}{Eq.~\ref{eq:app_sac_policy}} for a maskable action space since if $a \notin A(s)$, the $\pi_{\phi}(a\mid s)=0$. This ensures that the masked actions are not considered with the update.
\begin{equation} \label{eq:app_q_bellman_sac}
J_{Q}(\theta)= \mathbb{E}_{s,a,{s}',{a}' \sim D} \left [ \left (Q_{\theta}(s,a)-\mathcal{B}Q({s}',{a}') \right)^{2}\right ].
\end{equation}
The loss for the Q-networks (Eq. \ref{eq:app_q_bellman_sac}) is the mean square error between the prediction and the Bellman update error calculated through:
\begin{equation} \label{eq:sac_standard_up}
    \mathcal{B}Q({s}',{a}') = r + \gamma \mathbb{E}[\pi_{\phi}({s}')\min(Q_{\theta_1}({s}'),Q_{\theta_2}({s}'))] 
\end{equation}
In Eq.~\ref{eq:sac_standard_up} policy $\pi$ is used to calculate the target for the Q-network. \textcolor{black}{Therefore, the whole probability distribution of $\pi$ is used to calculate the target, such that the probability of all the actions is taken into account.} 

Previously, we stated that d-SAC and d-mSAC uses entropy regularization through $\alpha_{\text{temp}}$. $\alpha_{\text{temp}}$ is either a set hyperparameter or a learnable parameter~\citep{discrete_sac}. However, Christodoulou et al. did not consider maskable action spaces, meaning that $\alpha_{\text{temp}}$ is not learnable for \textcolor{black}{maskable action spaces} without modifying the update.

When $\alpha_{\text{temp}}$ is learned, $\alpha_{\text{temp}}$ is updated based on the current and target entropy. Within discrete SAC~\citep{discrete_sac}, this is updated through:
\begin{equation} \label{eq:target_entropy_static}
   J(\alpha_{\text{temp}}) = \pi_{\phi}(s_{t})^{T}[-\alpha_{\text{temp}}(\log(\pi_{\phi}(s_{t})) + \bar{H})], 
\end{equation}
where $\bar{H}$ is the desirable target entropy and is a set hyperparameter. This does not function with a maskable action space because we require different target entropies for different action space sizes. \textcolor{black}{Therefore, we reformulate the calculation of the entropy loss of Eq. \ref{eq:target_entropy_static} to:
\begin{equation}\label{eq:target_entropy_dynamic}
    J(\alpha_{\text{temp}}) = \pi_{\phi}(s_{t})^{T}\bigg[-\alpha_{\text{temp}}\log(\pi_{\phi}(s_{t})) + c_{\hat{H}} \log\big(|A(s_t)|\big)\bigg], 
\end{equation} 
where $A(s_t)$ is the action space at state $s_{t}$, and $c_{\hat{H}}$ is a hyperparameter that should be between 0 and 1 and sets how much the target entropy should be compared to the maximum entropy. This maximum entropy in Eq.~\ref{eq:target_entropy_dynamic} is represented as $\log\big(|A(s_t)|\big) = -\sum_{i=1}^{|A(s_t)|}\frac{1}{|A(s_t)|}\log\left(\frac{1}{|A(s_t)|}\right)$, the maximum entropy for a discrete action space of size $|A(s_t)|$~\citep{joram_soch_2025_14646799}.}  


\subsection{Network Architecture}
Offline-LD adapts the network architecture of L2D \citep{l2d}, which uses a modified \textcolor{black}{Graph Isomorphism Networks} (GIN) \citep{gin_network} with two key enhancements.  First, the GIN network is adapted to handle directed graphs, enabling the representation of job-precedence relationships. Second, instead of processing the complete disjunctive graph at each timestep $t$, the architecture considers a simplified graph as input $\bar{G}_{\mathcal{D}}=(\mathcal{O}, \mathcal{C} \cup \mathcal{D}_{u}(t))$ that excludes undirected disjunctive edges.  The GIN network outputs two types of embeddings: node embeddings $h_{O}(s_t)$ for all available operations $O \in \mathcal{O}$, and a graph embedding $h_{G}(s_t)$ calculated as the mean pool \textcolor{black}{the embeddings of} of available operations $O$ at timestep $t$. These embeddings are concatenated and used as input for the Q-networks $Q_{\theta}(h_{O}(s_t) \mathbin\Vert h_{G}(s_t))$ and policy network as $\pi_{\phi}(h_{O}(s_t) \mathbin\Vert h_{G}(s_t))$.

To support the offline learning, we added Dropout in the Q-networks of both d-mSAC and mQRDQN. Although dropout is typically avoided in online RL due to its variance-inducing properties, it serves as an effective regularizer in the offline setting with CQL, helping prevent overfitting in Q-networks \citep{workflow_offline_rl}.




\subsection{Dataset Generation} \label{sec:method_dataset_generation}
\begin{figure}[ht]
    \centering
    \includegraphics[width=1\columnwidth]{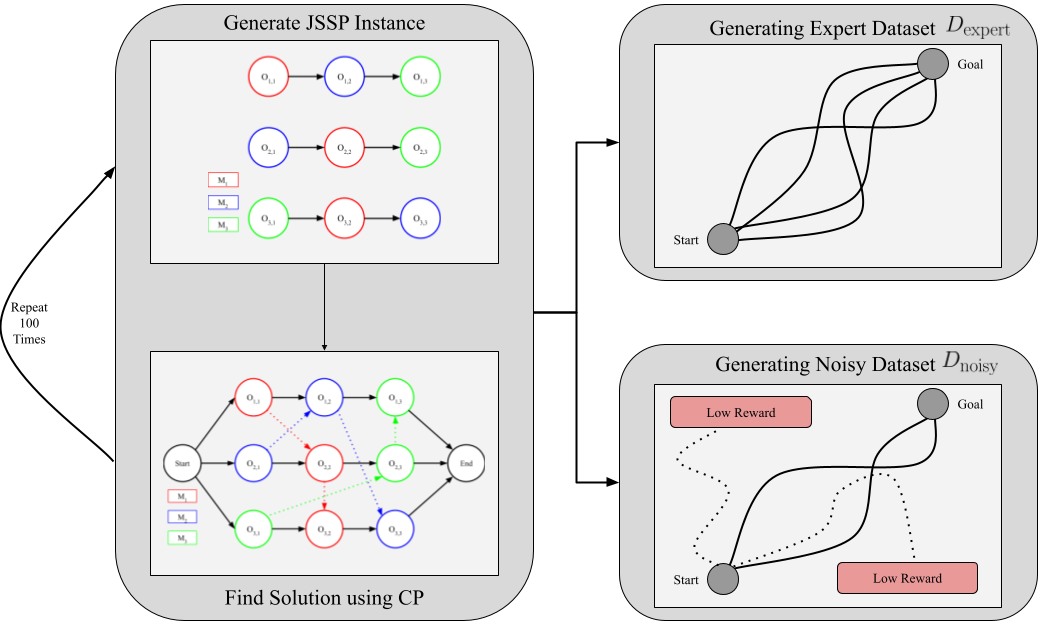}
    \caption{For the datasets, we start by generating 100 instances. We use constraint programming to find valid solutions for these instances. From these 100 solutions, we create an expert dataset $D_\text{expert}$ and a noisy dataset $D_\text{noisy}$. \textcolor{black}{The expert dataset $D_\text{expert}$ use the original solutions found through the constraint programming, whereas the noisy-expert dataset $D_\text{noisy}$ modifies 50\% of the solution, by changing the scheduled operations.}}
    \label{fig:dataset_generation_image}
\end{figure}
An effective offline RL dataset generation strategy is essential to the performance since the collected datatset must address the exploration-exploitation trade-off. Therefore, the training dataset \textcolor{black}{must satisfy two key properties: contain enough high-reward trajectories and ensure enough state space coverage~\citep{offline_dataset}. Most existing research on offline RL is based on robotics, whereby the collected dataset is either human demonstrations~\citep{offline_human} or other RL agents~\citep{offline_dataset}. In our case, we propose a  specialized dataset generation strategy  for JSSP.}

Figure~\ref{fig:dataset_generation_image} provides an overview of our dataset generation process. It begins by creating JSSP instances with $|\mathcal{J}|$ jobs and $|\mathcal{M}|$ machines. We use Constraint Programming (CP) to find optimal or near-optimal solutions for these instances. This approach ensures that our expert dataset $D_{\text{expert}}$ contains high-reward trajectories that typically difficult to discover for online RL methods~\citep{l2d}.

\textcolor{black}{Note that data sets generated only through CP lack diversity, as they contain near-optimal or optimal trajectories}. To address this, we generate noisy datasets $D_{\text{noisy}}$ that include counterfactual examples, showing the agent both suboptimal actions and their consequences~\citep{bc_offline_rl}. Typically, an expert policy selects $\epsilon$-greedy actions to generate noisy datasets. However, taking $\epsilon$-greedy actions with CP would result in almost no high-reward trajectories since the solutions generated by CP are static and thus cannot adapt to noisy actions. This would likely result in suboptimal trained policies~\citep{10.5555/3666122.3666343}.

Instead, we introduce noise selectively: each episode/solution has probability $p_{\text{noisy}}$ of containing noise, with noisy actions generated using $\epsilon$-greedy policy. When this process generates invalid actions, we randomly select from the available valid actions, maintaining solution feasibility while introducing controlled variability.


\subsection{Reward Preprocessing}
A common issue whenever RL is applied to JSSP and other combinatorial optimization problems is designing the reward function. The main challenge is that the maximum obtainable return can vary significantly between instances of the same problem~\citep{rl_co_survey}. For example, in JSSP, the minimum obtainable makespan for two instances of the same size can differ substantially due to varying processing times. Therefore, any online RL method should learn both which actions yield higher returns and how to value these returns for each specific instance, increasing the learning complexity. A simple solution would be to normalize the reward based on some variable of the given problem. However, it is impractical in online RL settings where training instances are generated dynamically, as computing normalization factors would either impose significant computational overhead or yield unreliable results.

However, with offline RL, we can use techniques similar to preprocessing training datasets for supervised learning; namely, we normalize the reward based on the near-optimal or optimal makespan found by CP. This normalization would not affect the runtime, unlike with online RL, because the dataset is static, and preprocessing is done only once. \textcolor{black}{We normalize the reward through the following equation:}
\begin{align}
    N{(D, D_{\text{expert}})} = &\frac{r_i}{C_{\max, i}}, \forall i \in \{1, 2, \ldots, |D|-1, |D|\} \nonumber\\
    &\text{where } C_{\max, i} \in D_{\text{expert}}, \label{eq:norm_reward}
\end{align}
where \textcolor{black}{$D$ is the dataset, $D_{\text{expert}}$ is the expert dataset, and} $C_{\max, i} \in D_{\text{expert}}$ is corresponding episodes makespan found in $D_{\text{expert}}$, for reward $r_{i}$. This normalization scheme offers two key advantages: it standardizes reward magnitudes across different instances and ensures consistent scaling of rewards in $D_\text{noisy}$ by leveraging the expert dataset's makespan $C_{\max, i} \in D_{\text{expert}}$ as the normalization factor.
\section{Experiments and Results}
\subsection{Experimental Setup}
For our experiments, we generated training sets of 100 instances for \textcolor{black}{each of} five problem sizes (number of jobs and number of machines): $6\times 6$, $10 \times 10$, $15 \times 15$, $20 \times 20$, and $30 \times 20$. Following the standard set by \citet{TAILLARD1993278}, we assign each operation a processing time between 1 and 99. We used the Constraint Programming (CP) implementation provided in \citet{reijnen2023job} to find a solution for each generated instance, with a time limit of 60 minutes. These CP solutions were used to create two distinct training datasets: an expert dataset $D_{\text{expert}}$ containing the optimal \textcolor{black}{or near optimal} CP solutions, and a noisy-expert dataset $D_{\text{noisy}}$ generated following the procedure described in Section \ref{sec:method_dataset_generation}, using parameters $p_\text{noisy}=0.5$ and $p_{\epsilon}=0.1$.

\textcolor{black}{
For evaluation and testing, we generated the following three different types of problem instances: (1) 100 new instances for each problem size of $6\times 6$, $10 \times10$, $15\times 15$, $20 \times 20$, and $30 \times 20$. These were generated in the same way as the training data; (2) 100 new instances for each of the following problem sizes: $20 \times 10$, $20 \times 15$, $30 \times 15$, $50 \times 20$, and $100 \times 20$. Since these sized instances are not seen in the training data, we would like to test whether the learned policies on specific sized data can perform well on these different sized instances; and in additionally (3) the well-known benchmark instances, including Taillard~\citep{TAILLARD1993278}, and Demirkol~\citep{DEMIRKOL1998137}, which are used to test whether the learned policies can generalize well to instances different than the ones it is trained on. The Taillard instances show mainly different types of complexity in the number of jobs and machines, whereas the Demirkol have a larger range processing times than the training instances and show how Offline-LD generalizes to unseen processing times.}

 \textcolor{black}{We use the performance gap as measurement of performance, with a lower gap signifying better performance. The performance gap for our Offline-LD method and baselines is calculated relative to a reference makespan $C^*$. This reference $C^*$ is defined as: (1) for generated instances, the makespan found by Constraint Programming (CP) with a time-limit of 60 minutes; (2) for the Taillard and Demirkol benchmark instances, the best known upper bound makespan\footnote{Upper bounds obtained from https://optimizizer.com/TA.php and https://optimizizer.com/DMU.php}, which is the best found solution for a given instance. Given $C$ as the makespan found by the evaluated method, the gap is computed as: $\text{Gap}=\frac{C - C^{*}}{C^{*}} \times 100.$} 



We tested Offline-LD with mQRDQN and d-mSAC to compare the effectiveness of a value-based and actor-critic method for offline RL in JSSP. We trained both methods on $D_{\text{expert}}$ and $D_{\text{noisy}}$. We compared Offline-LD with L2D~\citep{l2d} to assess the relative performance of offline versus online RL approaches in JSSP. Our evaluation also includes Behavioral Cloning (BC) trained on $D_{\text{expert}}$ and three widely-used Priority Dispatching Rules (PDRs): Shortest Processing Time (SPT), Most Operation Remaining (MOR), and Most Work Remaining (MWKR). These PDRs were selected based on their established effectiveness in JSSP applications~\citep{dispatching_rules}.

We trained each RL method, including L2D, five times using seeds 600-604. We trained the offline RL approaches for $50,000$ training steps, whereas L2D is trained for $10,000$ episodes, with four different trajectories collected per episode~\citep{l2d}. \textcolor{black}{This means that L2D required a minimum of 1,440,000 training steps for $6 \times 6$ and up to 24,000,000 training steps for $30\times 20$}. For dataset generation, we used seed 200 for the training set, 300 for the evaluation set, and 0 for creating $D_{\text{noisy}}$.  We used $\gamma=1$ and a network architecture consisting of a two-layer GIN setup for the graph encoder (with sum node aggregation and average graph pooling), and a two-layer MLP ([128, 32, 1])  for network outputs.  For Offline-LD, we employed Dropout with $p=0.4$ in the MLPs and set $\alpha_{CQL}=1$. The target entropy ratio for d-mSAC is set to $c_{\bar{H}}=0.98$ and mQRDQN uses $N=32$ quantiles. All networks were trained with a learning rate of \textcolor{black}{$2 \times 10^{-5}$} and batch size 64, with target Q-networks updated every 2500 steps. A complete list of hyperparameters is provided in Table \ref{tab:hyperparameters}. The experiments were carried out on a single NVIDIA A100 GPU with 256 GB of RAM and an Intel Xeon Platinum 8360Y CPU. We made our code available\footnote{https://github.com/jesserem/Offline-LD}.

\begin{table}[ht]
\caption{Hyperparameters used in the experiments.}
\centering
\begin{tabular}{p{6cm}c}
\toprule
\multicolumn{2}{c}{\textbf{Shared Parameters}} \\
\midrule
Learning Rate & $2 \times 10^{-5}$  \\
Batch Size & 64 \\
Discount Factor ($\gamma$) & 1.0 \\
$\alpha_{\text{CQL}}$ & 1.0 \\
\midrule
\multicolumn{2}{l}{\textit{Network Architecture}} \\
MLP Architecture & [128, 32, 1] \\
MLP (GIN) & [64, 64, 64] \\
GIN Layers & 2 \\
GIN Node Aggregation & Sum \\
GIN Graph Pool & Average \\
Activation & ReLU \\
Dropout Rate ($p_{\text{dropout}}$) & 0.4 \\
Target Update Step Q-networks & 2500 \\
\midrule
\multicolumn{2}{c}{\textbf{Algorithm-Specific Parameters}} \\
\midrule
\multicolumn{2}{l}{\textit{mQRDQN}} \\
Number of Quantiles ($N$) & 32 \\
\midrule
\multicolumn{2}{l}{\textit{d-mSAC}} \\
Target Entropy Ratio ($c_{\bar{H}}$) & 0.98 \\
\bottomrule
\end{tabular}

\label{tab:hyperparameters}
\end{table}
\subsection{Results}

    
\begin{sidewaystable}
\caption{The results for the generated instances of size $6\times6$, $10\times 10$, $15\times 15$, $20\times 20$, and $30\times 20$, for which the RL approaches are trained. \textcolor{black}{\textbf{Gap}: Difference from CP solution. \textbf{Time}: Runtime (s). \textbf{Bold}: Best RL approach. *: Offline-LD significantly outperforms L2D (t-test, $p=0.05$). Opt. Rate (\%): Percentage of instances solved optimally by CP}}
\centering
\begin{tabular}{lccccccccccc}
\toprule
\multirow{2}{*}{Method} & \multicolumn{2}{c}{$6\times6$} & \multicolumn{2}{c}{$10\times10$} & \multicolumn{2}{c}{$15\times15$} & \multicolumn{2}{c}{$20\times20$} & \multicolumn{2}{c}{$30\times20$} \\
\cmidrule(lr){2-3} \cmidrule(lr){4-5} \cmidrule(lr){6-7} \cmidrule(lr){8-9} \cmidrule(lr){10-11}
& Gap & Time & Gap & Time & Gap & Time & Gap & Time & Gap & Time \\
\midrule
\multicolumn{11}{l}{\textbf{PDR}} \\
\midrule
SPT & 41.6\% & 0.01 & 51.7\% & 0.01 & 58\% & 0.02 & 63.5\% & 0.05 & 67.1\% & 0.09 \\
MOR & \textcolor{black}{20.7}\% & 0.01 & \textcolor{black}{33}\% & 0.01 & \textcolor{black}{41.4}\% & 0.02 & \textcolor{black}{47.1}\% & 0.05 & \textcolor{black}{49.1}\% & 0.09 \\
MWKR & \textcolor{black}{32.4}\% & 0.01 & \textcolor{black}{47.5}\% & 0.01 & \textcolor{black}{57.6}\% & 0.02 & \textcolor{black}{61.5}\% & 0.06 & \textcolor{black}{69.4}\% & 0.12 \\
\midrule
\multicolumn{11}{l}{\textbf{Baselines}} \\
\midrule
L2D & 15.8\%±6.5\% & 0.05 & 25.3\%±8.5\% & 0.14 & 28.1\%±5.3\% & 0.35 & 30.8\%±7.3\% & 0.7 & 29.4\%±2.3\% & 1.22 \\
BC-$D_{\text{expert}}$ & 30.9\%±7\% & 0.05 & 33.2\%±8.0\% & 0.14 & 37.1\%±6.7\% & 0.35 & 39.4\%±6.1\% & 0.7 & 42.6\%±5.2\% & 1.21 \\
\midrule
\multicolumn{11}{l}{\textbf{Offline-LD}} \\
\midrule
mQRDQN-$D_{\text{expert}}$ & 14.5\%±1\%$^*$ & 0.04 & 20.6\%±1.8\%$^*$ & 0.12 & \textbf{26.0\%±1.4\%}$^*$ & 0.31 & 27.7\%±1\%$^*$ & 0.62 & 31\%±4.6\% & 1.17 \\
d-mSAC-$D_{\text{expert}}$ & 14.5\%±2.6\%$^*$ & 0.05 & 21\%±2.5\%$^*$ & 0.14 & 26.3\%±3.3\%$^*$ & 0.35 & 28.5\%±4\%$^*$ & 0.7 & 30\%±4.1\% & 1.26 \\
mQRDQN-$D_{\text{noisy}}$ & 14.5\%±1.4\%$^*$ & 0.04 & \textbf{20.4\%±0.6\%}$^*$ & 0.12 & \textbf{26.0\%±0.5\%}$^*$ & 0.32 & \textbf{27.6\%±1.7\%}$^*$ & 0.63 & 29\%±2\%$^*$ & 1.17 \\
d-mSAC-$D_{\text{noisy}}$ & \textbf{14.3\%±1.5\%}$^*$ & 0.05 & 20.7\%±0.9\%$^*$ & 0.14 & 26.1\%±1.8\%$^*$ & 0.36 & 27.7\%±3.1\%$^*$ & 0.69 & \textbf{28.9\%±3.7\%} & 1.26 \\
\midrule
Opt. Rate (\%) & \multicolumn{2}{c}{100\%} & \multicolumn{2}{c}{100\%} & \multicolumn{2}{c}{100\%} & \multicolumn{2}{c}{5\%} & \multicolumn{2}{c}{18\%} \\
\bottomrule
\end{tabular}

\label{tab:generated_instances}
\end{sidewaystable}
\begin{sidewaystable}
\caption{The results of the Taillard and Demirkol benchmark instances. \textcolor{black}{\textbf{Gap}: Difference from the upper bound of the best-known solution. \textbf{Time}: Runtime (s). \textbf{Bold}: Best RL approach. *: Offline-LD significantly outperforms L2D (t-test, $p=0.05$). Opt. Rate (\%): Percentage of instances for which an optimal result is found.}}
\centering
\begin{tabular}{lccccccccccc}
\toprule
\multirow{2}{*}{Method} & \multicolumn{2}{c}{Taillard $15\times15$} & \multicolumn{2}{c}{Taillard $20\times20$} & \multicolumn{2}{c}{Taillard $30\times20$} & \multicolumn{2}{c}{Demirkol $20\times20$} & \multicolumn{2}{c}{Demirkol $30\times20$} \\
\cmidrule(lr){2-3} \cmidrule(lr){4-5} \cmidrule(lr){6-7} \cmidrule(lr){8-9} \cmidrule(lr){10-11}
& Gap & Time & Gap & Time & Gap & Time & Gap & Time & Gap & Time \\
\midrule
\multicolumn{11}{l}{\textbf{PDR}} \\
\midrule
SPT & 56.9\% & 0.02 & 65.3\% & 0.04 & 67.3\% & 0.12 & 64.8\% & 0.05 & 62.2\% & 0.10 \\
MOR & \textcolor{black}{41.4}\% & 0.02 & \textcolor{black}{44.4}\% & 0.04 & \textcolor{black}{54.6}\% & 0.09 & \textcolor{black}{58.1}\% & 0.05 & \textcolor{black}{64.2}\% & 0.09 \\
MWKR & \textcolor{black}{54.3}\% & 0.02 & \textcolor{black}{62.8}\% & 0.05 & \textcolor{black}{67.7}\% & 0.11 & \textcolor{black}{70.2}\% & 0.05 & \textcolor{black}{89.7}\% & 0.14 \\
\midrule
\multicolumn{11}{l}{\textbf{Baselines}} \\
\midrule
L2D & 27.4\%±5\% & 0.35 & 31.8\%±6.7\% & 0.70 & 33.6\%±2.8\% & 1.23 & 34.4\%±3.6\% & 0.70 & 36.7\%±1.8\% & 1.24 \\
BC-$D_{\text{expert}}$ & 36.1\%±6.7\% & 0.35 & 41.2\%±5.4\% & 0.70 & 45.2\%±4.6\% & 1.24 & 42.8\%±7.8\% & 0.71 & 47.9\%±5.2\% & 1.23 \\
\midrule
\multicolumn{11}{l}{\textbf{Offline-LD}} \\
\midrule
mQRDQN-$D_{\text{expert}}$ & 25.5\%±1.4\%$^*$ & 0.31 & 29\%±0.3\%$^*$ & 0.63 & 35.2\%±4.2\% & 1.19 & 32.8\%±4\% & 0.63 & 37.8\%±4.1\% & 1.20 \\
d-mSAC-$D_{\text{expert}}$ & \textbf{23.9\%±2.8\%}$^*$ & 0.35 & 28.6\%±3.1\%$^*$ & 0.71 & 34.0\%±4.4\% & 1.27 & \textbf{31.6\%±4.2\%}$^*$ & 0.70 & 37.6\%±3.2\% & 1.26 \\
mQRDQN-$D_{\text{noisy}}$ & 25.2\%±0.7\%$^*$ & 0.32 & 28.9\%±1.4\%$^*$ & 0.64 & 33.5\%±2\% & 1.18 & 32.8\%±2.5\% & 0.64 & \textbf{35.8\%±2.4\%} & 1.19 \\
d-mSAC-$D_{\text{noisy}}$ & 25.4\%±2\%$^*$ & 0.36 & \textbf{28.0\%±3.9\%}$^*$ & 0.69 & \textbf{32.8\%±4\%} & 1.28 & 33.4\%±3.6\% & 0.70 & 41.2\%±8.9\% & 1.27 \\
\midrule
\textcolor{black}{Opt. Rate (\%)} & \multicolumn{2}{c}{\textcolor{black}{100\%}} & \multicolumn{2}{c}{\textcolor{black}{30\%}} & \multicolumn{2}{c}{\textcolor{black}{0\%}} & \multicolumn{2}{c}{\textcolor{black}{0\%}} & \multicolumn{2}{c}{\textcolor{black}{10\%}} \\
\bottomrule
\end{tabular}

\label{tab:benchmark_results}
\end{sidewaystable}
\begin{figure*}[ht]
    \centering
    \includegraphics[width=1\linewidth]{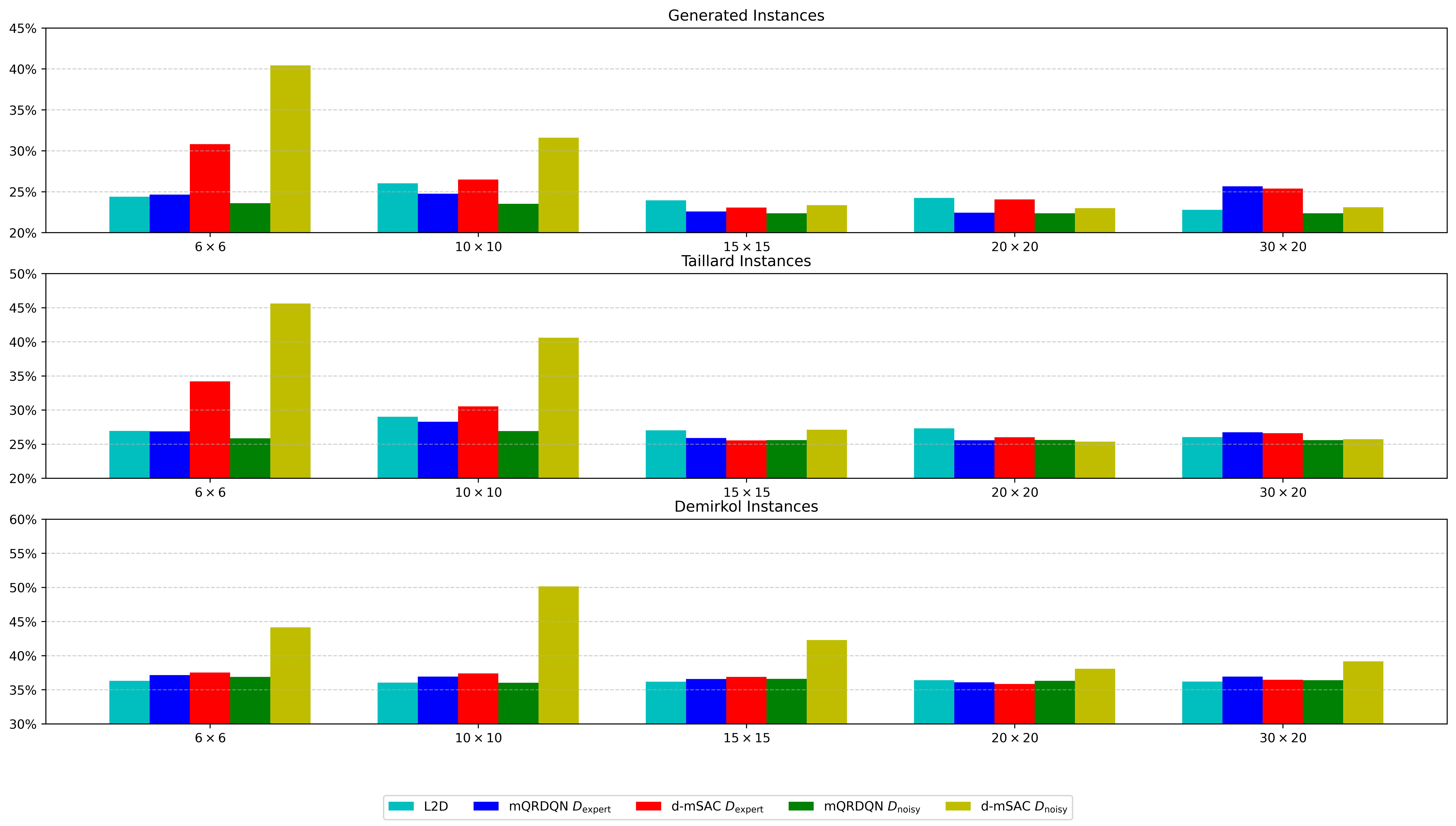}

    \caption{The figure shows the generalization of each trained approach by showing the average gap of the whole benchmark. The x-axis represents the trained instance size. The bars are the average gap. \textcolor{black}{Our results indicate that mQRDQN is able to generalize well to larger instance sizes, whereas d-mSAC does not, if trained on smaller instance sizes.}}
    \label{fig:benchmark_comparison}
\end{figure*}

\begin{table*}
\color{black}
\caption{Results for L2D, mQRDQN, d-mSAC with different training sizes, when evaluated on generated instances of sizes $50 \times 20$ and $100 \times 20$ instances. The results shows that mQRDQN is able to generalize well to larger instances ($50 \times 20$ and $100 \times 20$), when trained on smaller instances ($6 \times 6$ and $10 \times 10$), whereas d-mSAC does not. \textbf{Gap}: Difference from CP solution. \textbf{Time}: Runtime (s). \textbf{Bold}: Best RL approach. *: Offline-LD significantly outperforms L2D (t-test, $p=0.05$). Opt. Rate (\%): Percentage of instances solved optimally by CP.}
\label{tab:gen_big_general} 
\centering

\begin{tabular}{lcccc} 
\toprule
\multirow{2}{*}{Method} & \multicolumn{2}{c}{$50 \times 20$} & \multicolumn{2}{c}{$100 \times 20$} \\
\cmidrule(lr){2-3} \cmidrule(lr){4-5} 
& Gap & Time & Gap & Time \\
\midrule
\multicolumn{5}{l}{\textbf{Trained Size:} $6 \times 6$} \\
\midrule
L2D                   & 22.7\%$\pm$3.9\% & 2.84 & 9.4\%$\pm$1.9\% & 12.24 \\
mQRDQN-$D_{\text{expert}}$ & 22.5\%$\pm$2.2\%     & 2.68 & 8.8\%$\pm$1.2\%${}^{*}$ & 12.76 \\
d-mSAC-$D_{\text{expert}}$ & 27.4\%$\pm$4.6\%     & 2.81 & 10.7\%$\pm$2.2\%      & 12.52 \\
mQRDQN-$D_{\text{noisy}}$  & 21.7\%$\pm$2.3\%${}^{*}$ & 2.63 & 8.8\%$\pm$1.2\%${}^{*}$ & 12.61 \\
d-mSAC-$D_{\text{noisy}}$  & 41.3\%$\pm$10.0\%    & 2.9  & 18.6\%$\pm$7.2\%      & 11.92 \\
\midrule
\multicolumn{5}{l}{\textbf{Trained Size:} $10 \times 10$} \\
\midrule
L2D                   & 24.0\%$\pm$4.5\% & 2.7  & 9.7\%$\pm$2.0\% & 11.91 \\
mQRDQN-$D_{\text{expert}}$ & 23.2\%$\pm$2.6\%${}^{*}$ & 2.79 & 9.1\%$\pm$1.4\%${}^{*}$ & 12.52 \\
d-mSAC-$D_{\text{expert}}$ & 26.8\%$\pm$4.4\%     & 2.96 & 10.6\%$\pm$2.1\%      & 12.4  \\
mQRDQN-$D_{\text{noisy}}$  & 22.8\%$\pm$3.0\%${}^{*}$ & 2.75 & 8.9\%$\pm$1.7\%${}^{*}$ & 11.98 \\
d-mSAC-$D_{\text{noisy}}$  & 42.8\%$\pm$10.9\%    & 2.89 & 24.2\%$\pm$7.1\%      & 11.84 \\
\midrule
\multicolumn{5}{l}{\textbf{Trained Size:} $15 \times 15$} \\
\midrule
L2D                   & 22.5\%$\pm$3.3\% & 2.74 & 9.1\%$\pm$1.5\% & 12.07 \\
mQRDQN-$D_{\text{expert}}$ & 21.6\%$\pm$2.9\%${}^{*}$ & 2.68 & 8.8\%$\pm$1.4\%${}^{*}$ & 12.89 \\
d-mSAC-$D_{\text{expert}}$ & 21.7\%$\pm$2.9\%${}^{*}$ & 2.82 & 9.2\%$\pm$1.7\%       & 12.79 \\
mQRDQN-$D_{\text{noisy}}$  & 21.2\%$\pm$1.5\%${}^{*}$ & 2.67 & \textbf{8.5\%$\pm$1.0\%}${}^{*}$ & 12.42 \\
d-mSAC-$D_{\text{noisy}}$  & 25.6\%$\pm$7.8\%     & 2.9  & 13.5\%$\pm$7.4\%      & 12.32 \\
\midrule
\multicolumn{5}{l}{\textbf{Trained size:} $20 \times 20$} \\
\midrule
L2D                   & 23.0\%$\pm$4.3\% & 2.77 & 9.5\%$\pm$2.0\% & 12.2  \\
mQRDQN-$D_{\text{expert}}$ & 21.3\%$\pm$1.8\%${}^{*}$ & 2.63 & 8.6\%$\pm$1.1\%${}^{*}$ & 12.84 \\
d-mSAC-$D_{\text{expert}}$ & 21.7\%$\pm$2.3\%${}^{*}$ & 2.96 & 8.9\%$\pm$1.3\%${}^{*}$ & 13.32 \\
mQRDQN-$D_{\text{noisy}}$  & 21.2\%$\pm$1.8\%${}^{*}$ & 2.74 & 8.5\%$\pm$1.1\%${}^{*}$ & 12.68 \\
d-mSAC-$D_{\text{noisy}}$  & 21.5\%$\pm$2.4\%${}^{*}$ & 2.9  & 9.6\%$\pm$2.3\%       & 13.1  \\
\midrule
\multicolumn{5}{l}{\textbf{Trained Size:} $30 \times 20$} \\
\midrule
L2D                   & 21.5\%$\pm$1.7\%     & 2.88 & 8.8\%$\pm$0.9\% & 12.93 \\
mQRDQN-$D_{\text{expert}}$ & 22.8\%$\pm$3.6\%     & 2.8  & 9.4\%$\pm$1.8\%       & 12.81 \\
d-mSAC-$D_{\text{expert}}$ & 22.1\%$\pm$2.9\%     & 2.89 & 9.0\%$\pm$1.5\%       & 12.14 \\
mQRDQN-$D_{\text{noisy}}$  & \textbf{21.2\%$\pm$1.4\%}${}^{*}$ & 2.7  & 8.7\%$\pm$0.7\%       & 12.74 \\
d-mSAC-$D_{\text{noisy}}$  & 21.7\%$\pm$2.4\%     & 2.88 & 8.7\%$\pm$1.3\%       & 12.41 \\
\midrule
Opt. Rate (\%) & \multicolumn{2}{c}{89\%} & \multicolumn{2}{c}{100\%} \\
\bottomrule
\end{tabular}
\end{table*}



The results in Table~\ref{tab:generated_instances} show that with the generated instances, Offline-LD demonstrates superior performance, significantly outperforming baselines, BC and L2D, in 4 out of 5 instance sizes (denoted with *), while showing comparable performance for the remaining sizes. \textcolor{black}{These results highlight the robustness of Offline-LD across different problem setting and sizes. Moreover, Tables~\ref{tab:generated_instances} and~\ref{tab:benchmark_results} show that PDRs perform significantly worse compared to the learning-based methods, such as BC, L2D and Offline-LD.} Offline-LD demonstrates \textcolor{black}{(Table~\ref{tab:benchmark_results})} superior performance in 3 out of 5 instance sizes for the benchmark instances, \textcolor{black}{although} its performance diminishes for $30 \times 20$ instances and the Demirkol benchmark set, which feature a broader range of processing times compared to generated and Taillard instances.

Surprisingly, Behavioral Cloning (BC) consistently ranks as the worst performing approach \textcolor{black}{among learning-based methods} across different instance sizes and types. This outcome is particularly unexpected given BC's success in robotics applications with expert datasets~\citep{offline_human,implicit_bc}. This poor performance may stem from the distributional shift caused by the supervised training on different instances than the ones used for evaluation, compounded by the limited training dataset of only 100 instances~\citep{bc_offline_rl}.


Comparing the results of \textcolor{black}{Offline-LD trained with} noisy-expert ($D_{\text{noisy}}$) and expert ($D_{\text{expert}}$) datasets in Tables \ref{tab:generated_instances} and \ref{tab:benchmark_results}, we observe that training with a noisy-expert dataset generally yields better-performing and more stable policies across most instance sizes.  The most probable reason for this is that JSSP contains many ``critical states``, where a specific single action needs to be taken to get an optimal makespan. \textcolor{black}{By including counterfactual information (suboptimal actions and their consequences), the noisy dataset likely improves learning, particularly in critical states where specific actions are crucial~\citep{bc_offline_rl}.} 

\paragraph{Generalization to larger instances.} \textcolor{black}{In addition, we test whether the learned policies of our proposed offline-LD methods can perform well on larger test instances.} In Fig.~\ref{fig:benchmark_comparison}, we compare the performance of Offline-LD and L2D for each trained instance set with all instances in the corresponding benchmark\footnote{The full results can be found in Appendix \ref{app:results}.}. Offline-LD with mQRDQN with $D_{\text{noisy}}$ performed the best on both the generated and Taillard instances. \textcolor{black}{In Table~\ref{tab:gen_big_general}, we see that mQRDQN is able to generalize well to larger sizes, even if trained on smaller size. For example, Table~\ref{tab:gen_big_general} shows that if trained on $6\times6$ the gap with $D_\text{noisy}$ is $21.7\% \pm 2.3\%$ for the $50 \times 20$ instances, while with $30 \times 20$, this is only slightly better with $21.2\% \pm 1.4\%$. However, d-mSAC is not able to generalize well, especially with $D_{\text{noisy}}$. Table~\ref{tab:gen_big_general} shows that on the $100\times 20$ instance set, it scored $18.6\% \pm 7.2\%$ if trained on $6\times 6$, while the performance increases significantly if trained on $30\times 20$, where it achieved a gap of $8.7\%\pm 1.3\%$.}

\begin{wrapfigure}{r}{0.5\textwidth}
  \centering
    \includegraphics[width=0.48\textwidth]{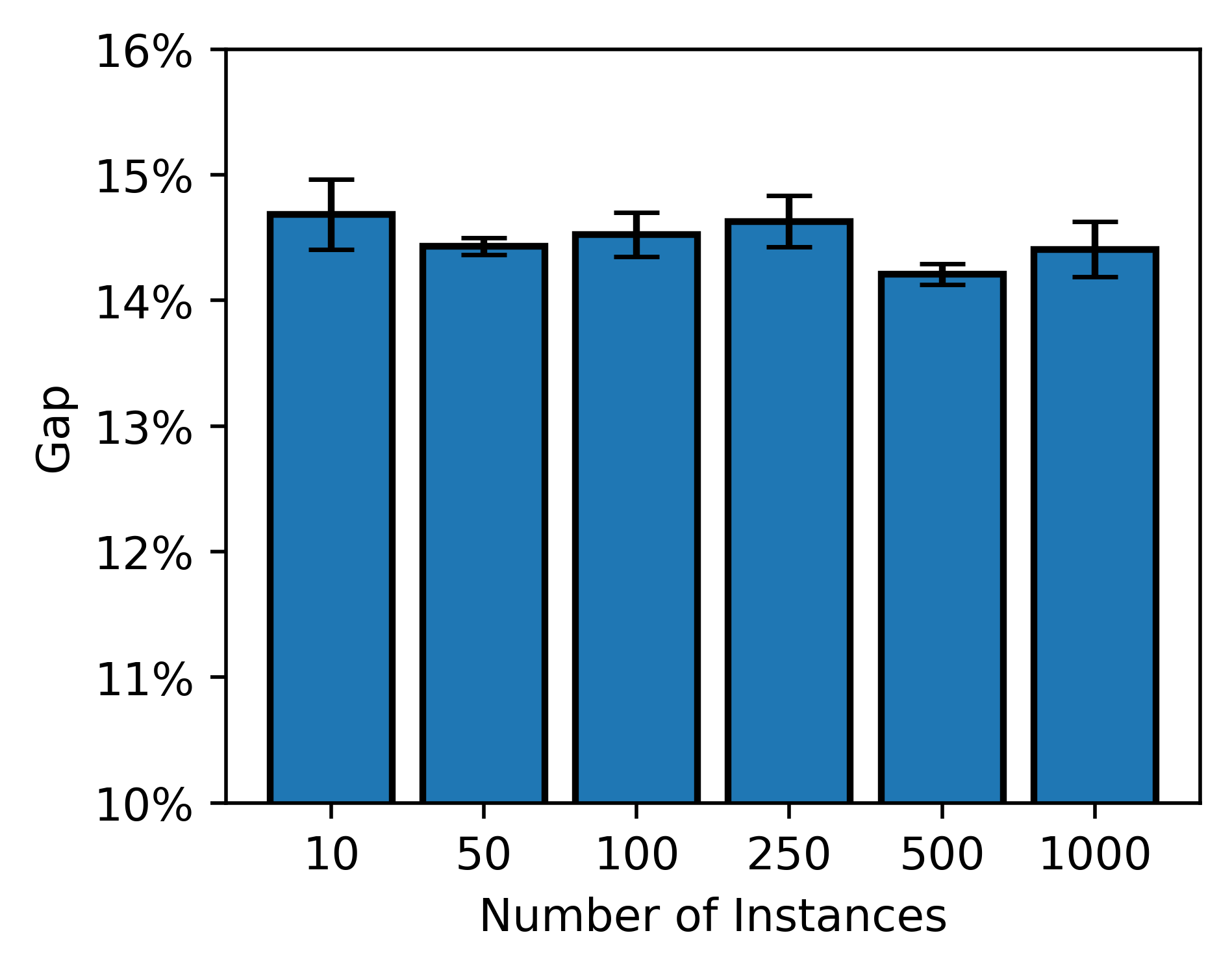}
  
  \caption{The results of different training dataset sizes with Offline-LD. The x-axis shows the number of instances in the training dataset used, and the y-axis shows the results on the $6 \times 6$ evaluation set.}
  \label{fig:dataset_size}
  \vspace{-30pt}
\end{wrapfigure}
\textcolor{black}{
\paragraph{Larger datasets.}
The results of Fig.~\ref{fig:benchmark_comparison} and Table~\ref{tab:gen_big_general} suggest that we might achieve higher performance by increasing the dataset size. This is trivial to evaluate for $6 \times 6$ instance size, as  CP required on average only 0.03 seconds to find an optimal solution of a given instance. To examine the effect of different dataset sizes, we both increased and decreased the number of instances from our original training set of 100 instances with $6 \times 6$ size.
Fig.~\ref{fig:dataset_size} shows the impact of varying the number of instances in a training dataset. For these experiments, we trained on mQRDQN on the $6 \times 6$ expert dataset, evaluated on the $6 \times 6$ evaluation set. When we applied a Wilcoxon signed rank test, we found no statistically significant difference between results. For example, the largest difference is between using 10 instances (Gap 14.68\% $\pm$ 0.28\%) and 500 instances (Gap 14.21\% $\pm$ 0.08\%), for which the Wilcoxon test showed $p\approx 0.22$, which means $p> 0.05$ and therefore does not reject the null hypothesis. This indicates that increasing the dataset size does not lead to a statistically significant improvement in performance, and that as few as 10 instances suffice to achieve competitive results.
}

\textcolor{black}{
These results contrasts prior research in offline RL~\citep{agarwal2020optimistic}, which showed that decreasing the dataset size has a significant effect on performance. However, \citet{agarwal2020optimistic} did their experiments for the Atari benchmark, which has significantly different properties and objectives than JSSP. Moreover, these and previous results indicate that Offline-LD can outperform online RL methods, such as L2D~\citep{l2d}, with significantly fewer training data. Offline-LD only required 100 instances, with one trajectory for each instance, while L2D required 10,000 instances, for which it collects four trajectories. This indicates the sample efficiency of Offline-LD in comparison to online RL methods.
}

\subsection{Ablation Study}
\begin{figure}[t]
    \centering
    \includegraphics[width=0.9\linewidth]{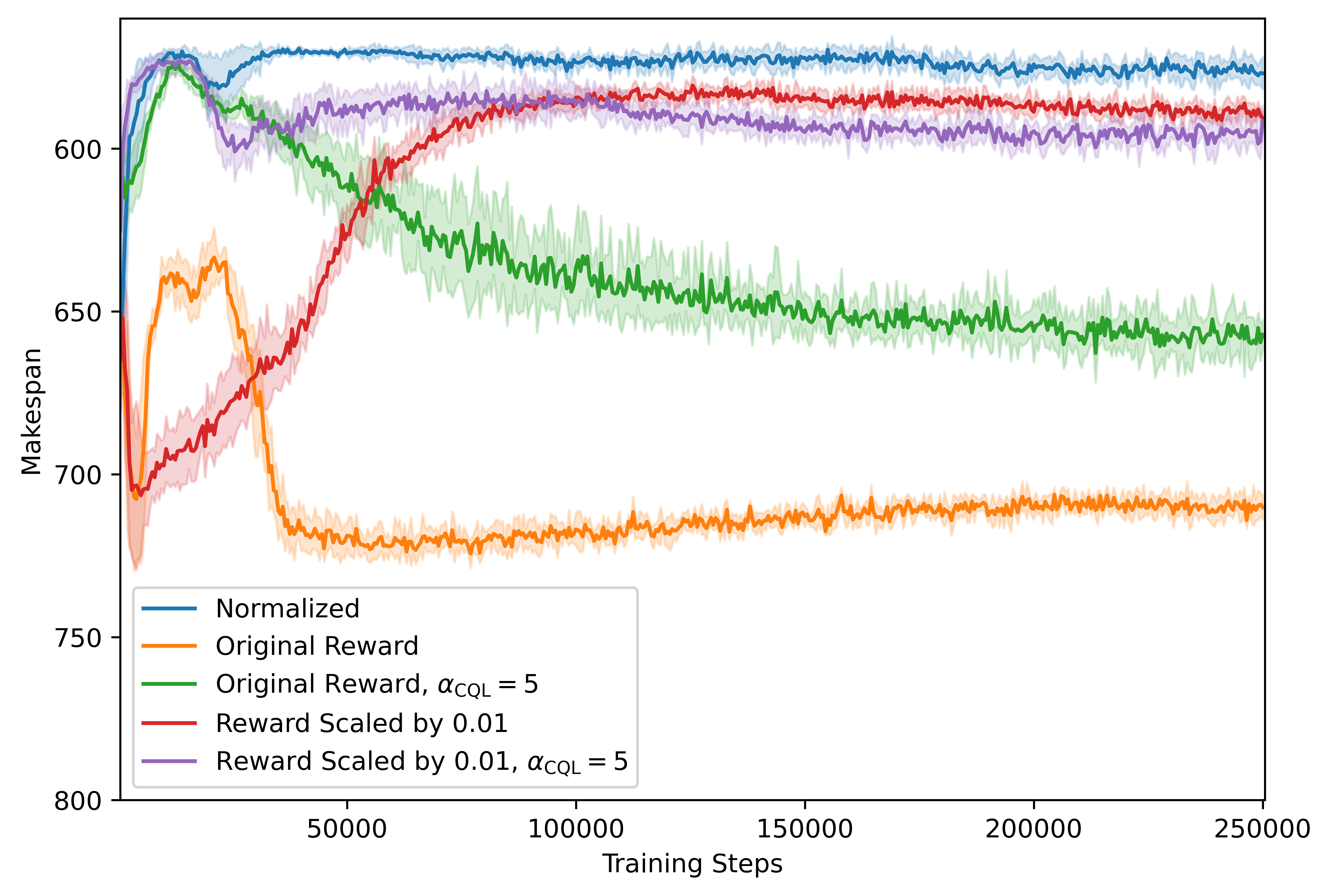}
    \caption{This figure shows the results of different training runs \textcolor{black}{of mQRDQN} for $6 \times 6$ with the expert dataset $D_{\text{expert}}$, whereby the line is the mean and the shaded area is the standard deviation. The results show that normalizing the reward stabilizes training and improves performance. \textcolor{black}{We clearly see that normalizing the reward results in better performance. If we neither scale or normalize the reward, the learned policy is highly unstable.}}
    \label{fig:exp_normalization}
\end{figure}
We conducted an ablation study to evaluate the effect of reward normalization. Our ablation experiments used Offline-LD with mQRDQN and trained on the $6\times 6$ expert dataset. We trained with the unnormalized dataset in four different configurations: standard, which uses the same hyperparameters as the normalized dataset; reward scaled by 0.01; setting $\alpha_\text{CQL}=5$, instead of $\alpha_\text{CQL}=1$; and combining both. We trained for 250,000 training steps instead of the normal 50,000 to test if other configurations would need more training steps. Each configuration is trained with five different seeds.

Fig. \ref{fig:exp_normalization} shows that normalizing the reward leads to the best results and training stability. The worst performance is given if the reward is neither normalized nor scaled. The most probable cause is that the reward function of \citet{l2d} has a relatively sparse and large output scale. This can result in inefficient and unstable training \citep{reward_scalar}. \textcolor{black}{This effect was not observed in \cite{l2d}, as PPO was used,} whereby before the policy is updated for $K$ epochs, the rewards gathered for these updates are normalized beforehand.

Fig. \ref{fig:exp_normalization} also shows that if $\alpha_{CQL}=5$, both scaled and unscaled policies perform roughly similarly to when rewards are normalized with the makespan until training step 25,000, whereafter they both converge to a worse makespan. The survival instinct of offline RL methods, such as CQL, most likely causes this result, whereby there exists a certain tolerance to faulty rewards if there is enough pessimism \citep{survival_offline_rl}.
\section{Conclusion}
This paper introduced Offline-LD, the first fully end-to-end offline RL approach for JSSP. Our results show that when trained on a small dataset of 100 instances only, we achieve either similar or better results than L2D, an online RL method. Moreover, the results improved when noisy training data was used as it contained counterfactual information. It also shows the robustness of offline RL in that it does not need optimal data for JSSP. 

In this paper, we present the first promising offline RL approach for JSSP and for solving combinatorial optimization problems. However, there are many potential improvements to the proposed offline RL approach. For example, our method builds up on a existing neural architecture and state-space representation, designed for online RL. For future research, we will design a network architecture more suitable for offline RL, and further explore how to assign rewards on collected data, specified for offline RL settings. \textcolor{black}{Furthermore, in this paper we relied on expert datasets generated through CP. While our results demonstrate competitive performance, even when training on as few as 10 instances, we recognize this as a limitation, as expert-generated data may not always be available in real-world settings. To address this, future work could explore training using datasets generated by alternative methods, including simple priority dispatching rules (PDRs) such as MOR, SPT, and MWKR. Such research could help determine whether performance depends strongly on training with high-quality solutions (e.g., from CP solvers), or whether comparable performance can be attained using data derived from simpler heuristics.}

Moreover, we believe that offline RL is also a promising research direction for other combinatorial optimization problems (COP) since many real-world COP are impractical to simulate due to natural inputs~\citep{cappart_gnn_survey}, such as weather forecasts for vehicle routing problems~\citep{weather_vrp}, or due to unexpected disturbances or machine breakdowns for scheduling problems~\citep{jssp_digital_twin, LI2023102443}. \textcolor{black}{This limits online RL for real-world COP problems, since it require accurate simulations of those problems to train~\citet{jssp_digital_twin, offline_rl_survey}.} In contrast, for many of these COPs, such as JSSP, there are real-world datasets, \textcolor{black}{that contains events like machine breakdowns,} which could be used to train offline RL methods. In addition, these datasets could be used to create benchmark datasets for offline RL.


\backmatter

\bmhead{Supplementary information}

Our supplementary material, such as our code for training and dataset generation, can be found in our GitHub repository (https://github.com/jesserem/Offline-LD)

\bmhead{Acknowledgements}
This research is partially supported by funding from the Dutch Research Council (NWO) and the municipality of Amsterdam under the Urbiquay program of the STABILITY (Grant NWA.1431.20.004). 

\section*{Declarations}
\bmhead{Conflict of Interests}
The authors declare that they have no conflict of interest.
\bmhead{Author Contributions}
All authors contributed to the conception of the work. J. van Remmerden wrote the main manuscript text. All authors reviewed and revised the manuscript.




\bibliography{sn-bibliography.bib}
\begin{appendices}

\section{Additional Results}\label{app:results}
\textcolor{black}{This appendix shows all the results for the generated instance (Tables \ref{tab:small_gen_trained}, \ref{tab:large_gen_trained}, \ref{tab:gen_small_small}, \ref{tab:gen_small_large} and \ref{tab:gen_big}), Taillard (Tables \ref{tab:tai_trained}, \ref{tab:tai_other_small} and \ref{tab:tai_other_large}), and Demirkol (Tables \ref{tab:dmu_trained}, \ref{tab:dmu_small} and \ref{tab:dmu_big}). These results show how Offline-LD performed on each instance set, with the trained instance size. We used these results to create Fig~\ref{fig:benchmark_comparison}.}

\textcolor{black}{We noticed that d-mSAC did not generalize well to larger sizes, when trained in smaller sizes such as $6 \times 6$ and $10 \times 10$, while mQRDQN did not encounter this issue. Moreover, both methods could generalize well when trained on larger sizes, to instance sizes of smaller sizes. However, we again notice that mQRDQN could generalize better to smaller sizes. The most probable cause for this is that mQRDQN uses quantile regression~\cite{qrdqn} to approximate the Q values, which could improve generalization to different sizes.}

\begin{table*}[ht]
\caption{The results for the generated instance for the following sizes: $6 \times 6$, $10 \times 10$, $15 \times 15$. \textbf{Gap} is the difference between the obtained result and the found CP solution. \textbf{Time} is the runtime in seconds. The \textbf{bold} result is the best RL approach. ${}^*$ signifies that an offline RL approach is a significant improvement compared to L2D of the same trained instance size, according to an independent t-test ($p=0.05$). The Opt. Rate (\%) is the percentage of the solutions that CP found an optimal result.}
\centering
\fontsize{6pt}{6pt}\selectfont
\setlength{\tabcolsep}{2mm}
\begin{tabular}{|ccccccc|}
\hline
\multicolumn{1}{|c|}{} & \multicolumn{2}{|c|}{$6 \times 6$}& \multicolumn{2}{|c|}{$10 \times 10$}& \multicolumn{2}{|c|}{$15 \times 15$}\\
\multicolumn{1}{|c|}{} & Gap & \multicolumn{1}{c|}{Time}& Gap & \multicolumn{1}{c|}{Time}& Gap & \multicolumn{1}{c|}{Time} \\ \hline 
\multicolumn{7}{|c|}{\textbf{PDR}} \\ \hline\multicolumn{1}{|c|}{SPT} & 41.5\% & \multicolumn{1}{c|}{0.01}& 51.6\% & \multicolumn{1}{c|}{0.01}& 58.0\% & \multicolumn{1}{c|}{0.02} \\ 
\multicolumn{1}{|c|}{MOR} & \textcolor{black}{20.7}\% & \multicolumn{1}{c|}{0.01}& \textcolor{black}{33}\% & \multicolumn{1}{c|}{0.01}& \textcolor{black}{41.4}\% & \multicolumn{1}{c|}{0.02} \\ 
\multicolumn{1}{|c|}{MWKR} & \textcolor{black}{32.4}\% & \multicolumn{1}{c|}{0.01}& \textcolor{black}{47.5}\% & \multicolumn{1}{c|}{0.01}& \textcolor{black}{57.6}\% & \multicolumn{1}{c|}{0.02} \\ \hline 
\multicolumn{7}{|c|}{\textbf{Baselines} $6 \times 6$ } \\ \hline 
\multicolumn{1}{|c|}{L2D} & 15.8\%± 6.5\% & \multicolumn{1}{c|}{0.05}& 23.7\%± 7.8\% & \multicolumn{1}{c|}{0.15}& 28.4\%± 7.3\% & \multicolumn{1}{c|}{0.36}\\ 
\multicolumn{1}{|c|}{BC-$D_{\text{expert}}$} & 30.9\%± 7.0\% & \multicolumn{1}{c|}{0.05}& 37.1\%± 10.5\% & \multicolumn{1}{c|}{0.14}& 41.9\%± 13.7\% & \multicolumn{1}{c|}{0.35} \\ \hline 
\multicolumn{7}{|c|}{\textbf{Offline-LD} $6 \times 6$ } \\ \hline 
\multicolumn{1}{|c|}{mQRDQN-$D_{\text{expert}}$}& 14.5\%± 1.0\%${}^{*}$ & \multicolumn{1}{c|}{0.04}& 24.9\%± 4.1\% & \multicolumn{1}{c|}{0.13}& 28.9\%± 3.7\% & \multicolumn{1}{c|}{0.32} \\
\multicolumn{1}{|c|}{d-mSAC-$D_{\text{expert}}$}& 14.5\%± 2.6\%${}^{*}$ & \multicolumn{1}{c|}{0.05}& 28.0\%± 8.5\% & \multicolumn{1}{c|}{0.14}& 39.8\%± 8.5\% & \multicolumn{1}{c|}{0.35} \\ 
\multicolumn{1}{|c|}{mQRDQN-$D_{\text{noisy}}$}& 14.5\%± 1.4\%${}^{*}$ & \multicolumn{1}{c|}{0.04}& 22.7\%± 2.4\% & \multicolumn{1}{c|}{0.12}& 27.8\%± 2.5\% & \multicolumn{1}{c|}{0.31} \\ 
\multicolumn{1}{|c|}{d-mSAC-$D_{\text{noisy}}$}& 14.3\%± 1.5\%${}^{*}$ & \multicolumn{1}{c|}{0.05}& 29.3\%± 11.1\% & \multicolumn{1}{c|}{0.14}& 50.5\%± 9.7\% & \multicolumn{1}{c|}{0.35} \\ \hline 
\multicolumn{7}{|c|}{\textbf{Baselines} $10 \times 10$ } \\ \hline 
\multicolumn{1}{|c|}{L2D} & 17.8\%± 7.8\% & \multicolumn{1}{c|}{0.05}& 25.3\%± 8.5\% & \multicolumn{1}{c|}{0.14}& 30.6\%± 7.8\% & \multicolumn{1}{c|}{0.34} \\ 
\multicolumn{1}{|c|}{BC-$D_{\text{expert}}$} & 33.1\%± 10.3\% & \multicolumn{1}{c|}{0.05}& 33.2\%± 8.0\% & \multicolumn{1}{c|}{0.14}& 39.5\%± 6.5\% & \multicolumn{1}{c|}{0.34} \\ \hline 
\multicolumn{7}{|c|}{\textbf{Offline-LD} $10 \times 10$ } \\ \hline 
\multicolumn{1}{|c|}{mQRDQN-$D_{\text{expert}}$}& 14.7\%± 1.0\%${}^{*}$ & \multicolumn{1}{c|}{0.04}& 20.6\%± 1.8\%${}^{*}$ & \multicolumn{1}{c|}{0.12}& 30.9\%± 4.7\% & \multicolumn{1}{c|}{0.31} \\ 
\multicolumn{1}{|c|}{d-mSAC-$D_{\text{expert}}$}& 14.8\%± 2.0\%${}^{*}$ & \multicolumn{1}{c|}{0.05}& 21.0\%± 2.5\%${}^{*}$ & \multicolumn{1}{c|}{0.14}& 29.8\%± 6.7\% & \multicolumn{1}{c|}{0.35} \\ 
\multicolumn{1}{|c|}{mQRDQN-$D_{\text{noisy}}$}& 14.4\%± 0.4\%${}^{*}$ & \multicolumn{1}{c|}{0.04}& 20.4\%± 0.6\%${}^{*}$ & \multicolumn{1}{c|}{0.12}& 27.5\%± 5.4\%${}^{*}$ & \multicolumn{1}{c|}{0.31} \\ 
\multicolumn{1}{|c|}{d-mSAC-$D_{\text{noisy}}$}& 14.5\%± 0.2\%${}^{*}$ & \multicolumn{1}{c|}{0.05}& 20.7\%± 0.9\%${}^{*}$ & \multicolumn{1}{c|}{0.14}& 27.3\%± 5.7\%${}^{*}$ & \multicolumn{1}{c|}{0.35} \\ \hline 
\multicolumn{7}{|c|}{\textbf{Baselines} $15 \times 15$ } \\ \hline 
\multicolumn{1}{|c|}{L2D} & 15.5\%± 5.6\% & \multicolumn{1}{c|}{0.05}& 22.3\%± 5.2\% & \multicolumn{1}{c|}{0.15}& 28.1\%± 5.3\% & \multicolumn{1}{c|}{0.35} \\ 
\multicolumn{1}{|c|}{BC-$D_{\text{expert}}$} & 36.2\%± 7.9\% & \multicolumn{1}{c|}{0.05}& 35.3\%± 7.0\% & \multicolumn{1}{c|}{0.14}& 37.1\%± 6.7\% & \multicolumn{1}{c|}{0.35} \\ \hline 
\multicolumn{7}{|c|}{\textbf{Offline-LD} $15 \times 15$ } \\ \hline 
\multicolumn{1}{|c|}{mQRDQN-$D_{\text{expert}}$}& 15.2\%± 2.6\% & \multicolumn{1}{c|}{0.04}& 20.5\%± 0.2\%${}^{*}$ & \multicolumn{1}{c|}{0.13}& 26.0\%± 1.4\%${}^{*}$ & \multicolumn{1}{c|}{0.31} \\ 
\multicolumn{1}{|c|}{d-mSAC-$D_{\text{expert}}$}& 16.9\%± 4.9\% & \multicolumn{1}{c|}{0.05}& 21.9\%± 3.6\% & \multicolumn{1}{c|}{0.14}& 26.3\%± 3.3\%${}^{*}$ & \multicolumn{1}{c|}{0.35} \\ 
\multicolumn{1}{|c|}{mQRDQN-$D_{\text{noisy}}$}& \textbf{14.3\%± 0.0}\%${}^{*}$ & \multicolumn{1}{c|}{0.04}& \textbf{20.4\%± 0.1\%}${}^{*}$ & \multicolumn{1}{c|}{0.13}& \textbf{26.0\%± 0.5\%}${}^{*}$ & \multicolumn{1}{c|}{0.32} \\ 
\multicolumn{1}{|c|}{d-mSAC-$D_{\text{noisy}}$}& 14.7\%± 1.3\% & \multicolumn{1}{c|}{0.05}& 20.8\%± 0.8\%${}^{*}$ & \multicolumn{1}{c|}{0.14}& 26.1\%± 1.8\%${}^{*}$ & \multicolumn{1}{c|}{0.36} \\ \hline 
\multicolumn{7}{|c|}{\textbf{Baselines} $20 \times 20$ } \\ \hline 
\multicolumn{1}{|c|}{L2D} & 15.2\%± 4.5\% & \multicolumn{1}{c|}{0.05}& 22.5\%± 5.7\% & \multicolumn{1}{c|}{0.15}& 28.2\%± 6.4\% & \multicolumn{1}{c|}{0.35} \\ 
\multicolumn{1}{|c|}{BC-$D_{\text{expert}}$} & 39.5\%± 8.6\% & \multicolumn{1}{c|}{0.05}& 39.6\%± 7.7\% & \multicolumn{1}{c|}{0.14}& 39.0\%± 6.7\% & \multicolumn{1}{c|}{0.35} \\ \hline 
\multicolumn{7}{|c|}{\textbf{Offline-LD} $20 \times 20$ } \\ \hline 
\multicolumn{1}{|c|}{mQRDQN-$D_{\text{expert}}$}& 14.6\%± 0.8\% & \multicolumn{1}{c|}{0.04}& 20.7\%± 0.8\%${}^{*}$ & \multicolumn{1}{c|}{0.12}& 25.9\%± 0.8\%${}^{*}$ & \multicolumn{1}{c|}{0.31} \\ 
\multicolumn{1}{|c|}{d-mSAC-$D_{\text{expert}}$}& 19.6\%± 6.0\% & \multicolumn{1}{c|}{0.05}& 23.7\%± 4.7\% & \multicolumn{1}{c|}{0.14}& 27.6\%± 4.0\% & \multicolumn{1}{c|}{0.36} \\ 
\multicolumn{1}{|c|}{mQRDQN-$D_{\text{noisy}}$}& 14.4\%± 0.5\%${}^{*}$ & \multicolumn{1}{c|}{0.04}& 20.5\%± 0.6\%${}^{*}$ & \multicolumn{1}{c|}{0.13}& 26.1\%± 0.6\%${}^{*}$ & \multicolumn{1}{c|}{0.32}\\ 
\multicolumn{1}{|c|}{d-mSAC-$D_{\text{noisy}}$}& 16.2\%± 3.3\% & \multicolumn{1}{c|}{0.05}& 21.5\%± 2.5\% & \multicolumn{1}{c|}{0.14}& 26.5\%± 3.2\%${}^{*}$ & \multicolumn{1}{c|}{0.35} \\ \hline 
\multicolumn{7}{|c|}{\textbf{Baselines} $30 \times 20$ } \\ \hline 
\multicolumn{1}{|c|}{L2D} & 14.6\%± 2.1\% & \multicolumn{1}{c|}{0.05}& 21.2\%± 5.0\% & \multicolumn{1}{c|}{0.15}& 26.6\%± 4.5\% & \multicolumn{1}{c|}{0.35}\\ 
\multicolumn{1}{|c|}{BC-$D_{\text{expert}}$} & 38.5\%± 9.6\% & \multicolumn{1}{c|}{0.05}& 43.8\%± 8.9\% & \multicolumn{1}{c|}{0.14}& 43.4\%± 7.5\% & \multicolumn{1}{c|}{0.35}\\ \hline 
\multicolumn{7}{|c|}{\textbf{Offline-LD} $30 \times 20$ } \\ \hline 
\multicolumn{1}{|c|}{mQRDQN-$D_{\text{expert}}$}& 26.6\%± 12.5\% & \multicolumn{1}{c|}{0.05}& 27.6\%± 10.2\% & \multicolumn{1}{c|}{0.13}& 28.3\%± 6.7\% & \multicolumn{1}{c|}{0.32} \\ 
\multicolumn{1}{|c|}{d-mSAC-$D_{\text{expert}}$}& 25.1\%± 13.7\% & \multicolumn{1}{c|}{0.05}& 25.5\%± 8.6\% & \multicolumn{1}{c|}{0.14}& 28.7\%± 6.6\% & \multicolumn{1}{c|}{0.35} \\ 
\multicolumn{1}{|c|}{mQRDQN-$D_{\text{noisy}}$}& 14.4\%± 0.9\% & \multicolumn{1}{c|}{0.04}& 20.4\%± 1.0\%${}^{*}$ & \multicolumn{1}{c|}{0.13}& 26.1\%± 1.9\% & \multicolumn{1}{c|}{0.31} \\ 
\multicolumn{1}{|c|}{d-mSAC-$D_{\text{noisy}}$}& 18.3\%± 6.9\% & \multicolumn{1}{c|}{0.05}& 22.2\%± 4.7\% & \multicolumn{1}{c|}{0.14}& 26.3\%± 3.9\% & \multicolumn{1}{c|}{0.36} \\ \hline 
\multicolumn{1}{|c|}{Opt. Rate (\%)} & \multicolumn{2}{c|}{100\%} & \multicolumn{2}{c|}{100\%} & \multicolumn{2}{c|}{100\%}  \\ \hline 
\end{tabular}

\label{tab:small_gen_trained}
\end{table*}

\begin{table*}[ht]
\caption{The results for the generated instance for the following sizes: $20 \times 20$, and $30 \times 20$. \textbf{Gap} is the difference between the obtained result and the found CP solution. \textbf{Time} is the runtime in seconds. The \textbf{bold} result is the best RL approach. ${}^*$ signifies that an offline RL approach is a significant improvement compared to L2D of the same trained instance size, according to an independent t-test ($p=0.05$). The Opt. Rate (\%) is the percentage of the solutions that CP found an optimal result.}
\centering
\fontsize{6pt}{6pt}\selectfont
\setlength{\tabcolsep}{2mm}
\begin{tabular}{|ccccc|}
\hline
\multicolumn{1}{|c|}{} & \multicolumn{2}{|c|}{$20 \times 20$}& \multicolumn{2}{|c|}{$30 \times 20$} \\
\multicolumn{1}{|c|}{} & Gap & \multicolumn{1}{c|}{Time}& Gap & \multicolumn{1}{c|}{Time} \\ \hline 
\multicolumn{5}{|c|}{\textbf{PDR}} \\ \hline\multicolumn{1}{|c|}{SPT} & 63.5\% & \multicolumn{1}{c|}{0.05}& 67.1\% & \multicolumn{1}{c|}{0.09} \\ 
\multicolumn{1}{|c|}{MOR} & 28.2\% & \multicolumn{1}{c|}{0.05}& 28.8\% & \multicolumn{1}{c|}{0.09} \\ 
\multicolumn{1}{|c|}{MWKR} & 27.1\% & \multicolumn{1}{c|}{0.06}& 29.6\% & \multicolumn{1}{c|}{0.12} \\ \hline 
\multicolumn{5}{|c|}{\textbf{Baselines} $6 \times 6$ } \\ \hline 
\multicolumn{1}{|c|}{L2D} & 30.1\%± 7.2\% & \multicolumn{1}{c|}{0.71}& 31.0\%± 5.3\% & \multicolumn{1}{c|}{1.27} \\ 
\multicolumn{1}{|c|}{BC-$D_{\text{expert}}$} & 45.6\%± 16.7\% & \multicolumn{1}{c|}{0.68}& 46.4\%± 17.6\% & \multicolumn{1}{c|}{1.24} \\ \hline 
\multicolumn{5}{|c|}{\textbf{Offline-LD} $6 \times 6$ } \\ \hline 
\multicolumn{1}{|c|}{mQRDQN-$D_{\text{expert}}$}& 30.3\%± 3.1\% & \multicolumn{1}{c|}{0.63}& 31.0\%± 2.9\% & \multicolumn{1}{c|}{1.17} \\ 
\multicolumn{1}{|c|}{d-mSAC-$D_{\text{expert}}$}& 41.2\%± 8.9\% & \multicolumn{1}{c|}{0.68}& 39.8\%± 7.1\% & \multicolumn{1}{c|}{1.25} \\ 
\multicolumn{1}{|c|}{mQRDQN-$D_{\text{noisy}}$}& 28.7\%± 2.3\%${}^{*}$ & \multicolumn{1}{c|}{0.62}& 30.2\%± 3.0\% & \multicolumn{1}{c|}{1.15} \\ 
\multicolumn{1}{|c|}{d-mSAC-$D_{\text{noisy}}$}& 56.2\%± 9.5\% & \multicolumn{1}{c|}{0.69}& 56.2\%± 9.5\% & \multicolumn{1}{c|}{1.26} \\ \hline 
\multicolumn{5}{|c|}{\textbf{Baselines} $10 \times 10$ } \\ \hline 
\multicolumn{1}{|c|}{L2D} & 32.8\%± 8.2\% & \multicolumn{1}{c|}{0.67}& 32.7\%± 6.2\% & \multicolumn{1}{c|}{1.19} \\ 
\multicolumn{1}{|c|}{BC-$D_{\text{expert}}$} & 45.8\%± 6.8\% & \multicolumn{1}{c|}{0.68}& 46.4\%± 7.3\% & \multicolumn{1}{c|}{1.23} \\ \hline 
\multicolumn{5}{|c|}{\textbf{Offline-LD} $10 \times 10$ } \\ \hline 
\multicolumn{1}{|c|}{mQRDQN-$D_{\text{expert}}$}& 32.5\%± 4.6\% & \multicolumn{1}{c|}{0.63}& 32.4\%± 3.7\% & \multicolumn{1}{c|}{1.16} \\ 
\multicolumn{1}{|c|}{d-mSAC-$D_{\text{expert}}$} & 39.7\%± 8.3\% & \multicolumn{1}{c|}{0.7}& 38.8\%± 7.0\% & \multicolumn{1}{c|}{1.26} \\ 
\multicolumn{1}{|c|}{mQRDQN-$D_{\text{noisy}}$}& 30.7\%± 4.9\%${}^{*}$ & \multicolumn{1}{c|}{0.62}& 31.3\%± 4.0\%${}^{*}$ & \multicolumn{1}{c|}{1.16} \\ 
\multicolumn{1}{|c|}{d-mSAC-$D_{\text{noisy}}$}& 48.2\%± 13.6\% & \multicolumn{1}{c|}{0.7}& 52.2\%± 14.2\% & \multicolumn{1}{c|}{1.28} \\ \hline 
\multicolumn{5}{|c|}{\textbf{Baselines} $15 \times 15$ } \\ \hline 
\multicolumn{1}{|c|}{L2D} & 30.2\%± 6.2\% & \multicolumn{1}{c|}{0.68}& 30.9\%± 4.2\% & \multicolumn{1}{c|}{1.21} \\ 
\multicolumn{1}{|c|}{BC-$D_{\text{expert}}$} & 42.8\%± 5.0\% & \multicolumn{1}{c|}{0.7}& 43.8\%± 5.7\% & \multicolumn{1}{c|}{1.26} \\ \hline 
\multicolumn{5}{|c|}{\textbf{Offline-LD} $15 \times 15$ } \\ \hline 
\multicolumn{1}{|c|}{mQRDQN-$D_{\text{expert}}$}& 28.3\%± 4.9\%${}^{*}$ & \multicolumn{1}{c|}{0.62}& 29.1\%± 4.0\%${}^{*}$ & \multicolumn{1}{c|}{1.17} \\ 
\multicolumn{1}{|c|}{d-mSAC-$D_{\text{expert}}$}& \textbf{27.4\%± 4.7\%}${}^{*}$ & \multicolumn{1}{c|}{0.69}& 29.1\%± 3.9\%${}^{*}$ & \multicolumn{1}{c|}{1.25} \\ 
\multicolumn{1}{|c|}{mQRDQN-$D_{\text{noisy}}$}& 27.6\%± 0.9\%${}^{*}$ & \multicolumn{1}{c|}{0.64}& 29.1\%± 1.0\%${}^{*}$ & \multicolumn{1}{c|}{1.17} \\ 
\multicolumn{1}{|c|}{d-mSAC-$D_{\text{noisy}}$}& 27.5\%± 2.5\%${}^{*}$ & \multicolumn{1}{c|}{0.7}& 29.5\%± 3.3\%${}^{*}$ & \multicolumn{1}{c|}{1.27} \\ \hline 
\multicolumn{5}{|c|}{\textbf{Baselines} $20 \times 20$ } \\ \hline 
\multicolumn{1}{|c|}{L2D} & 30.8\%± 7.3\% & \multicolumn{1}{c|}{0.7}& 31.4\%± 5.4\% & \multicolumn{1}{c|}{1.24} \\ 
\multicolumn{1}{|c|}{BC-$D_{\text{expert}}$} & 39.4\%± 6.1\% & \multicolumn{1}{c|}{0.7}& 40.0\%± 5.6\% & \multicolumn{1}{c|}{1.27} \\ \hline 
\multicolumn{5}{|c|}{\textbf{Offline-LD} $20 \times 20$ } \\ \hline 
\multicolumn{1}{|c|}{mQRDQN-$D_{\text{expert}}$}& 27.7\%± 1.0\%${}^{*}$ & \multicolumn{1}{c|}{0.62}& 29.2\%± 1.6\%${}^{*}$ & \multicolumn{1}{c|}{1.15} \\ 
\multicolumn{1}{|c|}{d-mSAC-$D_{\text{expert}}$}& 28.5\%± 4.0\%${}^{*}$ & \multicolumn{1}{c|}{0.7}& 29.2\%± 3.2\%${}^{*}$ & \multicolumn{1}{c|}{1.26} \\ 
\multicolumn{1}{|c|}{mQRDQN-$D_{\text{noisy}}$}& 27.6\%± 1.7\%${}^{*}$ & \multicolumn{1}{c|}{0.63}& 29.1\%± 2.2\%${}^{*}$ & \multicolumn{1}{c|}{1.19} \\ 
\multicolumn{1}{|c|}{d-mSAC-$D_{\text{noisy}}$}& 27.7\%± 3.1\%${}^{*}$ & \multicolumn{1}{c|}{0.69}& 28.7\%± 3.2\%${}^{*}$ & \multicolumn{1}{c|}{1.27} \\ \hline 
\multicolumn{5}{|c|}{\textbf{Baselines} $30 \times 20$ } \\ \hline 
\multicolumn{1}{|c|}{L2D} & 28.0\%± 2.8\% & \multicolumn{1}{c|}{0.68}& 29.4\%± 2.3\% & \multicolumn{1}{c|}{1.22} \\ 
\multicolumn{1}{|c|}{BC-$D_{\text{expert}}$} & 43.4\%± 6.0\% & \multicolumn{1}{c|}{0.68}& 42.6\%± 5.2\% & \multicolumn{1}{c|}{1.21} \\ \hline 
\multicolumn{5}{|c|}{\textbf{Offline-LD} $30 \times 20$ } \\ \hline 
\multicolumn{1}{|c|}{mQRDQN-$D_{\text{expert}}$} & 30.1\%± 5.7\% & \multicolumn{1}{c|}{0.63}& 31.0\%± 4.6\% & \multicolumn{1}{c|}{1.17} \\ 
\multicolumn{1}{|c|}{d-mSAC-$D_{\text{expert}}$}& 29.7\%± 5.0\% & \multicolumn{1}{c|}{0.69}& 30.0\%± 4.1\% & \multicolumn{1}{c|}{1.26} \\ 
\multicolumn{1}{|c|}{mQRDQN-$D_{\text{noisy}}$}& 27.6\%± 2.0\%${}^{*}$ & \multicolumn{1}{c|}{0.63}& \textbf{29.0\%± 2.0\%}${}^{*}$ & \multicolumn{1}{c|}{1.17} \\ 
\multicolumn{1}{|c|}{d-mSAC-$D_{\text{noisy}}$}& 27.8\%± 4.0\% & \multicolumn{1}{c|}{0.7}& 28.9\%± 3.7\% & \multicolumn{1}{c|}{1.26} \\ \hline 
\multicolumn{1}{|c|}{Opt. Rate (\%)} &  \multicolumn{2}{c|}{5\%} & \multicolumn{2}{c|}{18\%}  \\ \hline 
\end{tabular}

\label{tab:large_gen_trained}
\end{table*}

\begin{table*}[ht]
\caption{The results for the generated instance for the following sizes: $15 \times 10$ and $20 \times 10$. \textbf{Gap} is the difference between the obtained result and the found CP solution. \textbf{Time} is the runtime in seconds. The \textbf{bold} result is the best RL approach. ${}^*$ signifies that an offline RL approach is a significant improvement compared to L2D of the same trained instance size, according to an independent t-test ($p=0.05$). The Opt. Rate (\%) is the percentage of the solutions that CP found an optimal result.}
\centering
\fontsize{6pt}{6pt}\selectfont
\setlength{\tabcolsep}{2mm}
\begin{tabular}{|ccccc|}
\hline
\multicolumn{1}{|c|}{} & \multicolumn{2}{|c|}{$15 \times 10$}& \multicolumn{2}{|c|}{$20 \times 10$} \\
\multicolumn{1}{|c|}{} & Gap & \multicolumn{1}{c|}{Time}& Gap & \multicolumn{1}{c|}{Time} \\ \hline 
\multicolumn{5}{|c|}{\textbf{PDR}} \\ \hline\multicolumn{1}{|c|}{SPT} & 57.8\% & \multicolumn{1}{c|}{0.01}& 57.1\% & \multicolumn{1}{c|}{0.02} \\ 
\multicolumn{1}{|c|}{MOR} & \textcolor{black}{37.6}\% & \multicolumn{1}{c|}{0.01}& \textcolor{black}{35.8}\% & \multicolumn{1}{c|}{0.01} \\ 
\multicolumn{1}{|c|}{MWKR} & \textcolor{black}{55.4}\% & \multicolumn{1}{c|}{0.01}& \textcolor{black}{56.2}\% & \multicolumn{1}{c|}{0.02} \\ \hline 
\multicolumn{5}{|c|}{\textbf{Baselines} $6 \times 6$ } \\ \hline 
\multicolumn{1}{|c|}{L2D} & 26.4\%± 6.6\% & \multicolumn{1}{c|}{0.24}& 23.6\%± 5.4\% & \multicolumn{1}{c|}{0.32} \\ 
\multicolumn{1}{|c|}{BC-$D_{\text{expert}}$} & 39.5\%± 11.2\% & \multicolumn{1}{c|}{0.22}& 35.2\%± 11.8\% & \multicolumn{1}{c|}{0.3}\\ \hline 
\multicolumn{5}{|c|}{\textbf{Offline-LD} $6 \times 6$ } \\ \hline 
\multicolumn{1}{|c|}{mQRDQN-$D_{\text{expert}}$}& 26.9\%± 4.1\% & \multicolumn{1}{c|}{0.2}& 24.4\%± 3.9\% & \multicolumn{1}{c|}{0.28} \\ 
\multicolumn{1}{|c|}{d-mSAC-$D_{\text{expert}}$}& 32.4\%± 8.1\% & \multicolumn{1}{c|}{0.22}& 29.7\%± 7.7\% & \multicolumn{1}{c|}{0.31} \\ 
\multicolumn{1}{|c|}{mQRDQN-$D_{\text{noisy}}$}& 25.2\%± 2.9\%${}^{*}$ & \multicolumn{1}{c|}{0.19}& 23.4\%± 2.8\% & \multicolumn{1}{c|}{0.28} \\ 
\multicolumn{1}{|c|}{d-mSAC-$D_{\text{noisy}}$}& 39.1\%± 11.7\% & \multicolumn{1}{c|}{0.22}& 37.5\%± 10.3\% & \multicolumn{1}{c|}{0.31} \\ \hline 
\multicolumn{5}{|c|}{\textbf{Baselines} $10 \times 10$ } \\ \hline 
\multicolumn{1}{|c|}{L2D} & 27.9\%± 6.7\% & \multicolumn{1}{c|}{0.22}& 24.8\%± 6.1\% & \multicolumn{1}{c|}{0.3}\\ 
\multicolumn{1}{|c|}{BC-$D_{\text{expert}}$} & 35.1\%± 7.0\% & \multicolumn{1}{c|}{0.21}& 31.3\%± 6.4\% & \multicolumn{1}{c|}{0.3} \\ \hline 
\multicolumn{5}{|c|}{\textbf{Offline-LD} $10 \times 10$ } \\ \hline 
\multicolumn{1}{|c|}{mQRDQN-$D_{\text{expert}}$}& 25.4\%± 3.3\%${}^{*}$ & \multicolumn{1}{c|}{0.19}& 22.1\%± 3.8\%${}^{*}$ & \multicolumn{1}{c|}{0.27} \\ 
\multicolumn{1}{|c|}{d-mSAC-$D_{\text{expert}}$}& \textbf{24.2\%± 3.8\%}${}^{*}$ & \multicolumn{1}{c|}{0.22}& \textbf{21.6\%± 3.7\%}${}^{*}$ & \multicolumn{1}{c|}{0.31} \\ 
\multicolumn{1}{|c|}{mQRDQN-$D_{\text{noisy}}$}& 24.6\%± 1.4\%${}^{*}$ & \multicolumn{1}{c|}{0.19}& 21.7\%± 2.2\%${}^{*}$ & \multicolumn{1}{c|}{0.27}\\ 
\multicolumn{1}{|c|}{d-mSAC-$D_{\text{noisy}}$}& 24.4\%± 1.9\%${}^{*}$ & \multicolumn{1}{c|}{0.22}& 21.7\%± 2.5\%${}^{*}$ & \multicolumn{1}{c|}{0.31} \\ \hline 
\multicolumn{5}{|c|}{\textbf{Baselines} $15 \times 15$ } \\ \hline 
\multicolumn{1}{|c|}{L2D} & 25.8\%± 4.9\% & \multicolumn{1}{c|}{0.22}& 22.8\%± 4.1\% & \multicolumn{1}{c|}{0.31} \\ 
\multicolumn{1}{|c|}{BC-$D_{\text{expert}}$} & 36.9\%± 6.4\% & \multicolumn{1}{c|}{0.22}& 31.4\%± 5.5\% & \multicolumn{1}{c|}{0.31} \\ \hline 
\multicolumn{5}{|c|}{\textbf{Offline-LD} $15 \times 15$ } \\ \hline 
\multicolumn{1}{|c|}{mQRDQN-$D_{\text{expert}}$}& 24.4\%± 0.3\%${}^{*}$ & \multicolumn{1}{c|}{0.2}& 21.8\%± 0.4\%${}^{*}$ & \multicolumn{1}{c|}{0.27} \\ 
\multicolumn{1}{|c|}{d-mSAC-$D_{\text{expert}}$}& 25.2\%± 4.0\% & \multicolumn{1}{c|}{0.22}& 22.0\%± 3.4\% & \multicolumn{1}{c|}{0.31} \\ 
\multicolumn{1}{|c|}{mQRDQN-$D_{\text{noisy}}$}& 24.5\%± 0.2\%${}^{*}$ & \multicolumn{1}{c|}{0.2}& 21.8\%± 0.2\%${}^{*}$ & \multicolumn{1}{c|}{0.28} \\ 
\multicolumn{1}{|c|}{d-mSAC-$D_{\text{noisy}}$}& 24.4\%± 1.7\%${}^{*}$ & \multicolumn{1}{c|}{0.22}& 21.7\%± 2.1\%${}^{*}$ & \multicolumn{1}{c|}{0.31} \\ \hline 
\multicolumn{5}{|c|}{\textbf{Baselines} $20 \times 20$ } \\ \hline 
\multicolumn{1}{|c|}{L2D} & 25.9\%± 4.6\% & \multicolumn{1}{c|}{0.23}& 23.0\%± 4.1\% & \multicolumn{1}{c|}{0.31} \\ 
\multicolumn{1}{|c|}{BC-$D_{\text{expert}}$} & 40.5\%± 6.9\% & \multicolumn{1}{c|}{0.22}& 33.1\%± 5.2\% & \multicolumn{1}{c|}{0.31} \\ \hline 
\multicolumn{5}{|c|}{\textbf{Offline-LD} $20 \times 20$ } \\ \hline 
\multicolumn{1}{|c|}{mQRDQN-$D_{\text{expert}}$}& 24.4\%± 1.3\%${}^{*}$ & \multicolumn{1}{c|}{0.19}& 21.7\%± 2.1\%${}^{*}$ & \multicolumn{1}{c|}{0.27} \\ 
\multicolumn{1}{|c|}{d-mSAC-$D_{\text{expert}}$}& 26.4\%± 4.7\% & \multicolumn{1}{c|}{0.22}& 23.7\%± 3.9\% & \multicolumn{1}{c|}{0.31} \\ 
\multicolumn{1}{|c|}{mQRDQN-$D_{\text{noisy}}$}& 24.4\%± 0.5\%${}^{*}$ & \multicolumn{1}{c|}{0.2}& 21.7\%± 1.3\%${}^{*}$ & \multicolumn{1}{c|}{0.28} \\ 
\multicolumn{1}{|c|}{d-mSAC-$D_{\text{noisy}}$}& 25.0\%± 3.1\% & \multicolumn{1}{c|}{0.22}& 22.2\%± 3.0\% & \multicolumn{1}{c|}{0.31} \\ \hline 
\multicolumn{5}{|c|}{\textbf{Baselines} $30 \times 20$ } \\ \hline 
\multicolumn{1}{|c|}{L2D} & 24.9\%± 4.5\% & \multicolumn{1}{c|}{0.22}& 22.0\%± 3.7\% & \multicolumn{1}{c|}{0.31} \\ 
\multicolumn{1}{|c|}{BC-$D_{\text{expert}}$} & 44.8\%± 8.0\% & \multicolumn{1}{c|}{0.22}& 35.9\%± 6.7\% & \multicolumn{1}{c|}{0.3} \\ \hline 
\multicolumn{5}{|c|}{\textbf{Offline-LD} $30 \times 20$ } \\ \hline 
\multicolumn{1}{|c|}{mQRDQN-$D_{\text{expert}}$}& 27.6\%± 7.6\% & \multicolumn{1}{c|}{0.2}& 23.7\%± 5.6\% & \multicolumn{1}{c|}{0.28} \\ 
\multicolumn{1}{|c|}{d-mSAC-$D_{\text{expert}}$}& 28.3\%± 7.7\% & \multicolumn{1}{c|}{0.22}& 24.4\%± 5.7\% & \multicolumn{1}{c|}{0.31} \\ 
\multicolumn{1}{|c|}{mQRDQN-$D_{\text{noisy}}$}& 24.4\%± 1.7\% & \multicolumn{1}{c|}{0.2}& 21.7\%± 1.6\% & \multicolumn{1}{c|}{0.28}\\ 
\multicolumn{1}{|c|}{d-mSAC-$D_{\text{noisy}}$}& 25.2\%± 3.7\% & \multicolumn{1}{c|}{0.22}& 21.9\%± 3.0\% & \multicolumn{1}{c|}{0.31} \\ \hline 
\multicolumn{1}{|c|}{Opt. Rate (\%)} & \multicolumn{2}{c|}{100\%} & \multicolumn{2}{c|}{98\%}  \\ \hline 
\end{tabular}

\label{tab:gen_small_small}
\end{table*}

\begin{table*}[ht]
\caption{The results for the generated instance for the following sizes: $20 \times 15$, and $30 \times 15$. \textbf{Gap} is the difference between the obtained result and the found CP solution. \textbf{Time} is the runtime in seconds. The \textbf{bold} result is the best RL approach. ${}^*$ signifies that an offline RL approach is a significant improvement compared to L2D of the same trained instance size, according to an independent t-test ($p=0.05$). The Opt. Rate (\%) is the percentage of the solutions that CP found an optimal result.}
\centering
\fontsize{6pt}{6pt}\selectfont
\setlength{\tabcolsep}{2mm}
\begin{tabular}{|ccccc|}
\hline
\multicolumn{1}{|c|}{} & \multicolumn{2}{|c|}{$20 \times 15$}& \multicolumn{2}{|c|}{$30 \times 15$} \\
\multicolumn{1}{|c|}{} & Gap & \multicolumn{1}{c|}{Time}& Gap & \multicolumn{1}{c|}{Time} \\ \hline 
\multicolumn{5}{|c|}{\textbf{PDR}} \\ \hline\multicolumn{1}{|c|}{SPT} & 65.6\% & \multicolumn{1}{c|}{0.03}& 60.0\% & \multicolumn{1}{c|}{0.05} \\ 
\multicolumn{1}{|c|}{MOR} & \textcolor{black}{46.6}\% & \multicolumn{1}{c|}{0.03}& \textcolor{black}{43.2}\% & \multicolumn{1}{c|}{0.06} \\ 
\multicolumn{1}{|c|}{MWKR} & \textcolor{black}{60.6}\% & \multicolumn{1}{c|}{0.04}& \textcolor{black}{64.2}\% & \multicolumn{1}{c|}{0.08} \\ \hline 
\multicolumn{5}{|c|}{\textbf{Baselines} $6 \times 6$ } \\ \hline 
\multicolumn{1}{|c|}{L2D} & 31.0\%± 6.4\% & \multicolumn{1}{c|}{0.5}& 26.1\%± 4.6\% & \multicolumn{1}{c|}{0.84} \\ 
\multicolumn{1}{|c|}{BC-$D_{\text{expert}}$} & 44.6\%± 14.9\% & \multicolumn{1}{c|}{0.49}& 38.9\%± 15.1\% & \multicolumn{1}{c|}{0.8} \\ \hline 
\multicolumn{5}{|c|}{\textbf{Offline-LD} $6 \times 6$ } \\ \hline 
\multicolumn{1}{|c|}{mQRDQN-$D_{\text{expert}}$}& 31.9\%± 4.0\% & \multicolumn{1}{c|}{0.45}& 27.0\%± 2.9\% & \multicolumn{1}{c|}{0.75} \\ 
\multicolumn{1}{|c|}{d-mSAC-$D_{\text{expert}}$}& 41.7\%± 7.6\% & \multicolumn{1}{c|}{0.49}& 33.7\%± 6.4\% & \multicolumn{1}{c|}{0.8} \\ 
\multicolumn{1}{|c|}{mQRDQN-$D_{\text{noisy}}$}& 30.4\%± 3.4\% & \multicolumn{1}{c|}{0.43}& 26.2\%± 2.7\% & \multicolumn{1}{c|}{0.72} \\ 
\multicolumn{1}{|c|}{d-mSAC-$D_{\text{noisy}}$}& 54.7\%± 8.7\% & \multicolumn{1}{c|}{0.49}& 47.0\%± 8.3\% & \multicolumn{1}{c|}{0.81} \\ \hline 
\multicolumn{5}{|c|}{\textbf{Baselines} $10 \times 10$ } \\ \hline 
\multicolumn{1}{|c|}{L2D} & 33.2\%± 7.3\% & \multicolumn{1}{c|}{0.47}& 27.5\%± 5.1\% & \multicolumn{1}{c|}{0.78} \\ 
\multicolumn{1}{|c|}{BC-$D_{\text{expert}}$} & 42.0\%± 6.4\% & \multicolumn{1}{c|}{0.47}& 36.0\%± 5.9\% & \multicolumn{1}{c|}{0.79} \\ \hline 
\multicolumn{5}{|c|}{\textbf{Offline-LD} $10 \times 10$ } \\ \hline 
\multicolumn{1}{|c|}{mQRDQN-$D_{\text{expert}}$}& 33.7\%± 4.5\% & \multicolumn{1}{c|}{0.44}& 27.6\%± 3.4\% & \multicolumn{1}{c|}{0.73} \\ 
\multicolumn{1}{|c|}{d-mSAC-$D_{\text{expert}}$}& 34.6\%± 6.7\% & \multicolumn{1}{c|}{0.49}& 29.5\%± 5.5\% & \multicolumn{1}{c|}{0.81} \\ 
\multicolumn{1}{|c|}{mQRDQN-$D_{\text{noisy}}$}& 30.3\%± 5.0\%${}^{*}$ & \multicolumn{1}{c|}{0.44}& 26.1\%± 4.3\%${}^{*}$ & \multicolumn{1}{c|}{0.73} \\ 
\multicolumn{1}{|c|}{d-mSAC-$D_{\text{noisy}}$}& 33.2\%± 7.8\% & \multicolumn{1}{c|}{0.5}& 38.6\%± 11.0\% & \multicolumn{1}{c|}{0.82} \\ \hline 
\multicolumn{5}{|c|}{\textbf{Baselines} $15 \times 15$ } \\ \hline 
\multicolumn{1}{|c|}{L2D} & 30.3\%± 5.1\% & \multicolumn{1}{c|}{0.48}& 25.7\%± 3.3\% & \multicolumn{1}{c|}{0.79} \\ 
\multicolumn{1}{|c|}{BC-$D_{\text{expert}}$} & 39.6\%± 6.1\% & \multicolumn{1}{c|}{0.49}& 34.2\%± 5.2\% & \multicolumn{1}{c|}{0.81} \\ \hline 
\multicolumn{5}{|c|}{\textbf{Offline-LD} $15 \times 15$ } \\ \hline 
\multicolumn{1}{|c|}{mQRDQN-$D_{\text{expert}}$}& 28.3\%± 1.7\%${}^{*}$ & \multicolumn{1}{c|}{0.44}& 24.4\%± 2.7\%${}^{*}$ & \multicolumn{1}{c|}{0.74} \\ 
\multicolumn{1}{|c|}{d-mSAC-$D_{\text{expert}}$}& 29.0\%± 3.5\%${}^{*}$ & \multicolumn{1}{c|}{0.49}& 25.0\%± 3.2\%${}^{*}$ & \multicolumn{1}{c|}{0.81} \\ 
\multicolumn{1}{|c|}{mQRDQN-$D_{\text{noisy}}$}& 28.2\%± 0.7\%${}^{*}$ & \multicolumn{1}{c|}{0.45}& 24.4\%± 1.1\%${}^{*}$ & \multicolumn{1}{c|}{0.74} \\ 
\multicolumn{1}{|c|}{d-mSAC-$D_{\text{noisy}}$}& 28.3\%± 2.4\%${}^{*}$ & \multicolumn{1}{c|}{0.5}& 24.6\%± 2.6\%${}^{*}$ & \multicolumn{1}{c|}{0.83} \\ \hline 
\multicolumn{5}{|c|}{\textbf{Baselines} $20 \times 20$ } \\ \hline 
\multicolumn{1}{|c|}{L2D} & 31.1\%± 6.4\% & \multicolumn{1}{c|}{0.49}& 26.1\%± 4.2\% & \multicolumn{1}{c|}{0.81} \\ 
\multicolumn{1}{|c|}{BC-$D_{\text{expert}}$} & 41.1\%± 6.3\% & \multicolumn{1}{c|}{0.49}& 33.8\%± 5.2\% & \multicolumn{1}{c|}{0.81} \\ \hline 
\multicolumn{5}{|c|}{\textbf{Offline-LD} $20 \times 20$ } \\ \hline 
\multicolumn{1}{|c|}{mQRDQN-$D_{\text{expert}}$}& 28.5\%± 1.6\%${}^{*}$ & \multicolumn{1}{c|}{0.44}& \textbf{24.2\%± 1.4\%}${}^{*}$ & \multicolumn{1}{c|}{0.73} \\ 
\multicolumn{1}{|c|}{d-mSAC-$D_{\text{expert}}$}& 29.6\%± 4.0\%${}^{*}$ & \multicolumn{1}{c|}{0.5}& 25.7\%± 3.4\% & \multicolumn{1}{c|}{0.81} \\ 
\multicolumn{1}{|c|}{mQRDQN-$D_{\text{noisy}}$}& 28.3\%± 1.4\%${}^{*}$ & \multicolumn{1}{c|}{0.45}& 24.3\%± 1.5\%${}^{*}$ & \multicolumn{1}{c|}{0.74} \\ 
\multicolumn{1}{|c|}{d-mSAC-$D_{\text{noisy}}$}& 29.0\%± 3.4\%${}^{*}$ & \multicolumn{1}{c|}{0.49}& 24.8\%± 3.1\%${}^{*}$ & \multicolumn{1}{c|}{0.81} \\ \hline 
\multicolumn{5}{|c|}{\textbf{Baselines} $30 \times 20$ } \\ \hline 
\multicolumn{1}{|c|}{L2D} & 29.1\%± 4.2\% & \multicolumn{1}{c|}{0.48}& 24.8\%± 3.1\% & \multicolumn{1}{c|}{0.8} \\ 
\multicolumn{1}{|c|}{BC-$D_{\text{expert}}$} & 43.9\%± 6.6\% & \multicolumn{1}{c|}{0.48}& 35.4\%± 5.1\% & \multicolumn{1}{c|}{0.79} \\ \hline 
\multicolumn{5}{|c|}{\textbf{Offline-LD} $30 \times 20$ } \\ \hline 
\multicolumn{1}{|c|}{mQRDQN-$D_{\text{expert}}$}& 29.7\%± 5.2\% & \multicolumn{1}{c|}{0.45}& 25.3\%± 4.0\% & \multicolumn{1}{c|}{0.75} \\ 
\multicolumn{1}{|c|}{d-mSAC-$D_{\text{expert}}$}& 30.6\%± 5.4\% & \multicolumn{1}{c|}{0.49}& 25.8\%± 3.8\% & \multicolumn{1}{c|}{0.81} \\ 
\multicolumn{1}{|c|}{mQRDQN-$D_{\text{noisy}}$}& \textbf{28.1\%± 2.0\%}${}^{*}$ & \multicolumn{1}{c|}{0.44}& 24.3\%± 1.8\%${}^{*}$ & \multicolumn{1}{c|}{0.74} \\ 
\multicolumn{1}{|c|}{d-mSAC-$D_{\text{noisy}}$}& 28.6\%± 3.5\% & \multicolumn{1}{c|}{0.49}& 24.5\%± 3.1\% & \multicolumn{1}{c|}{0.83} \\ \hline 
\multicolumn{1}{|c|}{Opt. Rate (\%)} & \multicolumn{2}{c|}{24\%} & \multicolumn{2}{c|}{82\%}  \\ \hline 
\end{tabular}

\label{tab:gen_small_large}
\end{table*}
\begin{table*}[ht]
\caption{The results for the generated instance for the following sizes: $50 \times 20$, and $100 \times 20$. \textbf{Gap} is the difference between the obtained result and the found CP solution. \textbf{Time} is the runtime in seconds. The \textbf{bold} result is the best RL approach. ${}^*$ signifies that an offline RL approach is a significant improvement compared to L2D of the same trained instance size, according to an independent t-test ($p=0.05$). The Opt. Rate (\%) is the percentage of the solutions that CP found an optimal result.}
\centering
\fontsize{6pt}{6pt}\selectfont
\setlength{\tabcolsep}{2mm}
\begin{tabular}{|ccccc|}
\hline
\multicolumn{1}{|c|}{} & \multicolumn{2}{|c|}{$50 \times 20$}& \multicolumn{2}{|c|}{$100 \times 20$} \\
\multicolumn{1}{|c|}{} & Gap & \multicolumn{1}{c|}{Time}& Gap & \multicolumn{1}{c|}{Time} \\ \hline 
\multicolumn{5}{|c|}{\textbf{PDR}} \\ \hline\multicolumn{1}{|c|}{SPT} & 57.0\% & \multicolumn{1}{c|}{0.49}& 37.1\% & \multicolumn{1}{c|}{4.1} \\ 
\multicolumn{1}{|c|}{MOR} & \textcolor{black}{39.7}\% & \multicolumn{1}{c|}{0.52}& \textcolor{black}{21.3}\% & \multicolumn{1}{c|}{4.19} \\ 
\multicolumn{1}{|c|}{MWKR} & \textcolor{black}{63.6}\% & \multicolumn{1}{c|}{0.57}& \textcolor{black}{46.2}\% & \multicolumn{1}{c|}{4.59} \\ \hline 
\multicolumn{5}{|c|}{\textbf{Baselines} $6 \times 6$ } \\ \hline 
\multicolumn{1}{|c|}{L2D} & 22.7\%± 3.9\% & \multicolumn{1}{c|}{2.84}& 9.4\%± 1.9\% & \multicolumn{1}{c|}{12.24} \\ 
\multicolumn{1}{|c|}{BC-$D_{\text{expert}}$} & 36.5\%± 16.5\% & \multicolumn{1}{c|}{2.8}& 18.5\%± 11.8\% & \multicolumn{1}{c|}{12.39} \\ \hline 
\multicolumn{5}{|c|}{\textbf{Offline-LD} $6 \times 6$ } \\ \hline 
\multicolumn{1}{|c|}{mQRDQN-$D_{\text{expert}}$}& 22.5\%± 2.2\% & \multicolumn{1}{c|}{2.68}& 8.8\%± 1.2\%${}^{*}$ & \multicolumn{1}{c|}{12.76} \\ 
\multicolumn{1}{|c|}{d-mSAC-$D_{\text{expert}}$}& 27.4\%± 4.6\% & \multicolumn{1}{c|}{2.81}& 10.7\%± 2.2\% & \multicolumn{1}{c|}{12.52} \\ 
\multicolumn{1}{|c|}{mQRDQN-$D_{\text{noisy}}$}& 21.7\%± 2.3\%${}^{*}$ & \multicolumn{1}{c|}{2.63}& 8.8\%± 1.2\%${}^{*}$ & \multicolumn{1}{c|}{12.61} \\ 
\multicolumn{1}{|c|}{d-mSAC-$D_{\text{noisy}}$}& 41.3\%± 10.0\% & \multicolumn{1}{c|}{2.9}& 18.6\%± 7.2\% & \multicolumn{1}{c|}{11.92} \\ \hline 
\multicolumn{5}{|c|}{\textbf{Baselines} $10 \times 10$ } \\ \hline 
\multicolumn{1}{|c|}{L2D} & 24.0\%± 4.5\% & \multicolumn{1}{c|}{2.7}& 9.7\%± 2.0\% & \multicolumn{1}{c|}{11.91} \\ 
\multicolumn{1}{|c|}{BC-$D_{\text{expert}}$} & 36.5\%± 7.9\% & \multicolumn{1}{c|}{2.84}& 17.8\%± 6.8\% & \multicolumn{1}{c|}{11.57} \\ \hline 
\multicolumn{5}{|c|}{\textbf{Offline-LD} $10 \times 10$ } \\ \hline 
\multicolumn{1}{|c|}{mQRDQN-$D_{\text{expert}}$}& 23.2\%± 2.6\%${}^{*}$ & \multicolumn{1}{c|}{2.79}& 9.1\%± 1.4\%${}^{*}$ & \multicolumn{1}{c|}{12.52} \\ 
\multicolumn{1}{|c|}{d-mSAC-$D_{\text{expert}}$}& 26.8\%± 4.4\% & \multicolumn{1}{c|}{2.96}& 10.6\%± 2.1\% & \multicolumn{1}{c|}{12.4} \\ 
\multicolumn{1}{|c|}{mQRDQN-$D_{\text{noisy}}$}& 22.8\%± 3.0\%${}^{*}$ & \multicolumn{1}{c|}{2.75}& 8.9\%± 1.7\%${}^{*}$ & \multicolumn{1}{c|}{11.98} \\ 
\multicolumn{1}{|c|}{d-mSAC-$D_{\text{noisy}}$}& 42.8\%± 10.9\% & \multicolumn{1}{c|}{2.89}& 24.2\%± 7.1\% & \multicolumn{1}{c|}{11.84} \\ \hline 
\multicolumn{5}{|c|}{\textbf{Baselines} $15 \times 15$ } \\ \hline 
\multicolumn{1}{|c|}{L2D} & 22.5\%± 3.3\% & \multicolumn{1}{c|}{2.74}& 9.1\%± 1.5\% & \multicolumn{1}{c|}{12.07} \\ 
\multicolumn{1}{|c|}{BC-$D_{\text{expert}}$} & 36.9\%± 7.1\% & \multicolumn{1}{c|}{2.79}& 19.8\%± 6.5\% & \multicolumn{1}{c|}{12.51} \\ \hline 
\multicolumn{5}{|c|}{\textbf{Offline-LD} $15 \times 15$ } \\ \hline 
\multicolumn{1}{|c|}{mQRDQN-$D_{\text{expert}}$}& 21.6\%± 2.9\%${}^{*}$ & \multicolumn{1}{c|}{2.68}& 8.8\%± 1.4\%${}^{*}$ & \multicolumn{1}{c|}{12.89} \\ 
\multicolumn{1}{|c|}{d-mSAC-$D_{\text{expert}}$}& 21.7\%± 2.9\%${}^{*}$ & \multicolumn{1}{c|}{2.82}& 9.2\%± 1.7\% & \multicolumn{1}{c|}{12.79} \\ 
\multicolumn{1}{|c|}{mQRDQN-$D_{\text{noisy}}$}& 21.2\%± 1.5\%${}^{*}$ & \multicolumn{1}{c|}{2.67}& \textbf{8.5\%± 1.0\%}${}^{*}$ & \multicolumn{1}{c|}{12.42} \\ 
\multicolumn{1}{|c|}{d-mSAC-$D_{\text{noisy}}$}& 25.6\%± 7.8\% & \multicolumn{1}{c|}{2.9}& 13.5\%± 7.4\% & \multicolumn{1}{c|}{12.32} \\ \hline 
\multicolumn{5}{|c|}{\textbf{Baselines} $20 \times 20$ } \\ \hline 
\multicolumn{1}{|c|}{L2D} & 23.0\%± 4.3\% & \multicolumn{1}{c|}{2.77}& 9.5\%± 2.0\% & \multicolumn{1}{c|}{12.2} \\ 
\multicolumn{1}{|c|}{BC-$D_{\text{expert}}$} & 32.6\%± 6.0\% & \multicolumn{1}{c|}{2.81}& 17.5\%± 6.1\% & \multicolumn{1}{c|}{12.33} \\ \hline 
\multicolumn{5}{|c|}{\textbf{Offline-LD} $20 \times 20$ } \\ \hline 
\multicolumn{1}{|c|}{mQRDQN-$D_{\text{expert}}$}& 21.3\%± 1.8\%${}^{*}$ & \multicolumn{1}{c|}{2.63}& 8.6\%± 1.1\%${}^{*}$ & \multicolumn{1}{c|}{12.84} \\ 
\multicolumn{1}{|c|}{d-mSAC-$D_{\text{expert}}$}& 21.7\%± 2.3\%${}^{*}$ & \multicolumn{1}{c|}{2.96}& 8.9\%± 1.3\%${}^{*}$ & \multicolumn{1}{c|}{13.32} \\ 
\multicolumn{1}{|c|}{mQRDQN-$D_{\text{noisy}}$}& 21.2\%± 1.8\%${}^{*}$ & \multicolumn{1}{c|}{2.74}& 8.5\%± 1.1\%${}^{*}$ & \multicolumn{1}{c|}{12.68} \\ 
\multicolumn{1}{|c|}{d-mSAC-$D_{\text{noisy}}$}& 21.5\%± 2.4\%${}^{*}$ & \multicolumn{1}{c|}{2.9}& 9.6\%± 2.3\% & \multicolumn{1}{c|}{13.1} \\ \hline 
\multicolumn{5}{|c|}{\textbf{Baselines} $30 \times 20$ } \\ \hline 
\multicolumn{1}{|c|}{L2D} & 21.5\%± 1.7\% & \multicolumn{1}{c|}{2.88}& 8.8\%± 0.9\% & \multicolumn{1}{c|}{12.93} \\ 
\multicolumn{1}{|c|}{BC-$D_{\text{expert}}$} & 33.4\%± 4.1\% & \multicolumn{1}{c|}{2.82}& 17.2\%± 3.8\% & \multicolumn{1}{c|}{12.42} \\ \hline 
\multicolumn{5}{|c|}{\textbf{Offline-LD} $30 \times 20$ } \\ \hline 
\multicolumn{1}{|c|}{mQRDQN-$D_{\text{expert}}$}& 22.8\%± 3.6\% & \multicolumn{1}{c|}{2.8}& 9.4\%± 1.8\% & \multicolumn{1}{c|}{12.81} \\ 
\multicolumn{1}{|c|}{d-mSAC-$D_{\text{expert}}$}& 22.1\%± 2.9\% & \multicolumn{1}{c|}{2.89}& 9.0\%± 1.5\% & \multicolumn{1}{c|}{12.14} \\ 
\multicolumn{1}{|c|}{mQRDQN-$D_{\text{noisy}}$}& \textbf{21.2\%± 1.4\%}${}^{*}$ & \multicolumn{1}{c|}{2.7}& 8.7\%± 0.7\% & \multicolumn{1}{c|}{12.74} \\ 
\multicolumn{1}{|c|}{d-mSAC-$D_{\text{noisy}}$}& 21.7\%± 2.4\% & \multicolumn{1}{c|}{2.88}& 8.7\%± 1.3\% & \multicolumn{1}{c|}{12.41} \\ \hline 
\multicolumn{1}{|c|}{Opt. Rate (\%)} & \multicolumn{2}{c|}{89\%} & \multicolumn{2}{c|}{100\%}  \\ \hline 
\end{tabular}

\label{tab:gen_big}
\end{table*}
\begin{table*}[ht]
\caption{The results for the Taillard instance for the following sizes: $15 \times 15$, $20 \times 20$, and $30 \times 20$. \textbf{Gap} is the difference between the obtained result and the upper bound. \textbf{Time} is the runtime in seconds. The \textbf{bold} result is the best RL approach. ${}^*$ signifies that an offline RL approach is a significant improvement compared to L2D of the same trained instance size, according to an independent t-test ($p=0.05$). The Opt. Rate (\%) is the percentage of the solutions that CP found an optimal result.}
\centering
\fontsize{6pt}{6pt}\selectfont
\setlength{\tabcolsep}{2mm}
\begin{tabular}{|ccccccc|}
\hline
\multicolumn{1}{|c|}{} & \multicolumn{2}{|c|}{\begin{tabular}[c]{@{}c@{}}Taillard\\$15 \times 15$\end{tabular}}& \multicolumn{2}{|c|}{\begin{tabular}[c]{@{}c@{}}Taillard\\$20 \times 20$\end{tabular}}& \multicolumn{2}{|c|}{\begin{tabular}[c]{@{}c@{}}Taillard\\$30 \times 20$\end{tabular}} \\
\multicolumn{1}{|c|}{} & Gap & \multicolumn{1}{c|}{Time}& Gap & \multicolumn{1}{c|}{Time}& Gap & \multicolumn{1}{c|}{Time} \\ \hline 
\multicolumn{7}{|c|}{\textbf{PDR}} \\ \hline\multicolumn{1}{|c|}{SPT} & 56.9\% & \multicolumn{1}{c|}{0.02}& 65.3\% & \multicolumn{1}{c|}{0.04}& 67.3\% & \multicolumn{1}{c|}{0.12} \\ 
\multicolumn{1}{|c|}{MOR} & \textcolor{black}{41.4}\% & \multicolumn{1}{c|}{0.02}& \textcolor{black}{44.4}\% & \multicolumn{1}{c|}{0.04}& \textcolor{black}{54.6}\% & \multicolumn{1}{c|}{0.09} \\ 
\multicolumn{1}{|c|}{MWKR} & \textcolor{black}{54.3}\% & \multicolumn{1}{c|}{0.02}& \textcolor{black}{62.8}\% & \multicolumn{1}{c|}{0.05}& \textcolor{black}{67.7}\% & \multicolumn{1}{c|}{0.11} \\ \hline 
\multicolumn{7}{|c|}{\textbf{Baselines} $6 \times 6$ } \\ \hline 
\multicolumn{1}{|c|}{L2D} & 28.0\%± 8.5\% & \multicolumn{1}{c|}{0.36}& 31.5\%± 7.4\% & \multicolumn{1}{c|}{0.72}& 34.2\%± 4.9\% & \multicolumn{1}{c|}{1.27} \\ 
\multicolumn{1}{|c|}{BC-$D_{\text{expert}}$} & 42.3\%± 14.1\% & \multicolumn{1}{c|}{0.35}& 43.9\%± 16.2\% & \multicolumn{1}{c|}{0.69}& 49.8\%± 19.3\% & \multicolumn{1}{c|}{1.24} \\ \hline 
\multicolumn{7}{|c|}{\textbf{Offline-LD} $6 \times 6$ } \\ \hline 
\multicolumn{1}{|c|}{mQRDQN-$D_{\text{expert}}$}& 27.8\%± 3.5\% & \multicolumn{1}{c|}{0.32}& 29.9\%± 2.6\% & \multicolumn{1}{c|}{0.63}& 33.8\%± 4.0\% & \multicolumn{1}{c|}{1.18} \\ 
\multicolumn{1}{|c|}{d-mSAC-$D_{\text{expert}}$}& 41.0\%± 7.1\% & \multicolumn{1}{c|}{0.35}& 40.5\%± 8.1\% & \multicolumn{1}{c|}{0.7}& 43.1\%± 7.1\% & \multicolumn{1}{c|}{1.26} \\ 
\multicolumn{1}{|c|}{mQRDQN-$D_{\text{noisy}}$}& 25.7\%± 3.1\% & \multicolumn{1}{c|}{0.31}& 29.0\%± 2.2\% & \multicolumn{1}{c|}{0.62}& 32.5\%± 3.6\% & \multicolumn{1}{c|}{1.15} \\ 
\multicolumn{1}{|c|}{d-mSAC-$D_{\text{noisy}}$}& 49.6\%± 7.9\% & \multicolumn{1}{c|}{0.35}& 55.6\%± 11.3\% & \multicolumn{1}{c|}{0.7}& 57.6\%± 9.6\% & \multicolumn{1}{c|}{1.26} \\ \hline 
\multicolumn{7}{|c|}{\textbf{Baselines} $10 \times 10$ } \\ \hline 
\multicolumn{1}{|c|}{L2D} & 31.7\%± 8.7\% & \multicolumn{1}{c|}{0.34}& 33.6\%± 8.1\% & \multicolumn{1}{c|}{0.67}& 36.3\%± 5.6\% & \multicolumn{1}{c|}{1.2} \\ 
\multicolumn{1}{|c|}{BC-$D_{\text{expert}}$} & 39.6\%± 6.6\% & \multicolumn{1}{c|}{0.34}& 46.2\%± 6.9\% & \multicolumn{1}{c|}{0.68}& 49.6\%± 7.3\% & \multicolumn{1}{c|}{1.23} \\ \hline 
\multicolumn{7}{|c|}{\textbf{Offline-LD} $10 \times 10$ } \\ \hline 
\multicolumn{1}{|c|}{mQRDQN-$D_{\text{expert}}$}& 31.0\%± 5.5\% & \multicolumn{1}{c|}{0.31}& 30.9\%± 2.8\% & \multicolumn{1}{c|}{0.63}& 36.3\%± 4.9\% & \multicolumn{1}{c|}{1.16} \\ 
\multicolumn{1}{|c|}{d-mSAC-$D_{\text{expert}}$}& 27.8\%± 6.2\%${}^{*}$ & \multicolumn{1}{c|}{0.35}& 39.7\%± 8.3\% & \multicolumn{1}{c|}{0.7}& 41.1\%± 6.4\% & \multicolumn{1}{c|}{1.27} \\ 
\multicolumn{1}{|c|}{mQRDQN-$D_{\text{noisy}}$}& 27.4\%± 4.9\%${}^{*}$ & \multicolumn{1}{c|}{0.31}& 29.3\%± 4.3\%${}^{*}$ & \multicolumn{1}{c|}{0.63}& 34.2\%± 5.2\% & \multicolumn{1}{c|}{1.14} \\ 
\multicolumn{1}{|c|}{d-mSAC-$D_{\text{noisy}}$}& 25.0\%± 5.4\%${}^{*}$ & \multicolumn{1}{c|}{0.36}& 51.6\%± 15.7\% & \multicolumn{1}{c|}{0.71}& 55.8\%± 12.5\% & \multicolumn{1}{c|}{1.28} \\ \hline 
\multicolumn{7}{|c|}{\textbf{Baselines} $15 \times 15$ } \\ \hline 
\multicolumn{1}{|c|}{L2D} & 27.4\%± 5.0\% & \multicolumn{1}{c|}{0.35}& 31.7\%± 6.4\% & \multicolumn{1}{c|}{0.68}& 34.7\%± 4.3\% & \multicolumn{1}{c|}{1.22} \\ 
\multicolumn{1}{|c|}{BC-$D_{\text{expert}}$} & 36.1\%± 6.7\% & \multicolumn{1}{c|}{0.35}& 44.4\%± 5.5\% & \multicolumn{1}{c|}{0.7}& 49.1\%± 5.4\% & \multicolumn{1}{c|}{1.27} \\ \hline 
\multicolumn{7}{|c|}{\textbf{Offline-LD} $15 \times 15$ } \\ \hline 
\multicolumn{1}{|c|}{mQRDQN-$D_{\text{expert}}$}& 25.5\%± 1.4\%${}^{*}$ & \multicolumn{1}{c|}{0.31}& 29.3\%± 6.2\% & \multicolumn{1}{c|}{0.64}& 34.2\%± 4.8\% & \multicolumn{1}{c|}{1.16} \\ 
\multicolumn{1}{|c|}{d-mSAC-$D_{\text{expert}}$}& 23.9\%± 2.8\%${}^{*}$ & \multicolumn{1}{c|}{0.35}& 28.1\%± 4.4\%${}^{*}$ & \multicolumn{1}{c|}{0.69}& 32.2\%± 3.9\%${}^{*}$ & \multicolumn{1}{c|}{1.26} \\ 
\multicolumn{1}{|c|}{mQRDQN-$D_{\text{noisy}}$}& 25.2\%± 0.7\%${}^{*}$ & \multicolumn{1}{c|}{0.32}& 29.2\%± 0.5\%${}^{*}$ & \multicolumn{1}{c|}{0.64}& 33.3\%± 1.4\%${}^{*}$ & \multicolumn{1}{c|}{1.17} \\ 
\multicolumn{1}{|c|}{d-mSAC-$D_{\text{noisy}}$}& 25.4\%± 2.0\%${}^{*}$ & \multicolumn{1}{c|}{0.36}& 28.7\%± 1.4\%${}^{*}$ & \multicolumn{1}{c|}{0.71}& 33.6\%± 3.6\% & \multicolumn{1}{c|}{1.27} \\ \hline 
\multicolumn{7}{|c|}{\textbf{Baselines} $20 \times 20$ } \\ \hline 
\multicolumn{1}{|c|}{L2D} & 28.1\%± 5.7\% & \multicolumn{1}{c|}{0.36}& 31.8\%± 6.7\% & \multicolumn{1}{c|}{0.7}& 35.2\%± 5.6\% & \multicolumn{1}{c|}{1.25} \\ 
\multicolumn{1}{|c|}{BC-$D_{\text{expert}}$} & 37.8\%± 6.8\% & \multicolumn{1}{c|}{0.36}& 41.2\%± 5.4\% & \multicolumn{1}{c|}{0.7}& 44.5\%± 7.1\% & \multicolumn{1}{c|}{1.28} \\ \hline 
\multicolumn{7}{|c|}{\textbf{Offline-LD} $20 \times 20$ } \\ \hline 
\multicolumn{1}{|c|}{mQRDQN-$D_{\text{expert}}$}& 25.8\%± 1.2\%${}^{*}$ & \multicolumn{1}{c|}{0.31}& 29.0\%± 0.3\%${}^{*}$ & \multicolumn{1}{c|}{0.63}& 33.4\%± 1.3\%${}^{*}$ & \multicolumn{1}{c|}{1.17} \\ 
\multicolumn{1}{|c|}{d-mSAC-$D_{\text{expert}}$}& 27.1\%± 3.4\% & \multicolumn{1}{c|}{0.36}& 28.6\%± 3.1\%${}^{*}$ & \multicolumn{1}{c|}{0.71}& \textbf{32.1\%± 3.1\%}${}^{*}$ & \multicolumn{1}{c|}{1.27} \\ 
\multicolumn{1}{|c|}{mQRDQN-$D_{\text{noisy}}$}& 25.8\%± 1.1\%${}^{*}$ & \multicolumn{1}{c|}{0.32}& 28.9\%± 1.4\%${}^{*}$ & \multicolumn{1}{c|}{0.64}& 33.1\%± 2.1\%${}^{*}$ & \multicolumn{1}{c|}{1.2} \\ 
\multicolumn{1}{|c|}{d-mSAC-$D_{\text{noisy}}$}& \textbf{22.5\%± 2.5\%}${}^{*}$ & \multicolumn{1}{c|}{0.35}& \textbf{28.0\%± 3.9}\%${}^{*}$ & \multicolumn{1}{c|}{0.69}& 32.9\%± 3.6\% & \multicolumn{1}{c|}{1.28} \\ \hline 
\multicolumn{7}{|c|}{\textbf{Baselines} $30 \times 20$ } \\ \hline 
\multicolumn{1}{|c|}{L2D} & 27.0\%± 5.3\% & \multicolumn{1}{c|}{0.35}& 29.9\%± 3.4\% & \multicolumn{1}{c|}{0.68}& 33.6\%± 2.8\% & \multicolumn{1}{c|}{1.23} \\ 
\multicolumn{1}{|c|}{BC-$D_{\text{expert}}$} & 41.2\%± 9.2\% & \multicolumn{1}{c|}{0.35}& 44.7\%± 4.6\% & \multicolumn{1}{c|}{0.68}& 45.2\%± 4.6\% & \multicolumn{1}{c|}{1.24} \\ \hline 
\multicolumn{7}{|c|}{\textbf{Offline-LD} $30 \times 20$ } \\ \hline 
\multicolumn{1}{|c|}{mQRDQN-$D_{\text{expert}}$}& 27.2\%± 6.7\% & \multicolumn{1}{c|}{0.32}& 30.2\%± 5.8\% & \multicolumn{1}{c|}{0.65}& 35.2\%± 4.2\% & \multicolumn{1}{c|}{1.19} \\ 
\multicolumn{1}{|c|}{d-mSAC-$D_{\text{expert}}$}& 26.9\%± 5.7\% & \multicolumn{1}{c|}{0.36}& 29.4\%± 4.9\% & \multicolumn{1}{c|}{0.71}& 34.0\%± 4.4\% & \multicolumn{1}{c|}{1.27} \\ 
\multicolumn{1}{|c|}{mQRDQN-$D_{\text{noisy}}$}& 25.3\%± 1.6\% & \multicolumn{1}{c|}{0.32}& 28.9\%± 1.9\% & \multicolumn{1}{c|}{0.64}& 33.5\%± 2.0\% & \multicolumn{1}{c|}{1.18} \\ 
\multicolumn{1}{|c|}{d-mSAC-$D_{\text{noisy}}$}& 24.9\%± 3.1\% & \multicolumn{1}{c|}{0.36}& 28.6\%± 3.5\% & \multicolumn{1}{c|}{0.71}& 32.8\%± 4.0\% & \multicolumn{1}{c|}{1.28} \\ \hline 
\multicolumn{1}{|c|}{Opt. Rate (\%)} & \multicolumn{2}{c|}{100\%} & \multicolumn{2}{c|}{30\%} & \multicolumn{2}{c|}{0\%}  \\ \hline 
\end{tabular}

\label{tab:tai_trained}
\end{table*}

\begin{table*}[ht]
\caption{The results for the Taillard instance for the following sizes: $20 \times 15$, $30 \times 15$ and $50 \times 15$. \textbf{Gap} is the difference between the obtained result and the upper bound. \textbf{Time} is the runtime in seconds. The \textbf{bold} result is the best RL approach. ${}^*$ signifies that an offline RL approach is a significant improvement compared to L2D of the same trained instance size, according to an independent t-test ($p=0.05$). The Opt. Rate (\%) is the percentage of the solutions that CP found an optimal result.}
\centering
\fontsize{6pt}{6pt}\selectfont
\setlength{\tabcolsep}{2mm}
\begin{tabular}{|ccccccc|}
\hline
\multicolumn{1}{|c|}{} & \multicolumn{2}{|c|}{\begin{tabular}[c]{@{}c@{}}Taillard\\$20 \times 15$\end{tabular}}& \multicolumn{2}{|c|}{\begin{tabular}[c]{@{}c@{}}Taillard\\$30 \times 15$\end{tabular}}& \multicolumn{2}{|c|}{\begin{tabular}[c]{@{}c@{}}Taillard\\$50 \times 15$\end{tabular}} \\
\multicolumn{1}{|c|}{} & Gap & \multicolumn{1}{c|}{Time}& Gap & \multicolumn{1}{c|}{Time}& Gap & \multicolumn{1}{c|}{Time} \\ \hline 
\multicolumn{7}{|c|}{\textbf{PDR}} \\ \hline\multicolumn{1}{|c|}{SPT} & 63.8\% & \multicolumn{1}{c|}{0.03}& 65.3\% & \multicolumn{1}{c|}{0.04}& 50.3\% & \multicolumn{1}{c|}{0.24} \\ 
\multicolumn{1}{|c|}{MOR} & \textcolor{black}{41.4}\% & \multicolumn{1}{c|}{0.03}& \textcolor{black}{43.6}\% & \multicolumn{1}{c|}{0.04}& \textcolor{black}{34}\% & \multicolumn{1}{c|}{0.25} \\ 
\multicolumn{1}{|c|}{MWKR} & \textcolor{black}{59.8}\% & \multicolumn{1}{c|}{0.04}& \textcolor{black}{66.9}\% & \multicolumn{1}{c|}{0.05}& \textcolor{black}{60.4}\% & \multicolumn{1}{c|}{0.29} \\ \hline 
\multicolumn{7}{|c|}{\textbf{Baselines} $6 \times 6$ } \\ \hline 
\multicolumn{1}{|c|}{L2D} & 32.4\%± 5.8\% & \multicolumn{1}{c|}{0.51}& 31.5\%± 7.4\% & \multicolumn{1}{c|}{0.72}& 20.3\%± 2.8\% & \multicolumn{1}{c|}{1.78} \\ 
\multicolumn{1}{|c|}{BC-$D_{\text{expert}}$} & 44.7\%± 15.7\% & \multicolumn{1}{c|}{0.49}& 43.9\%± 16.2\% & \multicolumn{1}{c|}{0.69}& 31.4\%± 13.7\% & \multicolumn{1}{c|}{1.77} \\ \hline 
\multicolumn{7}{|c|}{\textbf{Offline-LD} $6 \times 6$ } \\ \hline 
\multicolumn{1}{|c|}{mQRDQN-$D_{\text{expert}}$}& 32.2\%± 4.0\% & \multicolumn{1}{c|}{0.45}& 29.9\%± 2.6\% & \multicolumn{1}{c|}{0.63}& \textbf{19.2\%± 1.4\%} & \multicolumn{1}{c|}{1.67} \\ 
\multicolumn{1}{|c|}{d-mSAC-$D_{\text{expert}}$}& 42.0\%± 6.2\% & \multicolumn{1}{c|}{0.49}& 40.5\%± 8.1\% & \multicolumn{1}{c|}{0.7}& 23.2\%± 4.8\% & \multicolumn{1}{c|}{1.73} \\ 
\multicolumn{1}{|c|}{mQRDQN-$D_{\text{noisy}}$}& 30.2\%± 3.8\% & \multicolumn{1}{c|}{0.44}& 29.0\%± 2.2\% & \multicolumn{1}{c|}{0.62}& 19.8\%± 1.5\% & \multicolumn{1}{c|}{1.64} \\ 
\multicolumn{1}{|c|}{d-mSAC-$D_{\text{noisy}}$}& 54.9\%± 8.5\% & \multicolumn{1}{c|}{0.49}& 55.6\%± 11.3\% & \multicolumn{1}{c|}{0.7}& 31.5\%± 5.6\% & \multicolumn{1}{c|}{1.77} \\ \hline 
\multicolumn{7}{|c|}{\textbf{Baselines} $10 \times 10$ } \\ \hline 
\multicolumn{1}{|c|}{L2D} & 34.8\%± 6.6\% & \multicolumn{1}{c|}{0.48}& 33.6\%± 8.1\% & \multicolumn{1}{c|}{0.67}& 21.6\%± 3.3\% & \multicolumn{1}{c|}{1.68} \\ 
\multicolumn{1}{|c|}{BC-$D_{\text{expert}}$} & 41.7\%± 5.1\% & \multicolumn{1}{c|}{0.48}& 46.2\%± 6.9\% & \multicolumn{1}{c|}{0.68}& 28.8\%± 5.0\% & \multicolumn{1}{c|}{1.76} \\ \hline 
\multicolumn{7}{|c|}{\textbf{Offline-LD} $10 \times 10$ } \\ \hline 
\multicolumn{1}{|c|}{mQRDQN-$D_{\text{expert}}$}& 35.9\%± 4.2\% & \multicolumn{1}{c|}{0.44}& 30.9\%± 2.8\% & \multicolumn{1}{c|}{0.63}& 19.7\%± 2.6\% & \multicolumn{1}{c|}{1.68} \\ 
\multicolumn{1}{|c|}{d-mSAC-$D_{\text{expert}}$}& 36.0\%± 5.9\% & \multicolumn{1}{c|}{0.5}& 39.7\%± 8.3\% & \multicolumn{1}{c|}{0.7}& 22.2\%± 3.2\% & \multicolumn{1}{c|}{1.8} \\ 
\multicolumn{1}{|c|}{mQRDQN-$D_{\text{noisy}}$}& 34.1\%± 4.8\% & \multicolumn{1}{c|}{0.44}& 29.3\%± 4.3\% & \multicolumn{1}{c|}{0.63}& 20.3\%± 2.4\% & \multicolumn{1}{c|}{1.64} \\ 
\multicolumn{1}{|c|}{d-mSAC-$D_{\text{noisy}}$}& 38.7\%± 9.4\% & \multicolumn{1}{c|}{0.51}& 51.6\%± 15.7\% & \multicolumn{1}{c|}{0.71}& 35.3\%± 9.6\% & \multicolumn{1}{c|}{1.8} \\ \hline 
\multicolumn{7}{|c|}{\textbf{Baselines} $15 \times 15$ } \\ \hline 
\multicolumn{1}{|c|}{L2D} & 32.3\%± 4.9\% & \multicolumn{1}{c|}{0.48}& 31.7\%± 6.4\% & \multicolumn{1}{c|}{0.68}& 21.0\%± 2.9\% & \multicolumn{1}{c|}{1.71}\\ 
\multicolumn{1}{|c|}{BC-$D_{\text{expert}}$} & 40.7\%± 6.4\% & \multicolumn{1}{c|}{0.5}& 44.4\%± 5.5\% & \multicolumn{1}{c|}{0.7}& 28.0\%± 4.2\% & \multicolumn{1}{c|}{1.77} \\ \hline 
\multicolumn{7}{|c|}{\textbf{Offline-LD} $15 \times 15$ } \\ \hline 
\multicolumn{1}{|c|}{mQRDQN-$D_{\text{expert}}$}& 30.5\%± 1.1\%${}^{*}$ & \multicolumn{1}{c|}{0.44}& 29.3\%± 6.2\% & \multicolumn{1}{c|}{0.64}& 20.9\%± 2.6\% & \multicolumn{1}{c|}{1.66} \\ 
\multicolumn{1}{|c|}{d-mSAC-$D_{\text{expert}}$}& 32.0\%± 2.8\% & \multicolumn{1}{c|}{0.49}& 28.1\%± 4.4\% & \multicolumn{1}{c|}{0.69}& 21.4\%± 2.4\% & \multicolumn{1}{c|}{1.78}\\ 
\multicolumn{1}{|c|}{mQRDQN-$D_{\text{noisy}}$}& 30.5\%± 0.7\%${}^{*}$ & \multicolumn{1}{c|}{0.45}& 29.2\%± 0.5\%${}^{*}$ & \multicolumn{1}{c|}{0.64}& 20.4\%± 0.7\% & \multicolumn{1}{c|}{1.67} \\ 
\multicolumn{1}{|c|}{d-mSAC-$D_{\text{noisy}}$}& 30.7\%± 2.4\% & \multicolumn{1}{c|}{0.5}& 28.7\%± 1.4\% & \multicolumn{1}{c|}{0.71}& 23.1\%± 4.6\% & \multicolumn{1}{c|}{1.76} \\ \hline 
\multicolumn{7}{|c|}{\textbf{Baselines} $20 \times 20$ } \\ \hline 
\multicolumn{1}{|c|}{L2D} & 32.7\%± 5.7\% & \multicolumn{1}{c|}{0.5}& 31.8\%± 6.7\% & \multicolumn{1}{c|}{0.7}& 21.0\%± 3.1\% & \multicolumn{1}{c|}{1.74} \\ 
\multicolumn{1}{|c|}{BC-$D_{\text{expert}}$} & 38.9\%± 4.9\% & \multicolumn{1}{c|}{0.5}& 41.2\%± 5.4\% & \multicolumn{1}{c|}{0.7}& 26.5\%± 4.3\% & \multicolumn{1}{c|}{1.78} \\ \hline 
\multicolumn{7}{|c|}{\textbf{Offline-LD} $20 \times 20$ } \\ \hline 
\multicolumn{1}{|c|}{mQRDQN-$D_{\text{expert}}$}& \textbf{30.0\%± 1.1\%}${}^{*}$ & \multicolumn{1}{c|}{0.44}& 29.0\%± 0.3\%${}^{*}$ & \multicolumn{1}{c|}{0.63}& 21.0\%± 1.0\% & \multicolumn{1}{c|}{1.68} \\ 
\multicolumn{1}{|c|}{d-mSAC-$D_{\text{expert}}$}& 30.4\%± 4.7\%${}^{*}$ & \multicolumn{1}{c|}{0.5}& 28.6\%± 3.1\% & \multicolumn{1}{c|}{0.71}& 21.8\%± 1.8\% & \multicolumn{1}{c|}{1.81} \\ 
\multicolumn{1}{|c|}{mQRDQN-$D_{\text{noisy}}$}& 30.2\%± 1.2\%${}^{*}$ & \multicolumn{1}{c|}{0.46}& 28.9\%± 1.4\%${}^{*}$ & \multicolumn{1}{c|}{0.64}& 20.6\%± 1.3\% & \multicolumn{1}{c|}{1.67} \\ 
\multicolumn{1}{|c|}{d-mSAC-$D_{\text{noisy}}$}& 31.5\%± 2.7\% & \multicolumn{1}{c|}{0.5}& \textbf{28.0\%± 3.9\%} & \multicolumn{1}{c|}{0.69}& 21.1\%± 2.4\% & \multicolumn{1}{c|}{1.8} \\ \hline 
\multicolumn{7}{|c|}{\textbf{Baselines} $30 \times 20$ } \\ \hline 
\multicolumn{1}{|c|}{L2D} & 30.6\%± 4.0\% & \multicolumn{1}{c|}{0.48}& 29.9\%± 3.4\% & \multicolumn{1}{c|}{0.68}& 20.0\%± 2.2\% & \multicolumn{1}{c|}{1.75} \\ 
\multicolumn{1}{|c|}{BC-$D_{\text{expert}}$} & 43.1\%± 5.5\% & \multicolumn{1}{c|}{0.48}& 44.7\%± 4.6\% & \multicolumn{1}{c|}{0.68}& 26.4\%± 3.1\% & \multicolumn{1}{c|}{1.72}\\ \hline 
\multicolumn{7}{|c|}{\textbf{Offline-LD} $30 \times 20$ } \\ \hline 
\multicolumn{1}{|c|}{mQRDQN-$D_{\text{expert}}$}& 30.7\%± 3.7\% & \multicolumn{1}{c|}{0.45}& 30.2\%± 5.8\% & \multicolumn{1}{c|}{0.65}& 21.6\%± 3.2\% & \multicolumn{1}{c|}{1.68} \\ 
\multicolumn{1}{|c|}{d-mSAC-$D_{\text{expert}}$}& 31.9\%± 4.6\% & \multicolumn{1}{c|}{0.5}& 29.4\%± 4.9\% & \multicolumn{1}{c|}{0.71}& 21.9\%± 3.1\% & \multicolumn{1}{c|}{1.78} \\ 
\multicolumn{1}{|c|}{mQRDQN-$D_{\text{noisy}}$}& 30.9\%± 2.2\% & \multicolumn{1}{c|}{0.45}& 28.9\%± 1.9\% & \multicolumn{1}{c|}{0.64}& 20.0\%± 1.0\% & \multicolumn{1}{c|}{1.67} \\ 
\multicolumn{1}{|c|}{d-mSAC-$D_{\text{noisy}}$}& 31.5\%± 2.6\% & \multicolumn{1}{c|}{0.5}& 28.6\%± 3.5\% & \multicolumn{1}{c|}{0.71}& 20.8\%± 2.0\% & \multicolumn{1}{c|}{1.79} \\ \hline 
\multicolumn{1}{|c|}{Opt. Rate (\%)} & \multicolumn{2}{c|}{90\%} & \multicolumn{2}{c|}{30\%} & \multicolumn{2}{c|}{100\%}  \\ \hline 
\end{tabular}

\label{tab:tai_other_small}
\end{table*}

\begin{table*}[ht]
\caption{The results for the Taillard instance for the following sizes: $50 \times 15$, $50 \times 20$, and $100 \times 20$. \textbf{Gap} is the difference between the obtained result and the upper bound. \textbf{Time} is the runtime in seconds. The \textbf{bold} result is the best RL approach. ${}^*$ signifies that an offline RL approach is a significant improvement compared to L2D of the same trained instance size, according to an independent t-test ($p=0.05$). The Opt. Rate (\%) is the percentage of the solutions that CP found an optimal result.}
\centering
\fontsize{6pt}{6pt}\selectfont
\setlength{\tabcolsep}{2mm}
\begin{tabular}{|ccccc|}
\hline
\multicolumn{1}{|c|}{} & \multicolumn{2}{|c|}{\begin{tabular}[c]{@{}c@{}}Taillard\\$50 \times 20$\end{tabular}}& \multicolumn{2}{|c|}{\begin{tabular}[c]{@{}c@{}}Taillard\\$100 \times 20$\end{tabular}} \\
\multicolumn{1}{|c|}{} & Gap & \multicolumn{1}{c|}{Time}& Gap & \multicolumn{1}{c|}{Time} \\ \hline 
\multicolumn{5}{|c|}{\textbf{PDR}} \\ \hline\multicolumn{1}{|c|}{SPT} & 55.3\% & \multicolumn{1}{c|}{0.49}& 39.8\% & \multicolumn{1}{c|}{4.08} \\ 
\multicolumn{1}{|c|}{MOR} & \textcolor{black}{45.7}\% & \multicolumn{1}{c|}{0.54}& \textcolor{black}{25.4}\% & \multicolumn{1}{c|}{4.46} \\ 
\multicolumn{1}{|c|}{MWKR} & \textcolor{black}{64.5}\% & \multicolumn{1}{c|}{0.59}& \textcolor{black}{49.9}\% & \multicolumn{1}{c|}{4.38} \\ \hline 
\multicolumn{5}{|c|}{\textbf{Baselines} $6 \times 6$ } \\ \hline 
\multicolumn{1}{|c|}{L2D} & 24.9\%± 3.3\% & \multicolumn{1}{c|}{2.83}& 13.4\%± 1.9\% & \multicolumn{1}{c|}{12.32} \\ 
\multicolumn{1}{|c|}{BC-$D_{\text{expert}}$} & 39.3\%± 15.5\% & \multicolumn{1}{c|}{2.82}& 22.2\%± 11.9\% & \multicolumn{1}{c|}{12.53} \\ \hline 
\multicolumn{5}{|c|}{\textbf{Offline-LD} $6 \times 6$ } \\ \hline 
\multicolumn{1}{|c|}{mQRDQN-$D_{\text{expert}}$}& 26.6\%± 2.4\% & \multicolumn{1}{c|}{2.73}& 12.7\%± 1.2\% & \multicolumn{1}{c|}{12.91} \\ 
\multicolumn{1}{|c|}{d-mSAC-$D_{\text{expert}}$} & 30.4\%± 3.7\% & \multicolumn{1}{c|}{2.77}& 14.4\%± 2.0\% & \multicolumn{1}{c|}{13.03} \\ 
\multicolumn{1}{|c|}{mQRDQN-$D_{\text{noisy}}$} & 24.4\%± 2.1\% & \multicolumn{1}{c|}{2.6}& 12.5\%± 1.1\% & \multicolumn{1}{c|}{12.82} \\ 
\multicolumn{1}{|c|}{d-mSAC-$D_{\text{noisy}}$} & 45.1\%± 9.9\% & \multicolumn{1}{c|}{2.87}& 21.8\%± 7.4\% & \multicolumn{1}{c|}{12.0} \\ \hline 
\multicolumn{5}{|c|}{\textbf{Baselines} $10 \times 10$ } \\ \hline 
\multicolumn{1}{|c|}{L2D} & 27.2\%± 4.8\% & \multicolumn{1}{c|}{2.7}& 13.8\%± 2.4\% & \multicolumn{1}{c|}{12.17} \\ 
\multicolumn{1}{|c|}{BC-$D_{\text{expert}}$} & 38.7\%± 8.0\% & \multicolumn{1}{c|}{2.91}& 21.5\%± 7.5\% & \multicolumn{1}{c|}{11.64} \\ \hline 
\multicolumn{5}{|c|}{\textbf{Offline-LD} $10 \times 10$ } \\ \hline 
\multicolumn{1}{|c|}{mQRDQN-$D_{\text{expert}}$}& 27.5\%± 2.9\% & \multicolumn{1}{c|}{2.8}& 13.2\%± 1.6\% & \multicolumn{1}{c|}{12.62} \\ 
\multicolumn{1}{|c|}{d-mSAC-$D_{\text{expert}}$} & 29.9\%± 5.9\% & \multicolumn{1}{c|}{2.96}& 14.0\%± 2.7\% & \multicolumn{1}{c|}{12.68} \\ 
\multicolumn{1}{|c|}{mQRDQN-$D_{\text{noisy}}$} & 25.8\%± 3.0\% & \multicolumn{1}{c|}{2.77}& 13.1\%± 1.6\% & \multicolumn{1}{c|}{12.12} \\ 
\multicolumn{1}{|c|}{d-mSAC-$D_{\text{noisy}}$} & 44.8\%± 12.8\% & \multicolumn{1}{c|}{2.88}& 27.8\%± 8.2\% & \multicolumn{1}{c|}{11.9} \\ \hline 
\multicolumn{5}{|c|}{\textbf{Baselines} $15 \times 15$ } \\ \hline 
\multicolumn{1}{|c|}{L2D} & 25.6\%± 3.4\% & \multicolumn{1}{c|}{2.74}& 12.9\%± 1.3\% & \multicolumn{1}{c|}{11.92} \\ 
\multicolumn{1}{|c|}{BC-$D_{\text{expert}}$} & 40.1\%± 7.0\% & \multicolumn{1}{c|}{2.86}& 23.0\%± 6.6\% & \multicolumn{1}{c|}{12.11} \\ \hline 
\multicolumn{5}{|c|}{\textbf{Offline-LD} $15 \times 15$ } \\ \hline 
\multicolumn{1}{|c|}{mQRDQN-$D_{\text{expert}}$} & 23.9\%± 3.8\% & \multicolumn{1}{c|}{2.71}& 12.6\%± 1.5\% & \multicolumn{1}{c|}{12.68} \\ 
\multicolumn{1}{|c|}{d-mSAC-$D_{\text{expert}}$}& 24.5\%± 3.2\% & \multicolumn{1}{c|}{2.83}& 12.6\%± 1.5\% & \multicolumn{1}{c|}{12.66} \\ 
\multicolumn{1}{|c|}{mQRDQN-$D_{\text{noisy}}$} & 24.2\%± 1.1\%${}^{*}$ & \multicolumn{1}{c|}{2.69}& 13.0\%± 1.2\% & \multicolumn{1}{c|}{12.55} \\ 
\multicolumn{1}{|c|}{d-mSAC-$D_{\text{noisy}}$}& 27.6\%± 7.7\% & \multicolumn{1}{c|}{2.86}& 17.9\%± 7.8\% & \multicolumn{1}{c|}{12.58} \\ \hline 
\multicolumn{5}{|c|}{\textbf{Baselines} $20 \times 20$ } \\ \hline 
\multicolumn{1}{|c|}{L2D} & 26.1\%± 4.6\% & \multicolumn{1}{c|}{2.77}& 13.3\%± 1.8\% & \multicolumn{1}{c|}{12.11} \\ 
\multicolumn{1}{|c|}{BC-$D_{\text{expert}}$} & 34.4\%± 6.2\% & \multicolumn{1}{c|}{2.91}& 20.8\%± 5.8\% & \multicolumn{1}{c|}{12.8} \\ \hline 
\multicolumn{5}{|c|}{\textbf{Offline-LD} $20 \times 20$ } \\ \hline 
\multicolumn{1}{|c|}{mQRDQN-$D_{\text{expert}}$}& \textbf{23.7\%± 1.2\%}${}^{*}$ & \multicolumn{1}{c|}{2.68}& 12.9\%± 1.2\% & \multicolumn{1}{c|}{12.76} \\ 
\multicolumn{1}{|c|}{d-mSAC-$D_{\text{expert}}$}& 24.0\%± 2.0\% & \multicolumn{1}{c|}{2.99}& \textbf{12.4\%± 1.4\%} & \multicolumn{1}{c|}{13.33} \\ 
\multicolumn{1}{|c|}{mQRDQN-$D_{\text{noisy}}$}& 24.3\%± 1.8\%${}^{*}$ & \multicolumn{1}{c|}{2.75}& 12.7\%± 1.0\% & \multicolumn{1}{c|}{12.28} \\ 
\multicolumn{1}{|c|}{d-mSAC-$D_{\text{noisy}}$}& 23.5\%± 2.6\%${}^{*}$ & \multicolumn{1}{c|}{2.93}& 13.7\%± 2.7\% & \multicolumn{1}{c|}{13.23} \\ \hline 
\multicolumn{5}{|c|}{\textbf{Baselines} $30 \times 20$ } \\ \hline 
\multicolumn{1}{|c|}{L2D} & 24.5\%± 1.7\% & \multicolumn{1}{c|}{2.81}& 12.7\%± 0.8\% & \multicolumn{1}{c|}{12.55} \\ 
\multicolumn{1}{|c|}{BC-$D_{\text{expert}}$} & 36.7\%± 3.7\% & \multicolumn{1}{c|}{2.79}& 20.7\%± 4.5\% & \multicolumn{1}{c|}{12.31} \\ \hline 
\multicolumn{5}{|c|}{\textbf{Offline-LD} $30 \times 20$ } \\ \hline 
\multicolumn{1}{|c|}{mQRDQN-$D_{\text{expert}}$}& 24.9\%± 3.1\% & \multicolumn{1}{c|}{2.79}& 13.2\%± 2.0\% & \multicolumn{1}{c|}{12.66} \\ 
\multicolumn{1}{|c|}{d-mSAC-$D_{\text{expert}}$}& 23.8\%± 2.9\% & \multicolumn{1}{c|}{2.93}& 12.9\%± 1.6\% & \multicolumn{1}{c|}{13.16} \\ 
\multicolumn{1}{|c|}{mQRDQN-$D_{\text{noisy}}$}& 24.3\%± 1.0\% & \multicolumn{1}{c|}{2.78}& 12.7\%± 0.9\% & \multicolumn{1}{c|}{13.09} \\ 
\multicolumn{1}{|c|}{d-mSAC-$D_{\text{noisy}}$} & 23.9\%± 2.7\% & \multicolumn{1}{c|}{2.89}& 12.7\%± 1.6\% & \multicolumn{1}{c|}{12.73} \\ \hline 
\multicolumn{1}{|c|}{Opt. Rate (\%)} & \multicolumn{2}{c|}{100\%} & \multicolumn{2}{c|}{100\%}  \\ \hline 
\end{tabular}

\label{tab:tai_other_large}
\end{table*}
\begin{table*}[ht]
\caption{The results for the Demirkol instance for the following sizes: $20 \times 20$, and $30 \times 20$. \textbf{Gap} is the difference between the obtained result and the upper bound. \textbf{Time} is the runtime in seconds. The \textbf{bold} result is the best RL approach. ${}^*$ signifies that an offline RL approach is a significant improvement compared to L2D of the same trained instance size, according to an independent t-test ($p=0.05$). The Opt. Rate (\%) is the percentage of the solutions that CP found an optimal result.}
\centering
\fontsize{6pt}{6pt}\selectfont
\setlength{\tabcolsep}{2mm}
\begin{tabular}{|ccccc|}
\hline
\multicolumn{1}{|c|}{} & \multicolumn{2}{|c|}{\begin{tabular}[c]{@{}c@{}}Demirkol\\$20 \times 20$\end{tabular}}& \multicolumn{2}{|c|}{\begin{tabular}[c]{@{}c@{}}Demirkol\\$30 \times 20$\end{tabular}} \\
\multicolumn{1}{|c|}{} & Gap & \multicolumn{1}{c|}{Time}& Gap & \multicolumn{1}{c|}{Time} \\ \hline 
\multicolumn{5}{|c|}{\textbf{PDR}} \\ \hline\multicolumn{1}{|c|}{SPT} & 64.8\% & \multicolumn{1}{c|}{0.05}& 62.2\% & \multicolumn{1}{c|}{0.1} \\ 
\multicolumn{1}{|c|}{MOR} & \textcolor{black}{58.1}\% & \multicolumn{1}{c|}{0.05}& \textcolor{black}{64.2}\% & \multicolumn{1}{c|}{0.09} \\ 
\multicolumn{1}{|c|}{MWKR} & \textcolor{black}{70.2}\% & \multicolumn{1}{c|}{0.05}& \textcolor{black}{89.7}\% & \multicolumn{1}{c|}{0.14} \\ \hline 
\multicolumn{5}{|c|}{\textbf{Baselines} $6 \times 6$ } \\ \hline 
\multicolumn{1}{|c|}{L2D} & 34.5\%± 3.9\% & \multicolumn{1}{c|}{0.72}& 37.4\%± 3.7\% & \multicolumn{1}{c|}{1.28} \\ 
\multicolumn{1}{|c|}{BC-$D_{\text{expert}}$} & 46.0\%± 14.7\% & \multicolumn{1}{c|}{0.69}& 47.3\%± 13.3\% & \multicolumn{1}{c|}{1.26} \\ \hline 
\multicolumn{5}{|c|}{\textbf{Offline-LD} $6 \times 6$ } \\ \hline 
\multicolumn{1}{|c|}{mQRDQN-$D_{\text{expert}}$}& 35.3\%± 2.6\% & \multicolumn{1}{c|}{0.64}& 38.0\%± 3.7\% & \multicolumn{1}{c|}{1.17} \\ 
\multicolumn{1}{|c|}{d-mSAC-$D_{\text{expert}}$}& 38.3\%± 5.5\% & \multicolumn{1}{c|}{0.7}& 39.5\%± 5.5\% & \multicolumn{1}{c|}{1.26} \\ 
\multicolumn{1}{|c|}{mQRDQN-$D_{\text{noisy}}$}& 34.5\%± 3.1\% & \multicolumn{1}{c|}{0.62}& 37.3\%± 3.0\% & \multicolumn{1}{c|}{1.16} \\ 
\multicolumn{1}{|c|}{d-mSAC-$D_{\text{noisy}}$}& 48.1\%± 11.5\% & \multicolumn{1}{c|}{0.7}& 50.1\%± 10.8\% & \multicolumn{1}{c|}{1.28} \\ \hline 
\multicolumn{5}{|c|}{\textbf{Baselines} $10 \times 10$ } \\ \hline 
\multicolumn{1}{|c|}{L2D} & 35.0\%± 5.4\% & \multicolumn{1}{c|}{0.67}& 37.6\%± 5.9\% & \multicolumn{1}{c|}{1.2} \\ 
\multicolumn{1}{|c|}{BC-$D_{\text{expert}}$} & 46.5\%± 10.0\% & \multicolumn{1}{c|}{0.68}& 48.4\%± 8.6\% & \multicolumn{1}{c|}{1.24} \\ \hline 
\multicolumn{5}{|c|}{\textbf{Offline-LD} $10 \times 10$ } \\ \hline 
\multicolumn{1}{|c|}{mQRDQN-$D_{\text{expert}}$}& 34.8\%± 3.5\% & \multicolumn{1}{c|}{0.63}& 38.0\%± 3.3\% & \multicolumn{1}{c|}{1.16} \\ 
\multicolumn{1}{|c|}{d-mSAC-$D_{\text{expert}}$}& 40.3\%± 4.6\% & \multicolumn{1}{c|}{0.69}& 40.6\%± 3.6\% & \multicolumn{1}{c|}{1.27} \\ 
\multicolumn{1}{|c|}{mQRDQN-$D_{\text{noisy}}$}& 34.3\%± 4.1\% & \multicolumn{1}{c|}{0.63}& 37.2\%± 3.6\% & \multicolumn{1}{c|}{1.16} \\ 
\multicolumn{1}{|c|}{d-mSAC-$D_{\text{noisy}}$}& 53.9\%± 11.3\% & \multicolumn{1}{c|}{0.71}& 57.0\%± 10.2\% & \multicolumn{1}{c|}{1.29} \\ \hline 
\multicolumn{5}{|c|}{\textbf{Baselines} $15 \times 15$ } \\ \hline 
\multicolumn{1}{|c|}{L2D} & 34.4\%± 4.0\% & \multicolumn{1}{c|}{0.68}& 37.4\%± 4.1\% & \multicolumn{1}{c|}{1.22} \\ 
\multicolumn{1}{|c|}{BC-$D_{\text{expert}}$} & 47.9\%± 8.8\% & \multicolumn{1}{c|}{0.7}& 51.1\%± 8.3\% & \multicolumn{1}{c|}{1.27} \\ \hline 
\multicolumn{5}{|c|}{\textbf{Offline-LD} $15 \times 15$ } \\ \hline 
\multicolumn{1}{|c|}{mQRDQN-$D_{\text{expert}}$}& 33.6\%± 4.4\% & \multicolumn{1}{c|}{0.63}& 38.1\%± 4.2\% & \multicolumn{1}{c|}{1.16} \\ 
\multicolumn{1}{|c|}{d-mSAC-$D_{\text{expert}}$}& 36.0\%± 5.2\% & \multicolumn{1}{c|}{0.7}& 39.2\%± 4.7\% & \multicolumn{1}{c|}{1.25} \\ 
\multicolumn{1}{|c|}{mQRDQN-$D_{\text{noisy}}$}& 33.6\%± 2.5\% & \multicolumn{1}{c|}{0.64}& 36.6\%± 2.9\% & \multicolumn{1}{c|}{1.19} \\ 
\multicolumn{1}{|c|}{d-mSAC-$D_{\text{noisy}}$}& 38.8\%± 9.3\% & \multicolumn{1}{c|}{0.72}& 43.5\%± 9.0\% & \multicolumn{1}{c|}{1.29} \\ \hline 
\multicolumn{5}{|c|}{\textbf{Baselines} $20 \times 20$ } \\ \hline 
\multicolumn{1}{|c|}{L2D} & 34.4\%± 3.6\% & \multicolumn{1}{c|}{0.7}& 38.0\%± 5.8\% & \multicolumn{1}{c|}{1.25} \\ 
\multicolumn{1}{|c|}{BC-$D_{\text{expert}}$} & 42.8\%± 7.8\% & \multicolumn{1}{c|}{0.71}& 48.6\%± 9.3\% & \multicolumn{1}{c|}{1.28} \\ \hline 
\multicolumn{5}{|c|}{\textbf{Offline-LD} $20 \times 20$ } \\ \hline 
\multicolumn{1}{|c|}{mQRDQN-$D_{\text{expert}}$}& 32.8\%± 4.0\% & \multicolumn{1}{c|}{0.63}& 36.8\%± 2.7\% & \multicolumn{1}{c|}{1.16} \\ 
\multicolumn{1}{|c|}{d-mSAC-$D_{\text{expert}}$}& \textbf{31.6\%± 4.2\%} & \multicolumn{1}{c|}{0.7}& 36.7\%± 4.1\% & \multicolumn{1}{c|}{1.27} \\ 
\multicolumn{1}{|c|}{mQRDQN-$D_{\text{noisy}}$}& 32.8\%± 2.5\%${}^{*}$ & \multicolumn{1}{c|}{0.64}& 36.0\%± 2.5\% & \multicolumn{1}{c|}{1.2} \\ 
\multicolumn{1}{|c|}{d-mSAC-$D_{\text{noisy}}$}& 33.4\%± 3.6\% & \multicolumn{1}{c|}{0.7}& 38.5\%± 4.3\% & \multicolumn{1}{c|}{1.28} \\ \hline 
\multicolumn{5}{|c|}{\textbf{Baselines} $30 \times 20$ } \\ \hline 
\multicolumn{1}{|c|}{L2D} & 33.5\%± 2.1\% & \multicolumn{1}{c|}{0.68}& 36.7\%± 1.8\% & \multicolumn{1}{c|}{1.24} \\ 
\multicolumn{1}{|c|}{BC-$D_{\text{expert}}$} & 43.0\%± 4.4\% & \multicolumn{1}{c|}{0.69}& 47.9\%± 5.2\% & \multicolumn{1}{c|}{1.23} \\ \hline 
\multicolumn{5}{|c|}{\textbf{Offline-LD} $30 \times 20$ } \\ \hline 
\multicolumn{1}{|c|}{mQRDQN-$D_{\text{expert}}$}& 37.1\%± 6.6\% & \multicolumn{1}{c|}{0.64}& 37.8\%± 4.1\% & \multicolumn{1}{c|}{1.2} \\ 
\multicolumn{1}{|c|}{d-mSAC-$D_{\text{expert}}$}& 34.2\%± 5.2\% & \multicolumn{1}{c|}{0.7}& 37.6\%± 3.2\% & \multicolumn{1}{c|}{1.26} \\ 
\multicolumn{1}{|c|}{mQRDQN-$D_{\text{noisy}}$}& 32.9\%± 1.5\% & \multicolumn{1}{c|}{0.64}&\textbf{35.8\%± 2.4\%} & \multicolumn{1}{c|}{1.19} \\ 
\multicolumn{1}{|c|}{d-mSAC-$D_{\text{noisy}}$}& 33.7\%± 6.8\% & \multicolumn{1}{c|}{0.72}& 41.2\%± 8.9\% & \multicolumn{1}{c|}{1.27} \\ \hline 
\multicolumn{1}{|c|}{Opt. Rate (\%)} & \multicolumn{2}{c|}{0\%} & \multicolumn{2}{c|}{10\%}  \\ \hline 
\end{tabular}

\label{tab:dmu_trained}
\end{table*}
\begin{table*}[ht]
\caption{The results for the Demirkol instance for the following sizes: $20 \times 15$, $30 \times 15$, and $40 \times 15$. \textbf{Gap} is the difference between the obtained result and the upper bound. \textbf{Time} is the runtime in seconds. The \textbf{bold} result is the best RL approach. ${}^*$ signifies that an offline RL approach is a significant improvement compared to L2D of the same trained instance size, according to an independent t-test ($p=0.05$). The Opt. Rate (\%) is the percentage of the solutions that CP found an optimal result.}
\centering
\fontsize{6pt}{6pt}\selectfont
\setlength{\tabcolsep}{2mm}
\begin{tabular}{|ccccccc|}
\hline
\multicolumn{1}{|c|}{} & \multicolumn{2}{|c|}{\begin{tabular}[c]{@{}c@{}}Demirkol\\$20 \times 15$\end{tabular}}& \multicolumn{2}{|c|}{\begin{tabular}[c]{@{}c@{}}Demirkol\\$30 \times 15$\end{tabular}}& \multicolumn{2}{|c|}{\begin{tabular}[c]{@{}c@{}}Demirkol\\$40 \times 15$\end{tabular}} \\
\multicolumn{1}{|c|}{} & Gap & \multicolumn{1}{c|}{Time}& Gap & \multicolumn{1}{c|}{Time}& Gap & \multicolumn{1}{c|}{Time} \\ \hline 
\multicolumn{7}{|c|}{\textbf{PDR}} \\ \hline\multicolumn{1}{|c|}{SPT} & 61.5\% & \multicolumn{1}{c|}{0.03}& 60.5\% & \multicolumn{1}{c|}{0.05}& 57.1\% & \multicolumn{1}{c|}{0.09} \\ 
\multicolumn{1}{|c|}{MOR} & \textcolor{black}{62.4}\% & \multicolumn{1}{c|}{0.03}& \textcolor{black}{63.3}\% & \multicolumn{1}{c|}{0.06}& \textcolor{black}{58.6}\% & \multicolumn{1}{c|}{0.1} \\ 
\multicolumn{1}{|c|}{MWKR} & \textcolor{black}{75.9}\% & \multicolumn{1}{c|}{0.06}& \textcolor{black}{78.6}\% & \multicolumn{1}{c|}{0.07}& \textcolor{black}{84.3}\% & \multicolumn{1}{c|}{0.12} \\ \hline 
\multicolumn{7}{|c|}{\textbf{Baselines} $6 \times 6$ } \\ \hline 
\multicolumn{1}{|c|}{L2D} & 35.5\%± 5.0\% & \multicolumn{1}{c|}{0.53}& 38.2\%± 3.9\% & \multicolumn{1}{c|}{0.85}& \textbf{34.4\%± 4.0\%} & \multicolumn{1}{c|}{1.28} \\ 
\multicolumn{1}{|c|}{BC-$D_{\text{expert}}$} & 45.5\%± 13.5\% & \multicolumn{1}{c|}{0.5}& 45.4\%± 11.1\% & \multicolumn{1}{c|}{0.8}& 40.3\%± 11.6\% & \multicolumn{1}{c|}{1.25} \\ \hline 
\multicolumn{7}{|c|}{\textbf{Offline-LD} $6 \times 6$ } \\ \hline 
\multicolumn{1}{|c|}{mQRDQN-$D_{\text{expert}}$}& 36.4\%± 4.8\% & \multicolumn{1}{c|}{0.46}& 40.7\%± 2.8\% & \multicolumn{1}{c|}{0.75}& 36.2\%± 2.3\% & \multicolumn{1}{c|}{1.19} \\ 
\multicolumn{1}{|c|}{d-mSAC-$D_{\text{expert}}$}& 36.8\%± 4.0\% & \multicolumn{1}{c|}{0.51}& 41.6\%± 5.4\% & \multicolumn{1}{c|}{0.8}& 36.3\%± 4.9\% & \multicolumn{1}{c|}{1.27} \\ 
\multicolumn{1}{|c|}{mQRDQN-$D_{\text{noisy}}$}& 34.9\%± 3.1\% & \multicolumn{1}{c|}{0.45}& 40.5\%± 2.6\% & \multicolumn{1}{c|}{0.74}& 35.8\%± 2.4\% & \multicolumn{1}{c|}{1.15} \\ 
\multicolumn{1}{|c|}{d-mSAC-$D_{\text{noisy}}$}& 45.5\%± 7.2\% & \multicolumn{1}{c|}{0.52}& 46.9\%± 8.6\% & \multicolumn{1}{c|}{0.81}& 40.4\%± 8.3\% & \multicolumn{1}{c|}{1.27} \\ \hline 
\multicolumn{7}{|c|}{\textbf{Baselines} $10 \times 10$ } \\ \hline 
\multicolumn{1}{|c|}{L2D} & 35.2\%± 4.7\% & \multicolumn{1}{c|}{0.49}& 37.7\%± 5.8\% & \multicolumn{1}{c|}{0.79}& 34.5\%± 5.1\% & \multicolumn{1}{c|}{1.2} \\ 
\multicolumn{1}{|c|}{BC-$D_{\text{expert}}$} & 42.5\%± 8.2\% & \multicolumn{1}{c|}{0.49}& 44.3\%± 7.3\% & \multicolumn{1}{c|}{0.79}& 40.0\%± 7.2\% & \multicolumn{1}{c|}{1.24} \\ \hline 
\multicolumn{7}{|c|}{\textbf{Offline-LD} $10 \times 10$ } \\ \hline 
\multicolumn{1}{|c|}{mQRDQN-$D_{\text{expert}}$}& 36.1\%± 3.3\% & \multicolumn{1}{c|}{0.45}& 41.3\%± 3.4\% & \multicolumn{1}{c|}{0.74}& 35.2\%± 3.0\% & \multicolumn{1}{c|}{1.16} \\ 
\multicolumn{1}{|c|}{d-mSAC-$D_{\text{expert}}$}& 37.4\%± 4.8\% & \multicolumn{1}{c|}{0.51}& 40.7\%± 5.4\% & \multicolumn{1}{c|}{0.82}& 34.7\%± 3.2\% & \multicolumn{1}{c|}{1.26} \\ 
\multicolumn{1}{|c|}{mQRDQN-$D_{\text{noisy}}$}& \textbf{33.5\%± 3.6\%} & \multicolumn{1}{c|}{0.45}& 39.6\%± 3.1\% & \multicolumn{1}{c|}{0.74}& 34.8\%± 3.0\% & \multicolumn{1}{c|}{1.16} \\ 
\multicolumn{1}{|c|}{d-mSAC-$D_{\text{noisy}}$}& 50.0\%± 9.9\% & \multicolumn{1}{c|}{0.52}& 51.6\%± 9.4\% & \multicolumn{1}{c|}{0.83}& 47.4\%± 8.1\% & \multicolumn{1}{c|}{1.29} \\ \hline 
\multicolumn{7}{|c|}{\textbf{Baselines} $15 \times 15$ } \\ \hline 
\multicolumn{1}{|c|}{L2D} & 36.0\%± 4.2\% & \multicolumn{1}{c|}{0.5}& 37.2\%± 3.1\% & \multicolumn{1}{c|}{0.81}& 34.6\%± 2.5\% & \multicolumn{1}{c|}{1.22} \\ 
\multicolumn{1}{|c|}{BC-$D_{\text{expert}}$} & 45.7\%± 8.4\% & \multicolumn{1}{c|}{0.51}& 46.4\%± 7.3\% & \multicolumn{1}{c|}{0.82}& 40.3\%± 7.3\% & \multicolumn{1}{c|}{1.26} \\ \hline 
\multicolumn{7}{|c|}{\textbf{Offline-LD} $15 \times 15$ } \\ \hline 
\multicolumn{1}{|c|}{mQRDQN-$D_{\text{expert}}$}& 34.6\%± 4.5\% & \multicolumn{1}{c|}{0.46}& 38.9\%± 4.4\% & \multicolumn{1}{c|}{0.74}& 35.4\%± 3.0\% & \multicolumn{1}{c|}{1.19} \\ 
\multicolumn{1}{|c|}{d-mSAC-$D_{\text{expert}}$}& 34.9\%± 5.7\% & \multicolumn{1}{c|}{0.5}& 38.5\%± 4.7\% & \multicolumn{1}{c|}{0.82}& 35.6\%± 4.4\% & \multicolumn{1}{c|}{1.26} \\ 
\multicolumn{1}{|c|}{mQRDQN-$D_{\text{noisy}}$}& 35.7\%± 1.6\% & \multicolumn{1}{c|}{0.47}& 39.1\%± 2.0\% & \multicolumn{1}{c|}{0.75}& 35.7\%± 2.0\% & \multicolumn{1}{c|}{1.18} \\ 
\multicolumn{1}{|c|}{d-mSAC-$D_{\text{noisy}}$}& 40.8\%± 8.9\% & \multicolumn{1}{c|}{0.51}& 44.5\%± 9.0\% & \multicolumn{1}{c|}{0.83}& 40.6\%± 8.2\% & \multicolumn{1}{c|}{1.29} \\ \hline 
\multicolumn{7}{|c|}{\textbf{Baselines} $20 \times 20$ } \\ \hline 
\multicolumn{1}{|c|}{L2D} & 36.3\%± 4.2\% & \multicolumn{1}{c|}{0.51}& 37.8\%± 4.4\% & \multicolumn{1}{c|}{0.82}& 34.6\%± 4.3\% & \multicolumn{1}{c|}{1.25} \\ 
\multicolumn{1}{|c|}{BC-$D_{\text{expert}}$} & 42.6\%± 6.9\% & \multicolumn{1}{c|}{0.51}& 43.9\%± 6.9\% & \multicolumn{1}{c|}{0.82}& 38.7\%± 5.9\% & \multicolumn{1}{c|}{1.28} \\ \hline 
\multicolumn{7}{|c|}{\textbf{Offline-LD} $20 \times 20$ } \\ \hline 
\multicolumn{1}{|c|}{mQRDQN-$D_{\text{expert}}$}& 35.3\%± 4.1\% & \multicolumn{1}{c|}{0.46}& \textbf{37.0\%± 3.9\%} & \multicolumn{1}{c|}{0.75}& 34.7\%± 2.8\% & \multicolumn{1}{c|}{1.16} \\ 
\multicolumn{1}{|c|}{d-mSAC-$D_{\text{expert}}$}& 35.6\%± 3.7\% & \multicolumn{1}{c|}{0.51}& 37.4\%± 3.4\% & \multicolumn{1}{c|}{0.83}& 34.5\%± 2.9\% & \multicolumn{1}{c|}{1.27} \\ 
\multicolumn{1}{|c|}{mQRDQN-$D_{\text{noisy}}$}& 35.8\%± 2.4\% & \multicolumn{1}{c|}{0.46}& 38.8\%± 2.3\% & \multicolumn{1}{c|}{0.75}& 35.5\%± 2.2\% & \multicolumn{1}{c|}{1.2} \\ 
\multicolumn{1}{|c|}{d-mSAC-$D_{\text{noisy}}$}& 35.7\%± 3.9\% & \multicolumn{1}{c|}{0.51}& 39.9\%± 5.1\% & \multicolumn{1}{c|}{0.82}& 36.3\%± 3.2\% & \multicolumn{1}{c|}{1.28} \\ \hline 
\multicolumn{7}{|c|}{\textbf{Baselines} $30 \times 20$ } \\ \hline 
\multicolumn{1}{|c|}{L2D} & 35.1\%± 1.2\% & \multicolumn{1}{c|}{0.5}& 37.5\%± 1.9\% & \multicolumn{1}{c|}{0.81}& 34.9\%± 1.3\% & \multicolumn{1}{c|}{1.24} \\ 
\multicolumn{1}{|c|}{BC-$D_{\text{expert}}$} & 41.3\%± 6.1\% & \multicolumn{1}{c|}{0.49}& 43.2\%± 5.1\% & \multicolumn{1}{c|}{0.79}& 37.6\%± 5.1\% & \multicolumn{1}{c|}{1.24} \\ \hline 
\multicolumn{7}{|c|}{\textbf{Offline-LD} $30 \times 20$ } \\ \hline 
\multicolumn{1}{|c|}{mQRDQN-$D_{\text{expert}}$}& 36.1\%± 4.4\% & \multicolumn{1}{c|}{0.47}& 40.3\%± 5.0\% & \multicolumn{1}{c|}{0.76}& 34.6\%± 4.8\% & \multicolumn{1}{c|}{1.19} \\ 
\multicolumn{1}{|c|}{d-mSAC-$D_{\text{expert}}$}& 34.6\%± 4.4\% & \multicolumn{1}{c|}{0.51}& 38.6\%± 4.5\% & \multicolumn{1}{c|}{0.82}& 34.7\%± 3.5\% & \multicolumn{1}{c|}{1.29} \\ 
\multicolumn{1}{|c|}{mQRDQN-$D_{\text{noisy}}$}& 35.3\%± 2.1\% & \multicolumn{1}{c|}{0.46}& 38.1\%± 1.9\% & \multicolumn{1}{c|}{0.75}& 35.6\%± 2.3\% & \multicolumn{1}{c|}{1.2} \\ 
\multicolumn{1}{|c|}{d-mSAC-$D_{\text{noisy}}$}& 36.5\%± 5.3\% & \multicolumn{1}{c|}{0.52}& 39.6\%± 5.2\% & \multicolumn{1}{c|}{0.84}& 36.9\%± 5.0\% & \multicolumn{1}{c|}{1.29} \\ \hline 
\multicolumn{1}{|c|}{Opt. Rate (\%)} & \multicolumn{2}{c|}{20\%} & \multicolumn{2}{c|}{30\%} & \multicolumn{2}{c|}{50\%}  \\ \hline 
\end{tabular}

\label{tab:dmu_small}
\end{table*}
\begin{table*}[ht]
\caption{The results for the Demirkol instance for the following sizes: $40 \times 20$, $50 \times 15$, and $50 \times 20$. \textbf{Gap} is the difference between the obtained result and the upper bound. \textbf{Time} is the runtime in seconds. The \textbf{bold} result is the best RL approach. ${}^*$ signifies that an offline RL approach is a significant improvement compared to L2D of the same trained instance size, according to an independent t-test ($p=0.05$). The Opt. Rate (\%) is the percentage of the solutions that CP found an optimal result.}
\centering

\fontsize{6pt}{6pt}\selectfont
\setlength{\tabcolsep}{2mm}
\begin{tabular}{|ccccccc|}
\hline
\multicolumn{1}{|c|}{} & \multicolumn{2}{|c|}{\begin{tabular}[c]{@{}c@{}}Demirkol\\$40 \times 20$\end{tabular}}& \multicolumn{2}{|c|}{\begin{tabular}[c]{@{}c@{}}Demirkol\\$50 \times 15$\end{tabular}}& \multicolumn{2}{|c|}{\begin{tabular}[c]{@{}c@{}}Demirkol\\$50 \times 20$\end{tabular}} \\
\multicolumn{1}{|c|}{} & Gap & \multicolumn{1}{c|}{Time}& Gap & \multicolumn{1}{c|}{Time}& Gap & \multicolumn{1}{c|}{Time} \\ \hline 
\multicolumn{7}{|c|}{\textbf{PDR}} \\ \hline\multicolumn{1}{|c|}{SPT} & 61.8\% & \multicolumn{1}{c|}{0.27}& 50.4\% & \multicolumn{1}{c|}{0.23}& 58.5\% & \multicolumn{1}{c|}{0.49} \\ 
\multicolumn{1}{|c|}{MOR} & \textcolor{black}{64.4}\% & \multicolumn{1}{c|}{0.29}& \textcolor{black}{54.5}\% & \multicolumn{1}{c|}{0.24}& \textcolor{black}{63.9}\% & \multicolumn{1}{c|}{0.51} \\ 
\multicolumn{1}{|c|}{MWKR} & \textcolor{black}{85.3}\% & \multicolumn{1}{c|}{0.38}& \textcolor{black}{83.9}\% & \multicolumn{1}{c|}{0.29}& \textcolor{black}{92.3}\% & \multicolumn{1}{c|}{0.6} \\ \hline 
\multicolumn{7}{|c|}{\textbf{Baselines} $6 \times 6$ } \\ \hline 
\multicolumn{1}{|c|}{L2D} & 39.1\%± 3.5\% & \multicolumn{1}{c|}{1.95}& 33.7\%± 4.3\% & \multicolumn{1}{c|}{1.78}& 37.6\%± 4.2\% & \multicolumn{1}{c|}{2.84} \\ 
\multicolumn{1}{|c|}{BC-$D_{\text{expert}}$} & 45.5\%± 12.9\% & \multicolumn{1}{c|}{1.93}& 36.9\%± 10.2\% & \multicolumn{1}{c|}{1.75}& 42.5\%± 11.7\% & \multicolumn{1}{c|}{2.8} \\ \hline 
\multicolumn{7}{|c|}{\textbf{Offline-LD} $6 \times 6$ } \\ \hline 
\multicolumn{1}{|c|}{mQRDQN-$D_{\text{expert}}$}& 37.5\%± 3.1\% & \multicolumn{1}{c|}{1.84}& 35.5\%± 1.7\% & \multicolumn{1}{c|}{1.66}& 37.6\%± 2.0\% & \multicolumn{1}{c|}{2.74} \\ 
\multicolumn{1}{|c|}{d-mSAC-$D_{\text{expert}}$}& 37.7\%± 5.3\% & \multicolumn{1}{c|}{1.93}& 33.3\%± 4.9\% & \multicolumn{1}{c|}{1.73}& 36.7\%± 4.5\% & \multicolumn{1}{c|}{2.78} \\ 
\multicolumn{1}{|c|}{mQRDQN-$D_{\text{noisy}}$}& 37.7\%± 2.4\% & \multicolumn{1}{c|}{1.83}& 36.2\%± 1.6\% & \multicolumn{1}{c|}{1.65}& 38.1\%± 2.4\% & \multicolumn{1}{c|}{2.61} \\ 
\multicolumn{1}{|c|}{d-mSAC-$D_{\text{noisy}}$}& 44.7\%± 10.1\% & \multicolumn{1}{c|}{1.99}& 35.6\%± 8.5\% & \multicolumn{1}{c|}{1.77}& 41.8\%± 9.4\% & \multicolumn{1}{c|}{2.83} \\ \hline 
\multicolumn{7}{|c|}{\textbf{Baselines} $10 \times 10$ } \\ \hline 
\multicolumn{1}{|c|}{L2D} & 38.2\%± 4.6\% & \multicolumn{1}{c|}{1.84}& 32.7\%± 5.1\% & \multicolumn{1}{c|}{1.68}& 37.3\%± 3.9\% & \multicolumn{1}{c|}{2.7} \\ 
\multicolumn{1}{|c|}{BC-$D_{\text{expert}}$} & 43.8\%± 8.8\% & \multicolumn{1}{c|}{1.96}& 36.0\%± 9.1\% & \multicolumn{1}{c|}{1.76}& 40.5\%± 7.3\% & \multicolumn{1}{c|}{2.89} \\ \hline 
\multicolumn{7}{|c|}{\textbf{Offline-LD} $10 \times 10$ } \\ \hline 
\multicolumn{1}{|c|}{mQRDQN-$D_{\text{expert}}$}& \textbf{37.4\%± 2.6\%} & \multicolumn{1}{c|}{1.87}& 35.0\%± 2.6\% & \multicolumn{1}{c|}{1.68}& 37.6\%± 2.4\% & \multicolumn{1}{c|}{2.81} \\ 
\multicolumn{1}{|c|}{d-mSAC-$D_{\text{expert}}$}& 37.3\%± 3.5\% & \multicolumn{1}{c|}{1.97}& \textbf{32.0\%± 4.7\%} & \multicolumn{1}{c|}{1.8}& \textbf{36.1\%± 2.7\%} & \multicolumn{1}{c|}{2.96} \\ 
\multicolumn{1}{|c|}{mQRDQN-$D_{\text{noisy}}$}& 36.9\%± 3.2\% & \multicolumn{1}{c|}{1.83}& 34.2\%± 2.3\% & \multicolumn{1}{c|}{1.63}& 37.6\%± 2.5\% & \multicolumn{1}{c|}{2.77} \\ 
\multicolumn{1}{|c|}{d-mSAC-$D_{\text{noisy}}$}& 52.1\%± 7.7\% & \multicolumn{1}{c|}{2.0}& 41.1\%± 7.1\% & \multicolumn{1}{c|}{1.76}& 47.9\%± 7.3\% & \multicolumn{1}{c|}{2.89} \\ \hline 
\multicolumn{7}{|c|}{\textbf{Baselines} $15 \times 15$ } \\ \hline 
\multicolumn{1}{|c|}{L2D} & 38.7\%± 2.7\% & \multicolumn{1}{c|}{1.87}& 33.6\%± 4.3\% & \multicolumn{1}{c|}{1.72}& 37.7\%± 2.8\% & \multicolumn{1}{c|}{2.74} \\ 
\multicolumn{1}{|c|}{BC-$D_{\text{expert}}$} & 50.0\%± 7.8\% & \multicolumn{1}{c|}{1.94}& 37.5\%± 6.5\% & \multicolumn{1}{c|}{1.77}& 43.4\%± 7.1\% & \multicolumn{1}{c|}{2.84} \\ \hline 
\multicolumn{7}{|c|}{\textbf{Offline-LD} $15 \times 15$ } \\ \hline 
\multicolumn{1}{|c|}{mQRDQN-$D_{\text{expert}}$}& 38.6\%± 3.1\% & \multicolumn{1}{c|}{1.84}& 34.9\%± 3.1\% & \multicolumn{1}{c|}{1.66}& 38.4\%± 3.1\% & \multicolumn{1}{c|}{2.71} \\ 
\multicolumn{1}{|c|}{d-mSAC-$D_{\text{expert}}$}& 39.5\%± 4.6\% & \multicolumn{1}{c|}{1.97}& 34.3\%± 3.9\% & \multicolumn{1}{c|}{1.78}& 36.9\%± 4.4\% & \multicolumn{1}{c|}{2.82} \\ 
\multicolumn{1}{|c|}{mQRDQN-$D_{\text{noisy}}$}& 38.6\%± 2.1\% & \multicolumn{1}{c|}{1.84}& 33.9\%± 1.8\% & \multicolumn{1}{c|}{1.67}& 39.5\%± 2.4\% & \multicolumn{1}{c|}{2.71} \\ 
\multicolumn{1}{|c|}{d-mSAC-$D_{\text{noisy}}$}& 45.5\%± 10.5\% & \multicolumn{1}{c|}{1.97}& 39.7\%± 8.5\% & \multicolumn{1}{c|}{1.77}& 44.8\%± 9.9\% & \multicolumn{1}{c|}{2.84} \\ \hline 
\multicolumn{7}{|c|}{\textbf{Baselines} $20 \times 20$ } \\ \hline 
\multicolumn{1}{|c|}{L2D} & 39.2\%± 4.2\% & \multicolumn{1}{c|}{1.9}& 33.2\%± 5.9\% & \multicolumn{1}{c|}{1.75}& 37.7\%± 4.7\% & \multicolumn{1}{c|}{2.77} \\ 
\multicolumn{1}{|c|}{BC-$D_{\text{expert}}$} & 46.6\%± 8.2\% & \multicolumn{1}{c|}{1.97}& 36.7\%± 7.1\% & \multicolumn{1}{c|}{1.78}& 42.0\%± 7.3\% & \multicolumn{1}{c|}{2.91} \\ \hline 
\multicolumn{7}{|c|}{\textbf{Offline-LD} $20 \times 20$ } \\ \hline 
\multicolumn{1}{|c|}{mQRDQN-$D_{\text{expert}}$}& 38.9\%± 2.7\% & \multicolumn{1}{c|}{1.88}& 33.8\%± 2.6\% & \multicolumn{1}{c|}{1.68}& 39.4\%± 2.9\% & \multicolumn{1}{c|}{2.63} \\ 
\multicolumn{1}{|c|}{d-mSAC-$D_{\text{expert}}$}& 38.8\%± 4.1\% & \multicolumn{1}{c|}{2.02}& 33.3\%± 2.5\% & \multicolumn{1}{c|}{1.81}& 38.8\%± 2.2\% & \multicolumn{1}{c|}{2.98} \\ 
\multicolumn{1}{|c|}{mQRDQN-$D_{\text{noisy}}$}& 38.5\%± 2.4\% & \multicolumn{1}{c|}{1.88}& 34.1\%± 2.0\% & \multicolumn{1}{c|}{1.69}& 38.9\%± 2.4\% & \multicolumn{1}{c|}{2.73} \\ 
\multicolumn{1}{|c|}{d-mSAC-$D_{\text{noisy}}$}& 41.7\%± 5.2\% & \multicolumn{1}{c|}{1.98}& 36.2\%± 4.8\% & \multicolumn{1}{c|}{1.81}& 42.8\%± 5.3\% & \multicolumn{1}{c|}{2.94} \\ \hline 
\multicolumn{7}{|c|}{\textbf{Baselines} $30 \times 20$ } \\ \hline 
\multicolumn{1}{|c|}{L2D} & 38.8\%± 2.3\% & \multicolumn{1}{c|}{1.91}& 34.6\%± 1.1\% & \multicolumn{1}{c|}{1.74}& 38.5\%± 1.6\% & \multicolumn{1}{c|}{2.81} \\ 
\multicolumn{1}{|c|}{BC-$D_{\text{expert}}$} & 46.3\%± 6.3\% & \multicolumn{1}{c|}{1.93}& 36.0\%± 4.5\% & \multicolumn{1}{c|}{1.72}& 39.9\%± 4.6\% & \multicolumn{1}{c|}{2.79} \\ \hline 
\multicolumn{7}{|c|}{\textbf{Offline-LD} $30 \times 20$ } \\ \hline 
\multicolumn{1}{|c|}{mQRDQN-$D_{\text{expert}}$}& 38.4\%± 4.1\% & \multicolumn{1}{c|}{1.9}& 33.6\%± 3.7\% & \multicolumn{1}{c|}{1.69}& 37.4\%± 4.1\% & \multicolumn{1}{c|}{2.77} \\ 
\multicolumn{1}{|c|}{d-mSAC-$D_{\text{expert}}$}& 39.2\%± 4.0\% & \multicolumn{1}{c|}{2.01}& 34.0\%± 3.3\% & \multicolumn{1}{c|}{1.78}& 38.7\%± 3.0\% & \multicolumn{1}{c|}{2.93} \\ 
\multicolumn{1}{|c|}{mQRDQN-$D_{\text{noisy}}$}& 39.2\%± 2.4\% & \multicolumn{1}{c|}{1.85}& 34.1\%± 1.4\% & \multicolumn{1}{c|}{1.65}& 40.1\%± 1.9\% & \multicolumn{1}{c|}{2.77} \\ 
\multicolumn{1}{|c|}{d-mSAC-$D_{\text{noisy}}$}& 43.6\%± 7.6\% & \multicolumn{1}{c|}{2.0}& 38.3\%± 7.1\% & \multicolumn{1}{c|}{1.81}& 43.4\%± 7.8\% & \multicolumn{1}{c|}{2.89} \\ \hline 
\multicolumn{1}{|c|}{Opt. Rate (\%)} & \multicolumn{2}{c|}{50\%} & \multicolumn{2}{c|}{50\%} & \multicolumn{2}{c|}{50\%}  \\ \hline 
\end{tabular}

\label{tab:dmu_big}
\end{table*}

\end{appendices}



\end{document}